\documentclass[3p,oneThankscolumn]{elsarticle}

\makeatletter
\def\ps@pprintTitle{%
 \let\@oddhead\@empty
 \let\@evenhead\@empty
 \def\@oddfoot{\centerline{\thepage}}%
 \let\@evenfoot\@oddfoot}
\makeatother

\usepackage{filecontents}
\usepackage{amsthm}%
\usepackage{manyfoot}%
\usepackage{float}

\usepackage{stmaryrd}
\usepackage{url}
\usepackage{color}
\usepackage[export]{adjustbox}
\usepackage{xspace}
\usepackage{multirow}
\usepackage{dingbat}
\usepackage{csquotes}
\usepackage{textcomp}
\usepackage[table]{xcolor}
\usepackage{tabulary}

\usepackage{xspace}
\usepackage{mathtools}
\usepackage{soul}
\usepackage{hyperref}
\usepackage{fontawesome}

\newcommand{\latinphrase}[1]{\textit{#1}} 
\newcommand{\etal}{\latinphrase{et~al.}\xspace}

\newcommand{\etc}{\latinphrase{etc.}\xspace}

\begin{document}

\title{Image Colorization: A Survey and Dataset}
\author[1,2]{Saeed Anwar\fnref{fn1}}
\ead{saeed.anwar@anu.edu.au}
\author[3]{Muhammad Tahir}
\ead{m.tahir@nbc.nust.edu.pk}
\author[4]{Chongyi Li}
\ead{lichongyi@nankai.edu.cn}
\author[5]{Ajmal Mian}
\ead{ajmal.mian@uwa.edu.au}
\author[6,7]{Fahad Shahbaz Khan}
\ead{fahad.khan@mbzuai.ac.ae}
\author[8]{Abdul Wahab Muzaffar}
\ead{a.muzaffar@seu.edu.sa}

\address[1]{The Australian National University, Canberra, ACT, Australia}
\address[2]{The University of Canberra, Canberra, Act, Australia}
\address[3]{Department of Computer Science, National University of Sciences and Technology (NUST), Balochistan Campus, Quetta, Pakistan}
\address[4]{Nankai University, China}
\address[5]{The University of Western Australia, Australia}
\address[6]{Mohamed Bin Zayed University of Artificial Intelligence, Abu Dhabi, UAE}
\address[7]{Computer Vision Laboratory, Linkoping University, Sweden}
\address[8]{College of Computing and Informatics, Saudi Electronic University, Saudi Arabia}

\fntext[fn1]{Corresponding author: Saeed Anwar}

\begin{abstract}
Image colorization estimates RGB colors for grayscale images or video frames to improve their aesthetic and perceptual quality. Over the last decade, deep learning techniques for image colorization have significantly progressed, necessitating a systematic survey and benchmarking of these techniques. This article presents a comprehensive survey of recent state-of-the-art deep learning-based image colorization techniques, describing their fundamental block architectures, inputs, optimizers, loss functions, training protocols, training data, etc. It categorizes the existing colorization techniques into seven classes and discusses important factors governing their performance, such as benchmark datasets and evaluation metrics. We highlight the limitations of existing datasets and introduce a new dataset specific to colorization. We perform an extensive experimental evaluation of existing image colorization methods using both existing datasets and our proposed one. Finally, we discuss the limitations of existing methods and recommend possible solutions and future research directions for this rapidly evolving topic of deep image colorization. The dataset and codes for evaluation are publicly available at https://github.com/saeed-anwar/ColorSurvey.

\begin{keyword}
Image colorization, experimental survey, new colorization dataset, colorization review,  deep learning, CNN model classification.
\end{keyword}

\end{abstract}


\maketitle

\section{Introduction}
\label{sec:introduction}
\begin{flushright}
\enquote{Don\textquotesingle t ask what love can make or do! Look at the colors of the world.} - Rumi
\end{flushright}

Colorization of images is a challenging problem due to the varying conditions of imaging that need to be dealt with via a single algorithm. The problem is also severely ill-posed as two out of the three image dimensions are missing; although the scene semantics may be helpful in many cases, for example, the grass is usually green, clouds are usually white, and the sky is blue. However, such semantic priors are uncommon for many manufactured and natural objects, e.g., shirts, cars, flowers, etc. Moreover, the colorization problem also inherits the typical challenges of image enhancement, such as changes in illumination, variations in viewpoints, and occlusions.

With the rapid development of deep learning techniques, various image colorization models have been introduced, and state-of-the-art performance on current datasets has been reported. Diverse deep-learning models ranging from the early brute-force networks (e.g.,~\cite{cheng2015deep}) to the recently carefully designed Generative Adversarial Networks (GAN) (e.g.,~\cite{yoo2019coloring}) have been successfully used to tackle the colorization problem. These colorization networks differ in many major aspects, including network architecture, network depth, loss functions, learning strategies, etc.
This paper provides a comprehensive overview of single-image colorization and focuses on recent advances in deep neural networks for this task. To our knowledge, no survey of conventional or deep learning-based colorization techniques has been presented in the current literature. Our study concentrates on many important aspects, both systematically and comprehensively, in benchmarking recent advances in deep learning-based image colorization.
\vspace*{2mm}
\noindent
\textbf{Contributions:} Our contributions are as follows
\begin{enumerate}
\item We thoroughly review image colorization techniques, including problem settings, performance metrics and datasets. 

\item We propose a new taxonomy of colorization networks based on the differences in domain type, network structure, auxiliary input, and final output. 

\item We introduce a new benchmark dataset, the Natural-Color Dataset (NCD),  collected specifically for the colorization task.

\item We systematically evaluate state-of-the-art models on our Natural-Color Dataset.

\item We summarize the vital components of networks, provide new insights and discuss the challenges and possible future directions for image colorization. 
\end{enumerate}

\begin{figure}[h]
\begin{center}
\begin{tabular}{c} 
\includegraphics[trim={0cm 0cm  0cm  0cm}, clip, width=9cm, height= 14cm]{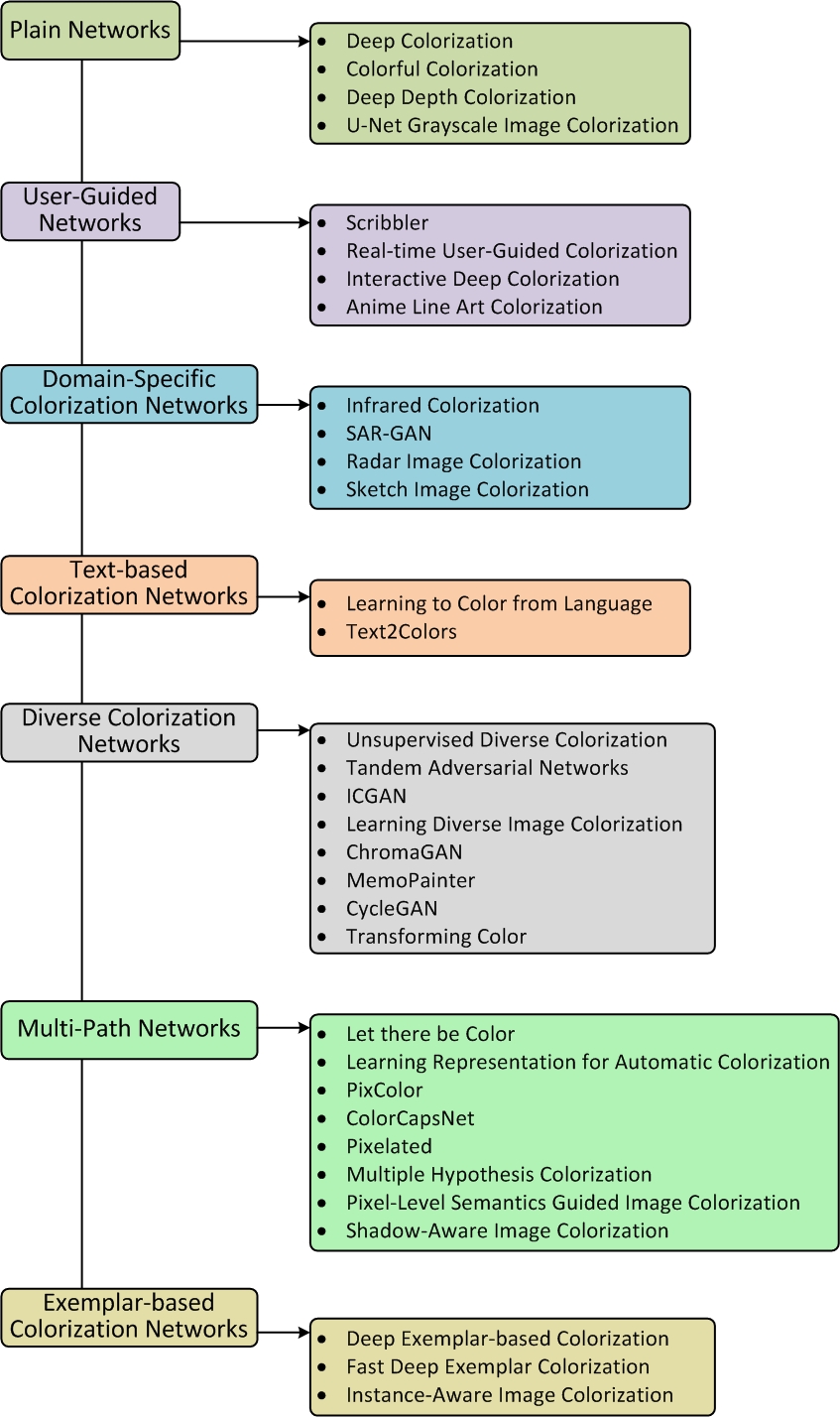}\\
\end{tabular}
\end{center}
\caption{\textbf{Taxonomy:} Classification of the colorization networks based on structure, input, domain, and type of network.}
\label{fig:Taxonomy}
\end{figure}

\noindent
\textbf{Need for a Survey:}  
Image colorization has been the focus of significant research efforts over the last two decades. Whereas most early image colorization methods were primarily influenced by conventional machine learning approaches~\cite{welsh2002transferring,levin2004colorization,huang2005adaptive,qu2006manga,yatziv2006fast,luan2007natural} - the past few years have seen a massive shift in focus to deep learning-based models due to their success in several different application domains~\cite{cheng2015deep,parnamaa2017accurate,xiao2019application,8416973,8627998,liu2022tt-cite1, wang2018cyber-cite2,wang2020mmdp-cite3,jiang2023medical-cite4}.
Automatic image colorization systems built upon deep learning have recently shown impressive performance. However, despite this success, no literature survey or review currently covers the latest developments in this field. Moreover, our goal is to streamline the direction of image colorization research, put existing research in context, identify gaps, and point out the focus areas for the future.
Therefore, inspired by surveys in deep image super-resolution~\cite{Saeed2020SR}, VQA~\cite{Aafaq2019VQA}, etc., we provide a comprehensive survey of deep image colorization.

\vspace*{2mm}
\noindent
\textbf{Need for a Dataset:} Datasets play a critical role in benchmarking the methods. The current datasets employed for evaluating algorithms in this field of research are not specific to colorization. The testing images may contain subjective colors such as walls, shirts, etc., and may have false colors; for example, the grass is blue. Similarly, the algorithms may be biased to training data, focus on a single object, color, or background, or spill the colors between objects or objects and the background. Hence, without a colorization-specific dataset, the evaluation may not accurately represent the algorithms. Therefore, we provide a colorization dataset and benchmark the state-of-the-art methods. We give further information about the dataset in the experimental section.

\vspace*{2mm}
\noindent
\textbf{Problem Formulation:} 
Colorization aims to estimate the RGB colors of a grayscale image, typically captured before color cameras were widely available and technological advancements were limited. Hence, this process is more of an image enhancement than an image restoration process. Another use for image colorization is to restore color images after they have been converted to grayscale or Y-Channel of YUV color space, for example, to save storage space or communication bandwidth. Therefore, in this case, a trivial formulation can be  written as:

\begin{equation}
I_g = \Phi(I_{rgb}),
\label{eq:naive}
\end{equation}
where $\Phi(\cdot)$ is a function that converts the RGB image $I_{rgb}$ to a grayscale image $I_g$, for example as follows:
\begin{equation}
I_g = 0.2989 * I_r + 0.5870 * I_g + 0.1140 * I_b. 
\label{eq:Convert}
\end{equation}

Typically, colorization methods aim to restore the color in YUV space, where the model needs to predict only two channels, i.e., U and V - instead of the three channels in RGB.

\section{Single-Image Deep Colorization}
\label{sec:SIDC}
This section introduces various deep-learning techniques for image colorization. As shown in Figure~\ref{fig:Taxonomy}, these colorization networks have been classified into various categories in terms of different factors, such as structural differences, input type, user assistance, etc. Some networks may belong to multiple categories; however, we assign each method to the most appropriate category. We also provide the strengths and weaknesses of each category in Table~\ref{tab:Strenght_weakness}.

\subsection{Plain Networks}
\begin{figure}[tbp]
\begin{center}
\begin{tabular}{cc} 
\includegraphics[trim={4.5cm 7.8cm  5.5cm  5.2cm },clip,width=0.5\columnwidth,valign=t]{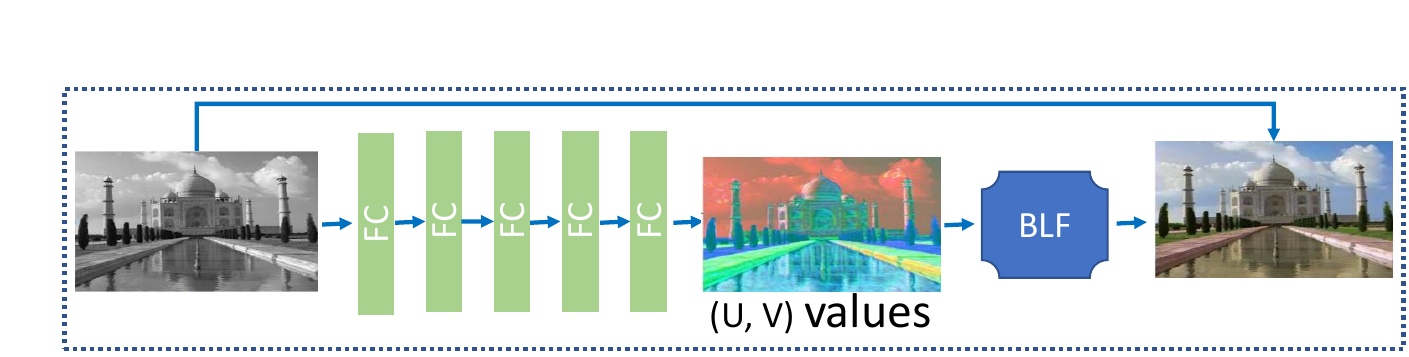}&

\includegraphics[trim={8.2cm 6.5cm  6.5cm  6.5cm },clip,width=0.5\columnwidth,valign=t]{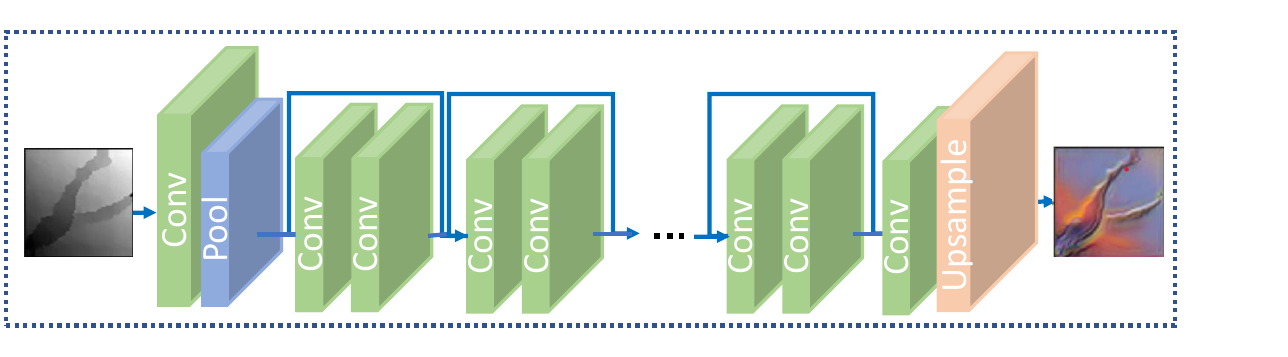}\\
Deep Colorization~\cite{cheng2015deep} & Depth Colorization (DE2CO)~\cite{carlucci20182}\\
\multicolumn{2}{c}{\includegraphics[trim={1cm 7.5cm  2.5cm  4cm },clip,width=0.5\columnwidth,valign=t]{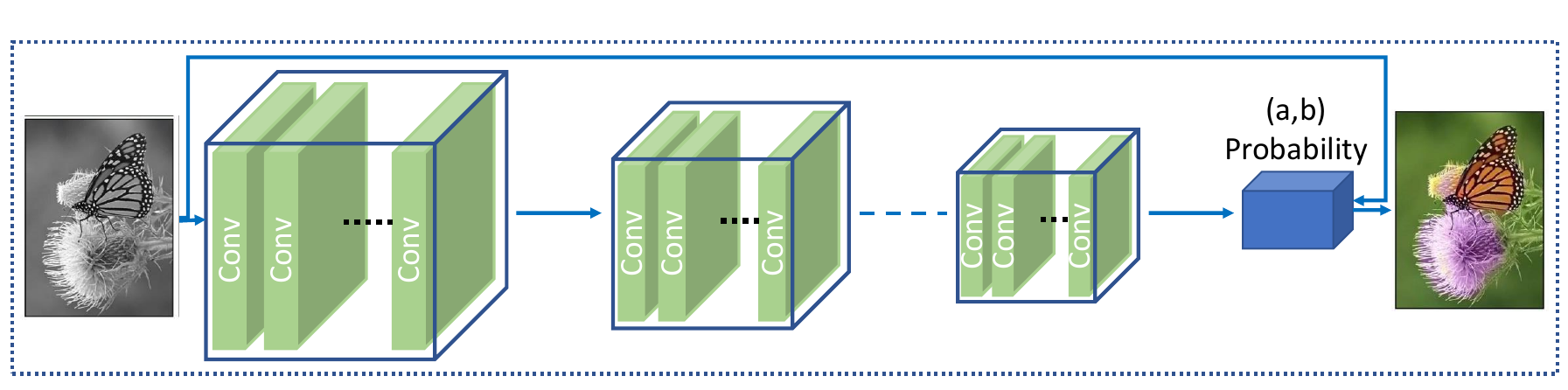}}\\
\\
\multicolumn{2}{c}{Colorful Colorization~\cite{zhang2016colorful}}\\
\end{tabular}
\end{center}
\caption{\textbf{Plain networks} are the earlier models with convolutional layer stacking with no skip or naive skip connections.}
\label{fig:plainNetworks}
\end{figure}

Similar to other CNN tasks, early colorization architectures were plain networks with a simple, straightforward architecture that stacked convolutional layers with no or naive skip connections. Networks that fall into this category are shown in Figure~\ref{fig:plainNetworks} and discussed below. These networks are straightforward to implement; however, the learning process is slow and requires more training data.

\subsubsection{Deep Colorization}
Deep colorization\footnote{Code available at \url{https://shorturl.at/cestD}}~\cite{cheng2015deep} can be regarded as the first method to incorporate convolutional neural networks (CNNs) for the colorization of images. This method, however, does not fully exploit CNNs; instead, it includes joint bilateral filtering~\cite{petschnigg2004digital} as a post-processing step to remove artifacts introduced by the CNN network.

In the training stage, five fully connected linear layers are followed by non-linear activations (ReLU), and the loss function is the least-squares error. In their model, the number of neurons in the first layer depends on the dimensions of the feature descriptor extracted from the grayscale patch, while the output layer has only two neurons, i.e., the U and V channel corresponding color pixel values. During testing/inference, features are extracted from the input grayscale image at three levels, i.e., low-level, mid-level, and high-level. The features at the low level are the sequential gray values, at the mid-level are DAISY features~\cite{tola2008fast}, and at the high level, semantic labeling is performed. The patch and DAISY features are concatenated and then passed through the network. As a final step for removing artifacts, joint bilateral filtering~\cite{petschnigg2004digital} is performed.


The network takes as input a 256 $\times$ 256 grayscale image and is composed of five layers: one input layer, three hidden layers, and one output layer. The model is trained on 2688 images from the Sun dataset~\cite{xiao2010sun}. The images are segmented into 47 object categories, including cars, buildings, sea, etc., for 47 high-level semantics. Furthermore, 32-dimensional mid-level DAISY features~\cite{tola2008fast} and 49-dimensional low-level features are utilized for colorization.

\subsubsection{Colorful Colorization}

Colorful image Colorization CNN\footnote{Code at \url{https://github.com/richzhang/colorization}}~\cite{zhang2016colorful} was one of the first attempts to colorize grayscale images. The network takes as input a grayscale image and predicts 313 \enquote{ab} pairs of the gamut showing the empirical probability distribution, which is then transformed to \enquote{a} and \enquote{b} channels of the \enquote{Lab} color space. The network linearly stacks convolutional layers, forming eight blocks. Each block comprises two or three convolutional layers followed by a ReLU layer and Batch Normalization (BatchNorm~\cite{ioffe2015batch}) layer. Instead of pooling, striding is used to decrease the size of the image. The input to the network is 256$\times$256, while the output is 224$\times$224; however, it is resized later to the original image size.

Colorful image colorization removes the influence (due to the background, e.g., sky, clouds, walls, etc.) of the low \enquote{ab} values by re-weighting the loss based on the rarity of pixels in training. The authors term this technique class rebalancing. The framework utilized to build the network is Caffe~\cite{jia2014caffe}, and ADAM~\cite{kingma2014adam} is used as the solver. The network is trained for 450k iterations with a learning rate of $3\times 10^{-5}$, which is reduced to $10^{-5}$ at 200k and $3\times 10^{-6}$ at 375k iterations. The kernel size is 3$\times$3, and the feature channels vary from 64 to 512.

\subsubsection{Deep Depth Colorization}
Deep depth colorization\footnote{Available at \url{https://github.com/engharat/SBADAGAN}} (DE)$^2$CO~\cite{carlucci20182} employs a pretrained ImageNet~\cite{deng2009imagenet} architecture for colorizing depth images. The system is primarily designed for object recognition by learning the mapping from the depths to RGB channels. The pretrained network weights are kept frozen, and only the last fully connected layer of (DE)$^2$CO is trained to classify the objects with a softmax classifier. The pretrained networks are merely used as feature extractors.

The input to the network is a 228$\times$228 depth map reduced via convolution followed by pooling to $64\times 57\times 57$. Subsequently, the features are passed through a series of residual blocks, composed of two convolutional layers, each followed by batch normalization~\cite{ioffe2015batch} and a non-linear activation function, i.e., leaky-ReLU~\cite{maas2013rectifier}. After the last residual block, the output is passed through a final convolutional layer to produce three channels, i.e., an RGB image as an output. To obtain the original resolution, the output is deconvolved as a final step. When an unseen dataset is encountered, only the last convolutional layer is retrained while keeping the weights across all other layers frozen. (DE)$^2$CO outperforms CaffeNet (a variant of Alexnet~\cite{krizhevsky2012AlexNet}), VGG16~\cite{simonyan2015vgg}, GoogleNet~\cite{googlenet}, and ResNet50~\cite{he2016ResNet} under the same settings on three benchmark datasets, including Washington-RGBD~\cite{lai2011WFGBD}, JHUIT50~\cite{li2015JHUIT50}, and BigBIRD~\cite{singh2014bigbird}.

\subsubsection{U-Net Grayscale Image Colorization}
Hu~\etal~\cite{hu2024grayscale-new2} utilized a fully convolutional neural network technique for grayscale image colorization tasks that involve the transformation of the color space followed by preprocessing of the output grayscale image and realization of an enhanced U-Net. The color space of input images is converted from RGB to Lab color space. Next, the output grayscale image is preprocessed using Heckbert's median cut method, which is used for gray-level quantization. Finally, the U-Net architecture is adopted to colorize the images. The U-Net network is trained on images from the Lab color space. Once trained, the network can generate pragmatic colors for grayscale images using the L component of the Lab color model after quantization with the median cut method. The outputs of each U-Net convolutional layer are obtained using the logistic activation function. The proposed method's performance was evaluated on the ImageNet database, and it achieved 84.81\% accuracy.

\subsection{User-guided networks}
\begin{figure}[t]
\begin{center}
\begin{tabular}{cc} 
\includegraphics[trim={5cm 6.5cm  10cm  0cm },clip,width=0.5\columnwidth,valign=t]{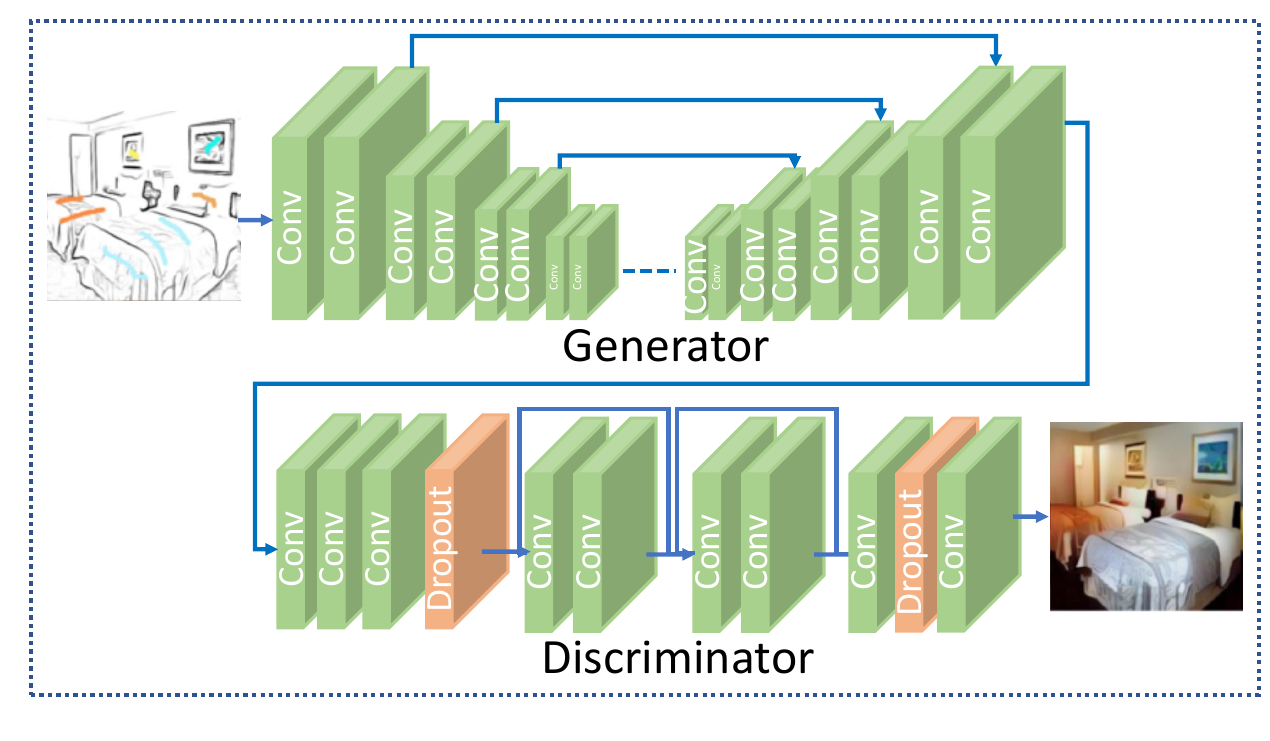}&

\includegraphics[trim={7.5cm 6.5cm  3.5cm  -2cm },clip,width=0.5\columnwidth,valign=t]{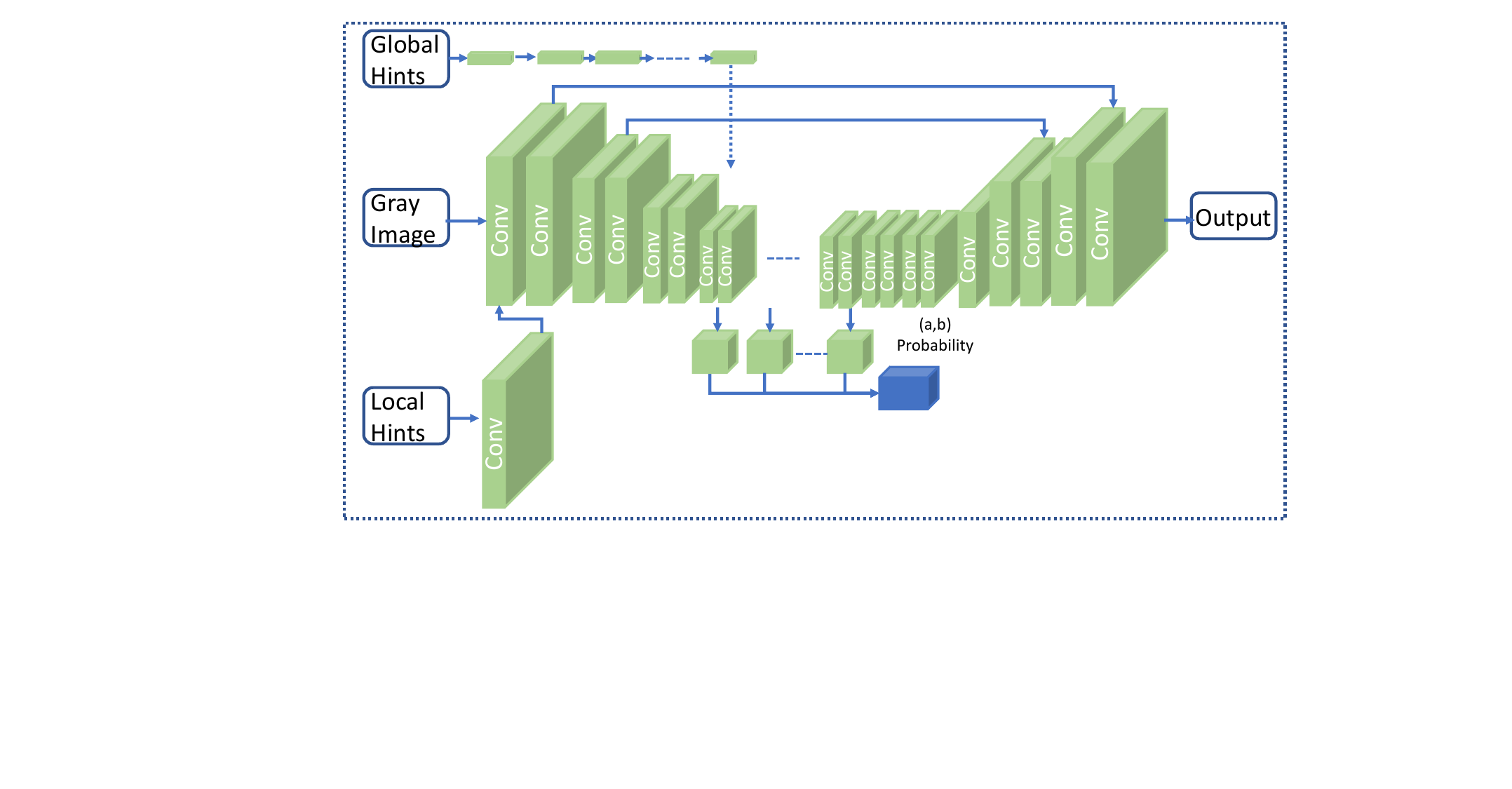}\\
Scribbler architecture~\cite{sangkloy2017scribbler} & Real-Time User guided Colorization~\cite{zhang2017real}\\
\includegraphics[trim={3.5cm 7cm  2cm  0cm },clip,width=0.5\columnwidth,valign=t]{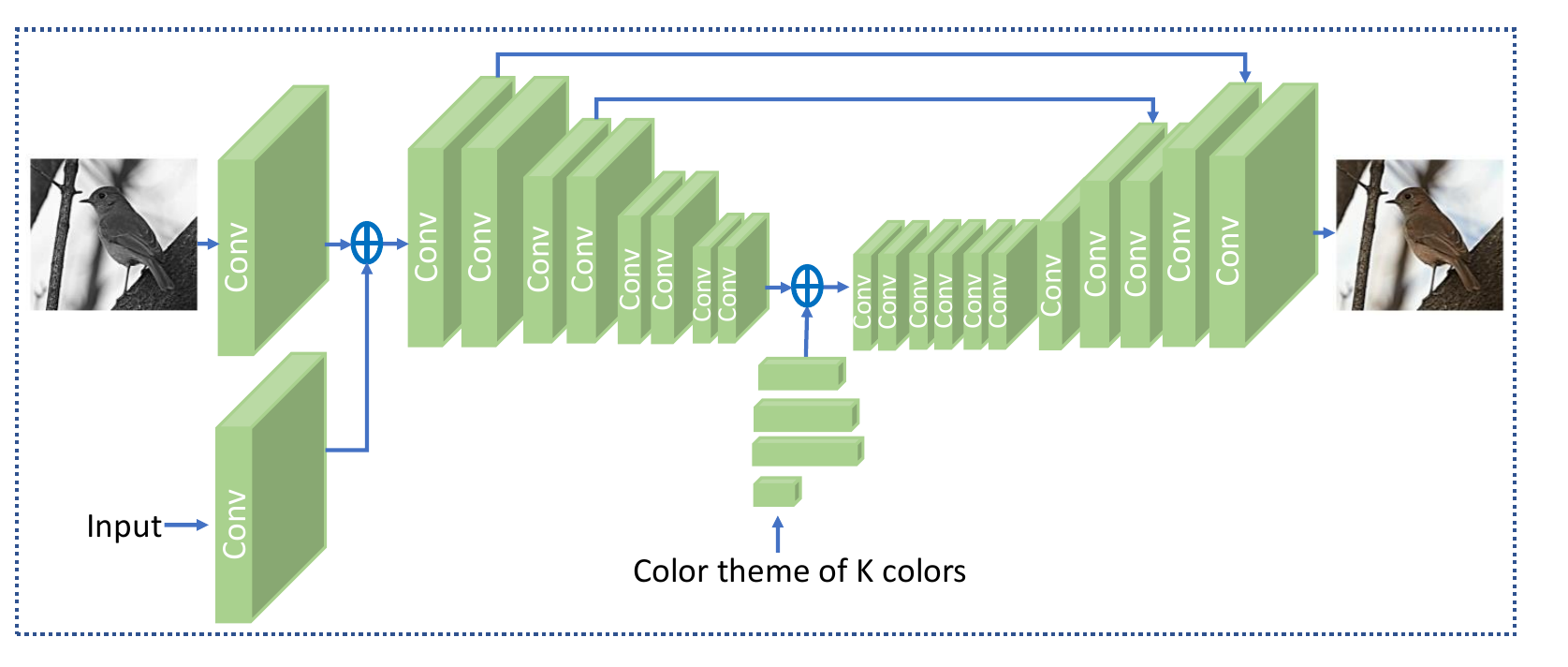}&
\includegraphics[trim={2cm 5cm  2cm  0cm },clip,width=0.5\columnwidth,valign=t]{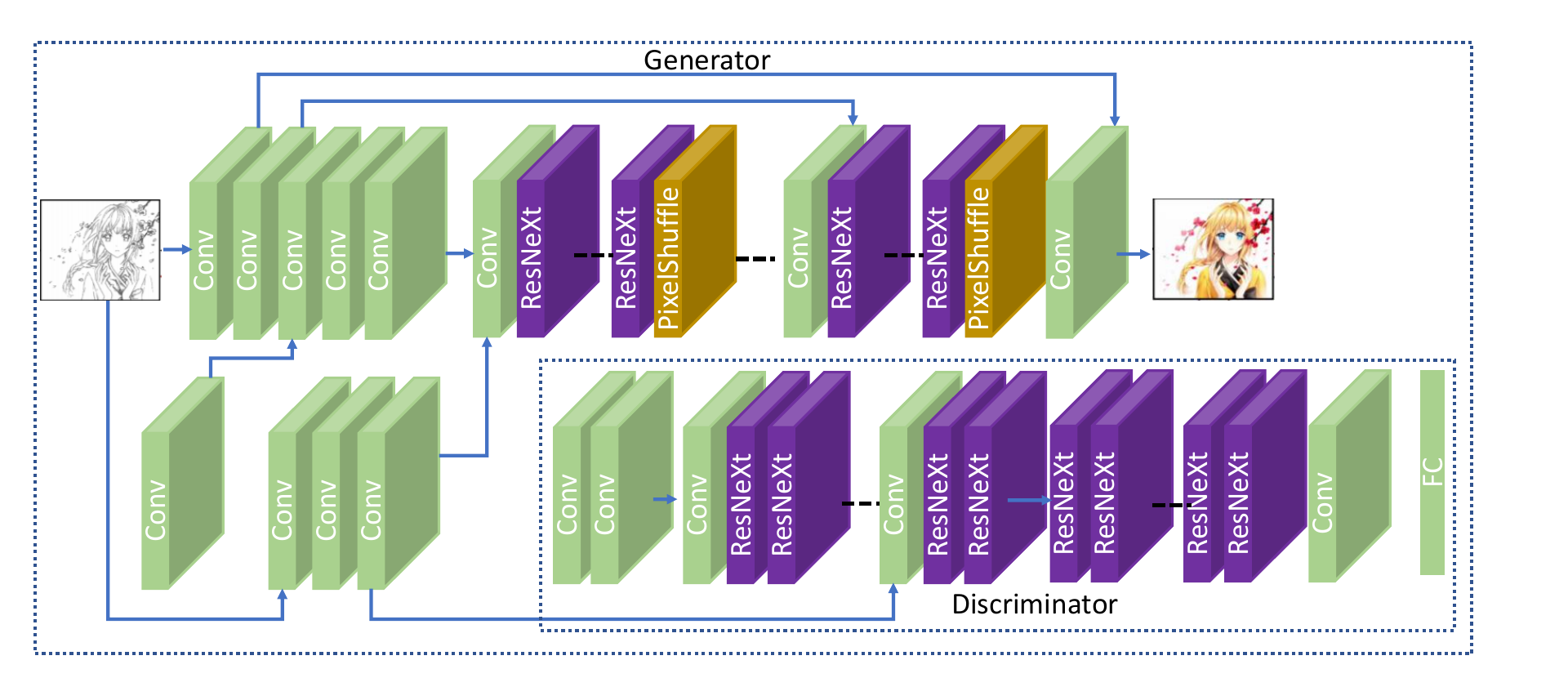}\\
\\
Interactive Deep Colorization~\cite{xiao2019interactive}& Hint-Guided Anime Colorization~\cite{ci2018user}\\
\end{tabular}
\end{center}
\caption{\textbf{User-Guided networks} are the ones that require a user to input the color at some stage of the network during colorization.}
\label{fig:UserGuided}
\end{figure}

User-guided networks require input from the user either in the form of points, strikes, or scribbles, as presented in Figure~\ref{fig:UserGuided}. Users can provide their input in real-time or offline. The following are examples of user-guided networks. User-guided colorization improves accuracy and provides control to users for direct input, enhancing complex and ambiguous regions colorization. Moreover, it may be useful for artistic purposes, precise historical restoration, handling small details and rare objects more effectively. However, all this is only possible due to increased user effort and a certain level of knowlege. This can be time-consuming and impractical for those without specialized knowledge. Although user-guided colorization is effective for individual images but these are less scalable for large datasets and can lead to inconsistent outputs when different users provide varying inputs for similar photos.

\subsubsection{Scribbler}
Sangkloy~\etal~\cite{sangkloy2017scribbler} used an end-to-end feed-forward deep generative adversarial architecture\footnote{\url{https://github.com/Pingxia/ConvolutionalSketchInversion}} to colorize images. To guide structural information and color patterns, user input in the form of sketches and color strokes is employed. The adversarial loss function enables the network to colorize the images more realistically.

The generator part of the proposed network adopts an encoder-decoder structure with residual blocks. Following the architecture of Sketch Inversion~\cite{guccluturk2016convolutional}, the augmented architecture consists of three downsampling layers and seven residual blocks preceded by three upsampling layers. The downsampling layers apply convolutions of stride two, whereas the upsampling layers utilize bilinear upsampling to substitute the deconvolutional layers, contrary to Sketch Inversion. All layers are followed by batch normalization~\cite{ioffe2015batch} and the ReLU activation function except the last layer, where the $TanH$ function is used. A fully convolutional network with five convolutional layers and two residual blocks makes up the discriminator part of the proposed network. Leaky-ReLU~\cite{maas2013rectifier} is used after each convolutional layer except the last one, where Sigmoid is applied. The model's performance is assessed qualitatively on datasets from three domains, including faces, cars, and bedrooms.

\subsubsection{Real-Time User-Guided Colorization}
Zhang~\etal \cite{zhang2017real} developed user interaction based on two variants, namely local hint and global hint networks, both of which utilize a common main branch for image colorization\footnote{\url{https://github.com/junyanz/interactive-deep-colorization}}. The local hint network is responsible for processing the user input and yielding a color distribution, whereas the global hint network accepts global statistics in the form of a global histogram and average image saturation. The loss functions include Huber and regression.

The main network comprises ten blocks, made up of two or three convolutional layers, followed by the ReLU. Similarly, each block is succeeded by batch normalization as well. Feature tensors are continuously halved in the first four blocks while doubling the feature dimensions. The process is repeated in reverse order for the last four convolutional blocks. Moreover, a dilated convolution (with a factor of two) is applied for the fifth and sixth blocks. Symmetric shortcut connections are also established between different blocks to recover spatial information. The kernel size for all convolutions is set to $3\times3$. However, at the last layer, a $1\times1$ kernel is used, mapping block ten and the final output.

\subsubsection{Interactive Deep Colorization}
To colorize grayscale images, Xiao~\etal~\cite{xiao2019interactive} developed an interactive colorization model based on the U-Net~\cite{ronneberger2015u} architecture, which can simultaneously utilize global and local inputs. The network includes a feature extraction module, a dilated module, a global input module, and a reconstruction module.

The feature extraction module (layers 2 to 14) receives the inputs, i.e., a grayscale image, local input, and gradient map merged via element-wise summation. Four convolution layers process the global input independently before combining it with the feature extraction module's output through element-wise summation. The dilated module takes the input from the extraction module (corresponding to convolutional layers ranging from 15 to 20). A reconstruction module, consisting of numerous deconvolution and convolution layers, further processes the output of the dilated module. The final step is combining the network's output with the input grayscale image to generate the colorized version. All the layers use ReLU except the final one, which employs a $TanH$ activation.

The proposed model modifies the Huber loss to fulfill its requirements. The testing set is composed of randomly chosen 1k images from ImageNet~\cite{deng2009imagenet}. The proposed model is trained for 300k iterations utilizing the remaining pictures from ImageNet~\cite{deng2009imagenet}, combined with 150k images from Places dataset~\cite{zhou2017places}.

\subsubsection{Anime Line Art Colorization}
Ci~\etal~\cite{ci2018user} proposed an end-to-end interactive deep conditional GAN (CGAN)~\cite{mirza2014cgan} for the colorization of synthetic anime line arts. The system operates on the user hints and the grayscale line art.

The discriminator is conditioned on the local features computed from a pre-trained network called Illustration2Vec~\cite{saito2015illustration2vec}. The generator adopts the architecture of U-Net~\cite{ronneberger2015u} that has two convolution blocks and the local feature network at the start. Afterward, four sub-networks with similar structures are employed, each of which is composed of convolutional layers at the front, followed by ResNeXt blocks~\cite{xie2017aggregated} with dilation and sub-pixel convolutional (PixelShuffle) layers. LeakyReLU serves as the activation function for each convolutional layer, except the final one, which employs $TanH$ activation. On the other hand, the discriminator is inspired by the architecture of SRGAN~\cite{ledig2017photo}; however, the basic blocks are replaced from the earlier mentioned generator to remove dilation, and an increased number of layers is used.

The generator employs the perceptual and adversarial losses, and the discriminator combines Wasserstein critic and penalty losses. The ADAM~\cite{kingma2014adam} optimizer is employed with $\beta\textsubscript{1}=0.5$, $\beta\textsubscript{1}=0.9$ and a batch size of four.

\subsection{Domain-Specific Colorization}
\begin{figure}[t]
\begin{center}
\begin{tabular}{cc} 
\includegraphics[trim={0cm 3cm  0cm  -2.5cm },clip,width=0.5\columnwidth,valign=t]{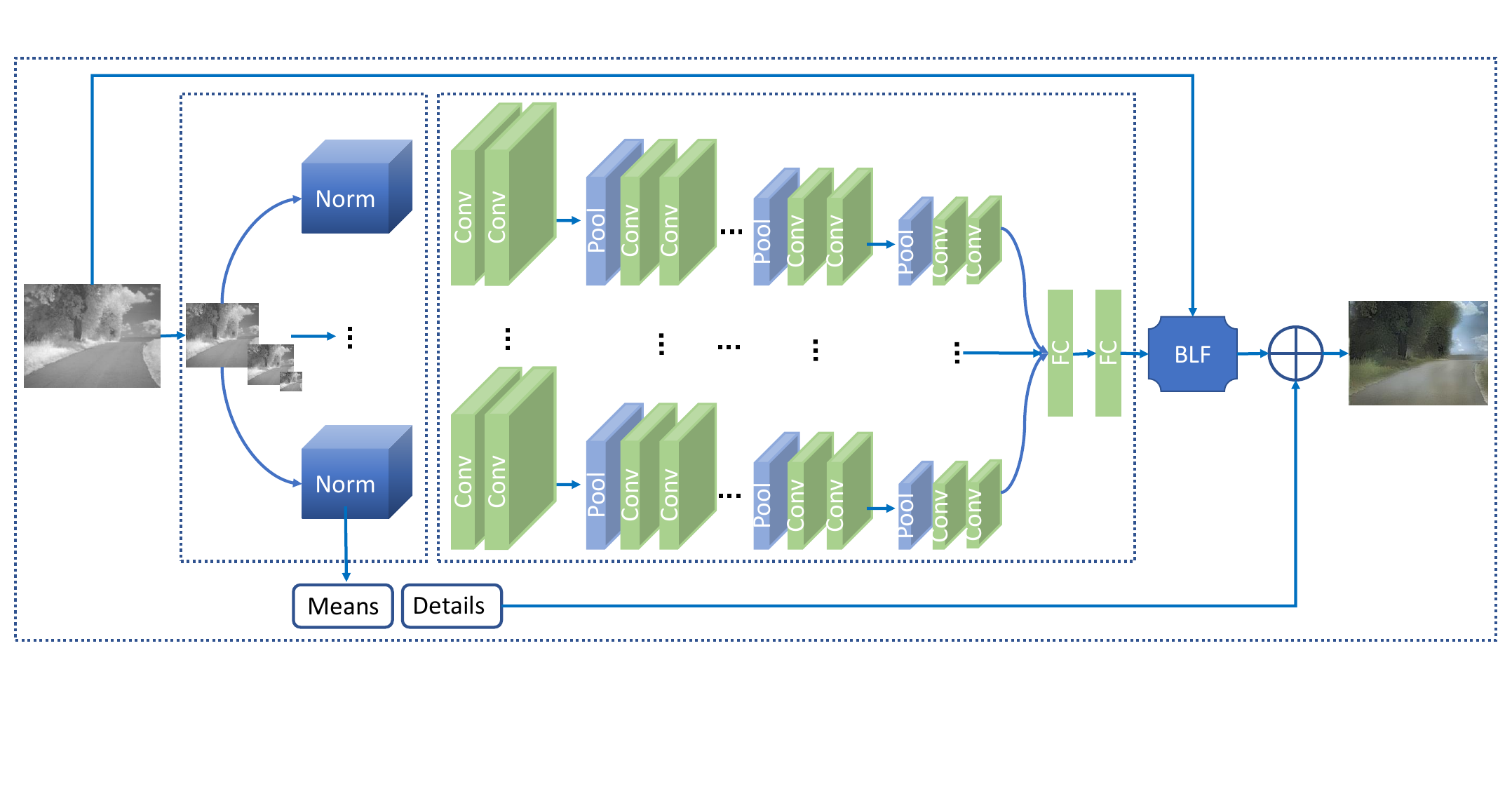}&
\includegraphics[trim={5cm 10cm  12cm  -2cm },clip,width=0.5\columnwidth,valign=t]{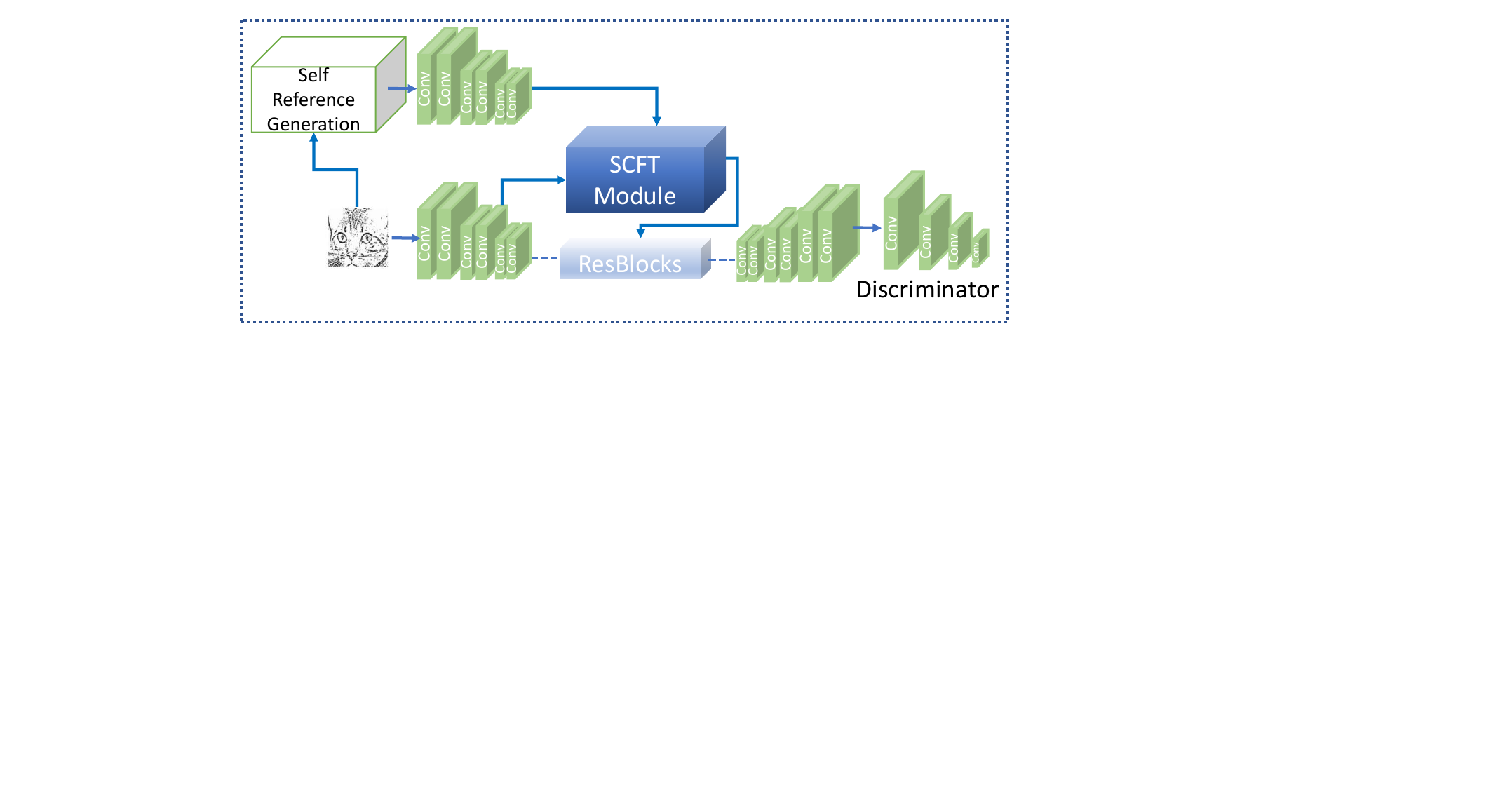}\\

NIR architecture\cite{limmer2016infrared} & SIC~\cite{lee2020CVPR}\\
\includegraphics[trim={7cm 7cm  10cm  0cm },clip,width=0.5\columnwidth,valign=t]{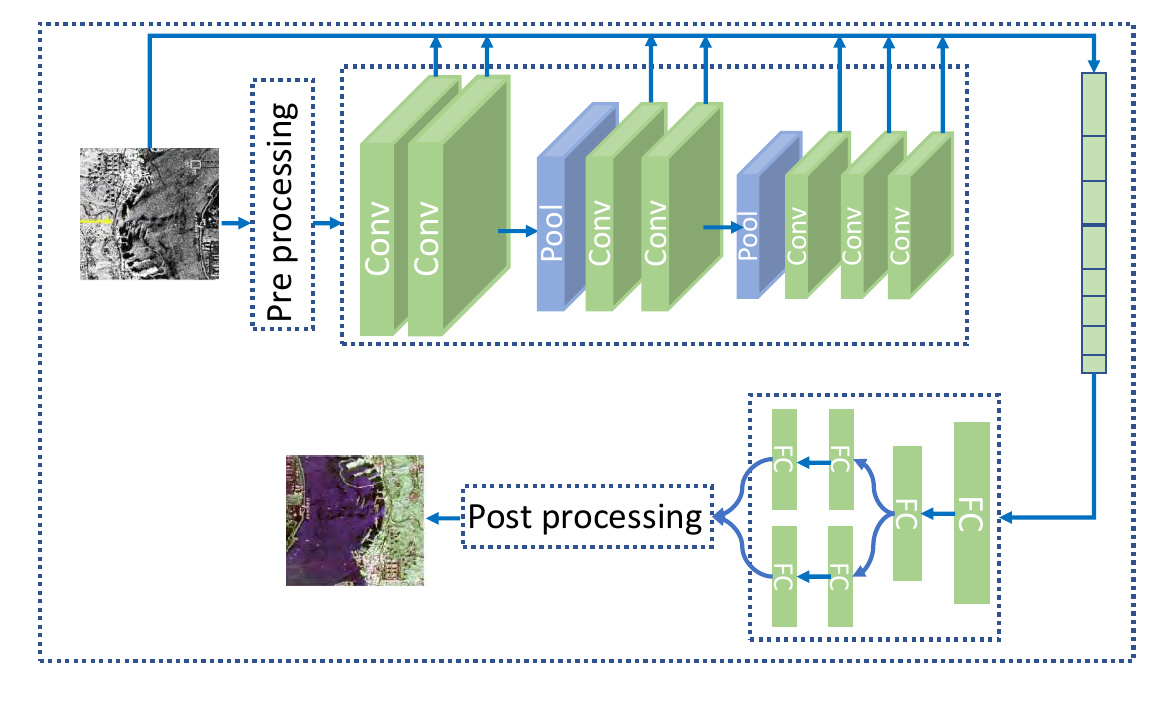}&
\includegraphics[trim={4cm 3.5cm  1cm  -1cm },clip,width=0.5\columnwidth,valign=t]{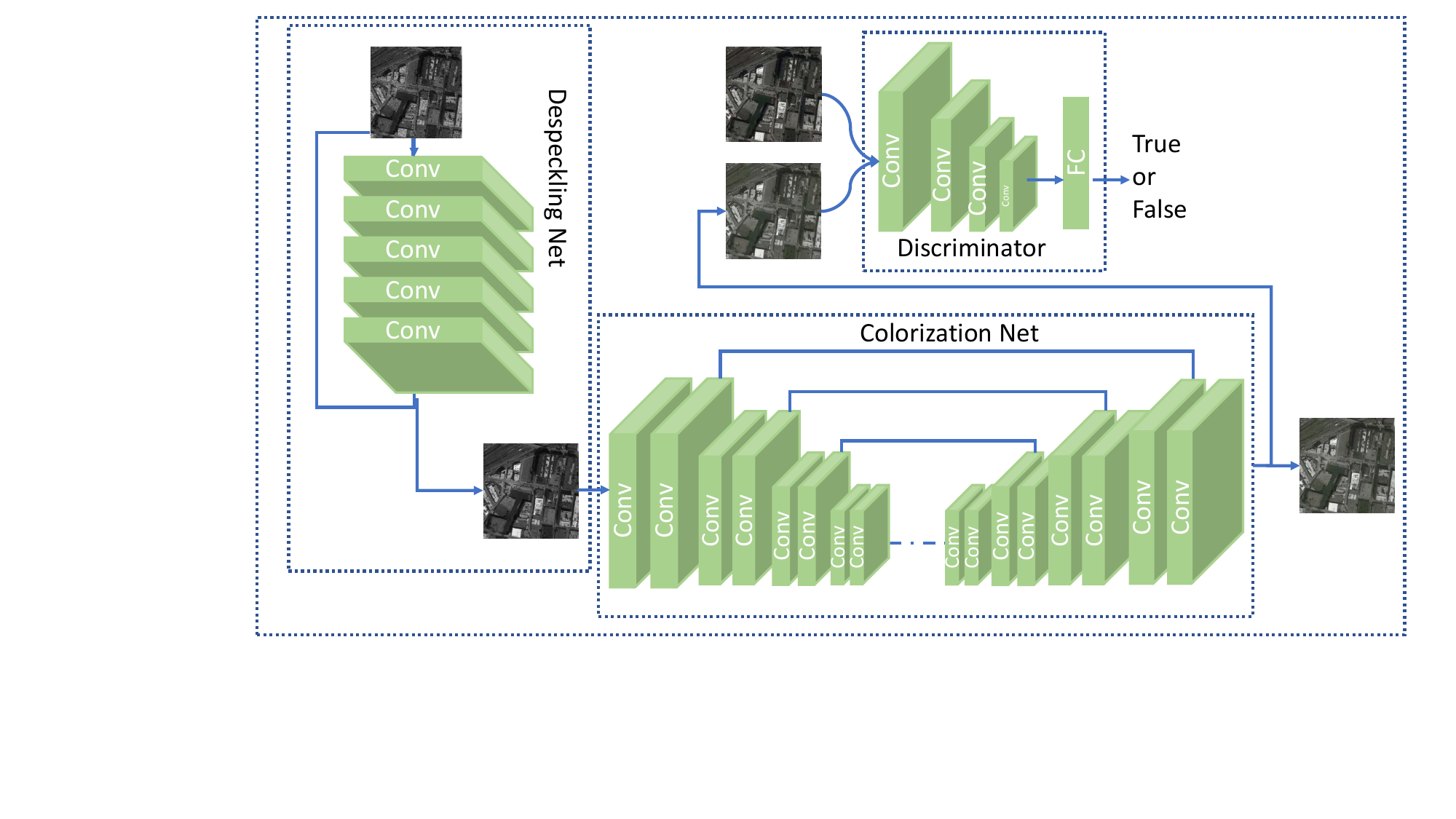}\\
RICNet architecture~\cite{song2017radar} & SAR-GAN architecture~\cite{wang2018generating} \\
\end{tabular}
\end{center}
\caption{\textbf{Domain-Specific Colorization networks} colorize images from different modalities such as infra-red, radar images.}
\label{fig:UnnaturalColorization}
\end{figure}

These networks aim to colorize images from different modalities, such as infrared, or domains, such as radar. Domain-specific colorization models perform better than general-purpose models because they understand the domain-specific subtleties, colors, and patterns. However, these networks have a limited capacity for generalization and frequently show inferior results on images outside of their trained domain. Furthermore, domain-specific colorization has limited usefulness, less applicability on more general images, and needs a significant amount of high-quality domain-specific data for training, which may only occasionally be available. We provide the details of such networks in the following sub-sections and show their architectures in Figure~\ref{fig:UnnaturalColorization}.

\subsubsection{Infrared Colorization}
An automatic Near Infrared (NIR)\footnote{Code is available at \url{https://bit.ly/2YZQhQ4}} image colorization technique~\cite{limmer2016infrared} was developed using a multi-branch deep CNN. NIR~\cite{limmer2016infrared} is capable of learning luminance and chrominance channels. Initially, the input image is preprocessed and converted into a pyramid, where each pyramid level is normalized to zero mean and unit variance. Next, each structurally similar branch of NIR~\cite{limmer2016infrared} is trained using a single input pyramid level without sharing weights between layers. All the branches are merged into a fully connected layer to produce the final output. Additionally, the mean input image is also fed directly to the fully connected layer.

The fundamental structure of NIR is inspired by \cite{zhang2016colorful}. Each branch is composed of stacked convolutional layers with pooling layers placed in the middle after regular intervals. The activation function after each convolutional layer is ReLU. To produce the colorized image, the raw output of the network is joint-bilaterally filtered, and the output is then further enhanced by incorporating high-frequency information directly from the original input image.

Each block has the same number of convolutional layers. The kernel size is 3$\times$3, while a downsampling of 2$\times$2 is performed in the pooling layers. Similarly, the first block employs the same number of feature maps, i.e., 16, but increases the number by a factor of 2 after each pooling layer. Moreover, the authors developed a real-world dataset by capturing road scenes in summer with a multi-CCD NIR/RGB camera. The proposed model is trained on about 32 image pairs and tested on 800 images from the mentioned dataset.

\subsubsection{SAR-GAN}
To colorize Synthetic Aperture Radar (SAR) images, Wang~\etal~\cite{wang2018generating} proposed SAR-GAN. The network uses a cascaded generative adversarial network as its underlying architecture. The input SAR images are first denoised from speckles and then colorized to produce high-quality visible images. The generator of SAR-GAN~\cite{wang2018generating} consists of a despeckling subnet and a colorization subnet. The despeckling subnet produces a noise-free SAR image, which the colorization subnet further processes to produce a colorized image.

The despeckling sub-network is made up of eight convolutional layers, an activation function, an element-wise division residual layer, and batch normalization~\cite{ioffe2015batch}. First, the speckle component in the SAR image is estimated and forwarded to the residual layer. To generate a noise-free image, a residual layer equipped with skip connections performs component-wise division of the input SAR image by the estimated speckle. The colorization sub-network has a symmetric encoder-decoder architecture with three skip connections and eight convolutional layers. This lets the input and output of the network share low-level properties.

The ADAM~\cite{kingma2014adam} optimization technique is adopted for training the entire network. The discriminator component of SAR-GAN~\cite{wang2018generating} uses a hybrid loss function developed by combining the pixel-level $\ell_1$ loss with the adversarial loss. SAR-GAN~\cite{wang2018generating} is tested on 85 out of 3292 synthetic images, as well as real SAR images.

\subsubsection{Radar Image Colorization}
Song~\etal~\cite{song2017radar} developed a feature-extractor network and a feature-translator network for the colorization of single-polarization radar grayscale images\footnote{Code at \url{https://github.com/fudanxu/SAR-Colorization}}. The first seven layers of the VGG16~\cite{simonyan2015vgg} pre-trained on ImageNet~\cite{deng2009imagenet} are used to construct the feature-extractor network. The output of the feature extractor network is a hyper-column descriptor obtained by concatenating the corresponding pixels from all layers with the input image.

The final hyper-column descriptor, thus obtained, is fed into the feature-translator network composed of five fully connected layers. As a final step, the feature-translator network uses the softmax function to obtain an output similar to that of classification, which is achieved by constructing nine groups of neurons representing nine polarimetrics. The ReLU activation function follows each convolutional layer in both networks. Furthermore, full polarimetric radar image patches are employed to train the feature extractor and feature translator.

\subsubsection{Sketch Image Colorization}
Lee~\etal~\cite{lee2020CVPR} proposed a reference-based sketch image colorization, where a sketch image is colorized according to an already-colored reference image. In contrast to grayscale images, which contain pixel intensity, sketch images are more information-scarce, which makes sketch image colorization more challenging.

In the training phase, the authors proposed an augmented self-reference generation method generated from the original image by color perturbation and geometric distortion. Specifically, an outline extractor first converts a color image into its sketch image. Then, the augmented self-reference image is generated by applying the thin plate splines transformation to the reference image. As the ground truth, the generated reference image contains most of the original image's content, thus providing full correspondence information for the sketch. Taking the sketch image and reference image as inputs, the network first encodes these two inputs by two independent encoders.
Furthermore, the authors proposed a spatially corresponding feature transfer module to transfer the contextual representations obtained from the reference into the sketch's spatially corresponding positions (i.e., integrating the sketch's feature representations and its reference image). The integrated features are passed through residual blocks and a decoder to produce the colored output. During training, a similarity-based triplet loss, $\ell_{1}$ reconstruction loss, adversarial loss, perceptual loss, and style loss are used to drive the learning of the proposed network.

The network has a U-net-like structure coupled with several residual blocks and a spatially corresponding feature transfer module between the encoder and decoder. The sketch and the color reference image with a size of 256 $\times$ 256 are fed to the network, which colorizes the sketch image.

\subsection{Text-based Colorization}
\begin{figure}[t]
\begin{center}
\begin{tabular}{cc} 
\includegraphics[trim={7.5cm 10cm  5cm  -2cm },clip,width=0.5\columnwidth,valign=t]{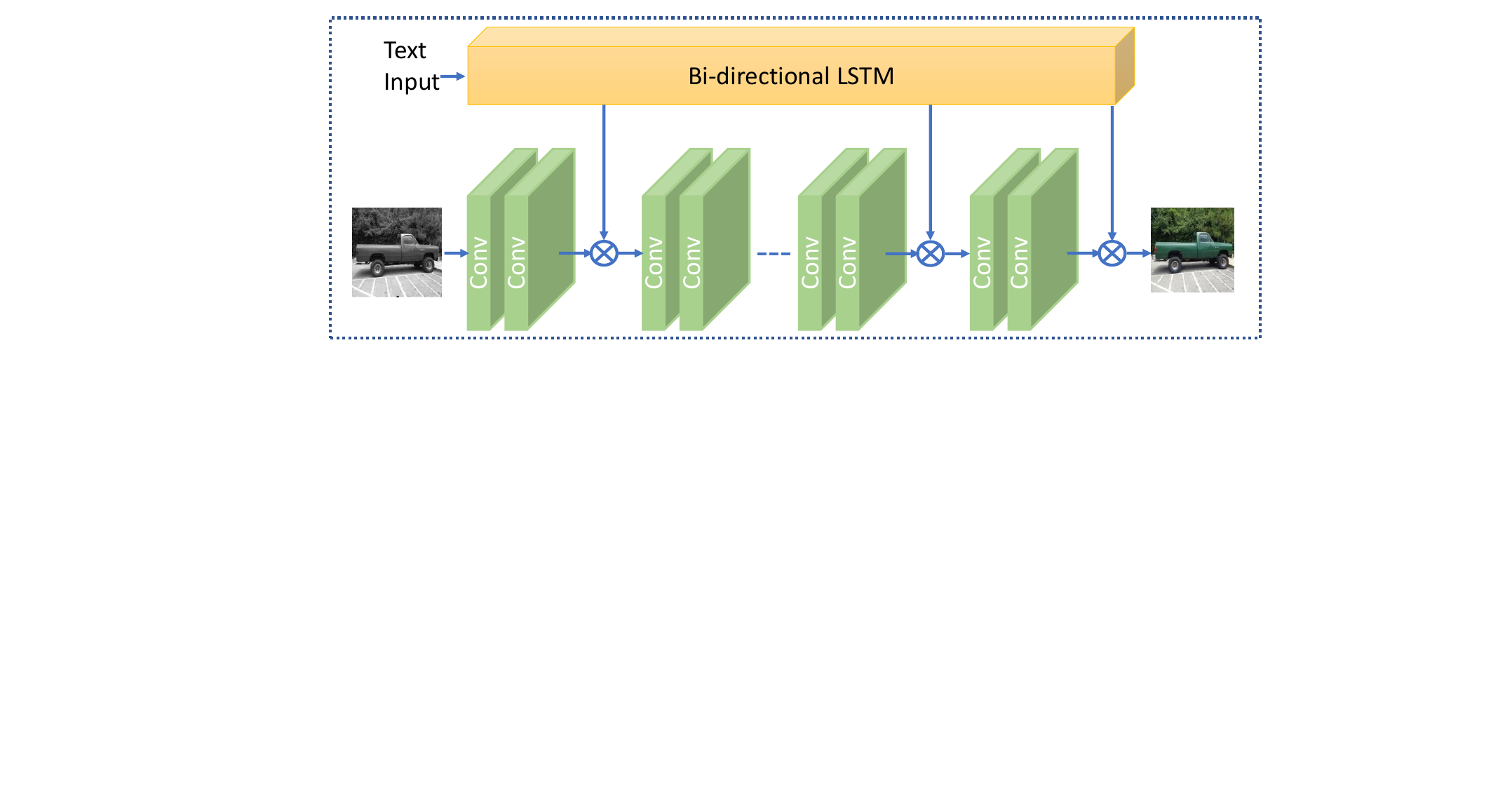}&

\includegraphics[trim={7.5cm 4cm  5cm  1.5cm },clip,width=0.5\columnwidth,valign=t]{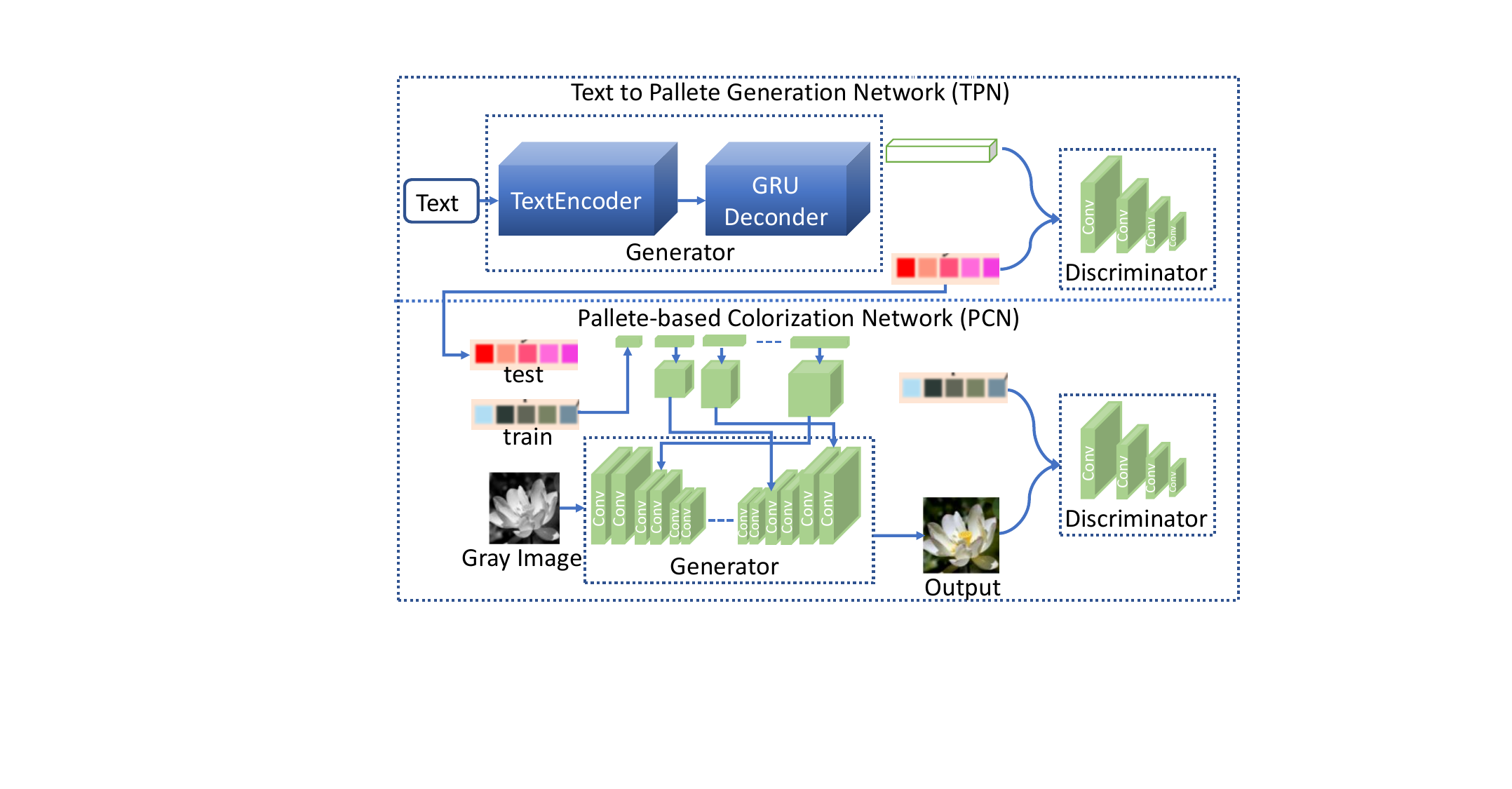}\\
LCL-Net architecture~\cite{manjunatha2018learning} & Text2Color~\cite{bahng2018coloring}\\
\end{tabular}
\end{center}
\caption{\textbf{Text-based colorization networks} are based on the text input with the grayscale image.}
\label{fig:TextBasedColorization}
\end{figure}

These types of networks colorize the images based on an additional input, usually text. Users can specify distinct color schemes, thematic components, and in-depth color preferences. However, instructions written in complex terminologies may hinder the network performance, leading to inconsistent outcomes. Hence, the model may require more precise descriptions; otherwise, the output reliability may decrease. We classify the following models as text-based colorization networks. Figure~\ref{fig:TextBasedColorization} presents the network architectures for this category.

\subsubsection{Learning to Color from Language}
To exploit additional text input for colorization, a language-conditioned colorization architectures\footnote{Available at \url{https://github.com/superhans/colorfromlanguage}}~\cite{manjunatha2018learning} was proposed to colorize grayscale images with additional learning from image captions. The authors employed an existing language-agnostic architecture called FCNN~\cite{zhang2016colorful}, providing image captions as an additional input. FCNN~\cite{zhang2016colorful} comprises eight blocks, each consisting of a sequence of convolutional layers followed by batch normalization. The authors experimented with the \emph{CONCAT} network~\cite{reed2016learning} to generate captions from images but settled on the \emph{FILM} network~\cite{perez2018film} due to its small number of parameters. To obtain the final output, the language-conditioned weights of the FILM~\cite{perez2018film} are utilized for affine transformation of the convolutional blocks' outputs.

Finally, the FCNN~\cite{zhang2016colorful} and the two language-conditioned architectures, i.e., CONCAT and FILM, were trained on images from the MS-COCO dataset~\cite{lin2014coco}. Automatic evaluations indicated that the FILM architecture achieves the highest accuracy. Moreover, crowd-sourced assessments validated the effectiveness of the models. The network's output is a 56$\times$56 color image.

\subsubsection{Text2Colors}
The Text2Colors\footnote{Code is \url{https://github.com/awesome-davian/Text2Colors/}}~\cite{bahng2018coloring} model is comprised of two conditional generative adversarial networks: Text-to-Pallette Generation Network (TPN) and Palette-based Colorization Network (PCN). TPN is responsible for constructing color palettes learned from the Palette-and-Text (PAT) dataset, which contains five color palettes for each of 10,183 textual phrases. PCN is responsible for colorizing the input grayscale image given the generated palette based on the input text.

The TPN generator learns the pairing between text and color palette, while its discriminator simultaneously learns the color palette and text characteristics to identify the real palettes from the fake ones. The Huber is the loss function in the TPN network.

On the other hand, the generator of the PCN network has two subnetworks: the colorization network based on U-Net~\cite{ronneberger2015u} to colorize the images and the conditioning network to apply the palette colors to the generated image. The PCN discriminator is based on the DCGAN architecture~\cite{radford2015unsupervised}. In the PCN's discriminator, first, the features from the input image and generated palette (by the TPN network) are jointly learned by a series of Conv-LeakyReLU layers. Then, a fully connected layer classifies the image as real or fake.

The discriminator and generator of TPN are first trained on the PAT dataset for 500 epochs and then trained on the ground truth image for 100 epochs. The trained generators of TPN and PCN are then used to colorize the input grayscale image at the testing stage using the input text's color palette. Adam optimizer is used with a learning rate of 0.0002 for all the networks.

\subsection{Diverse Colorization}
\begin{figure}[t]
\begin{center}
\begin{tabular}{cc} 
\includegraphics[trim={5cm 10cm  5cm  -2cm},clip,width=0.5\columnwidth,valign=t]{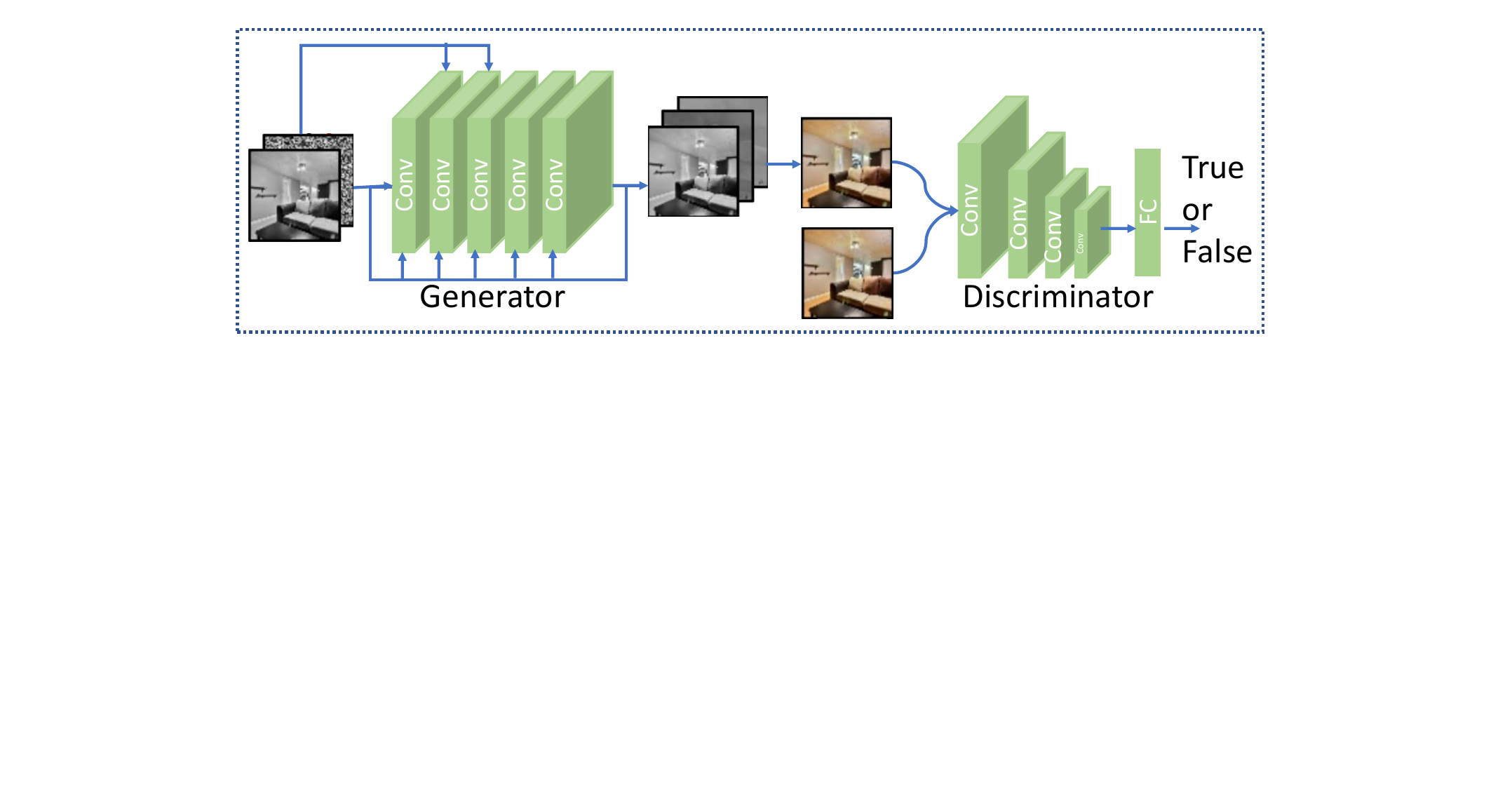}&
\includegraphics[trim={8.5cm 9cm  4cm  0cm},clip,width=0.5\columnwidth,valign=t]{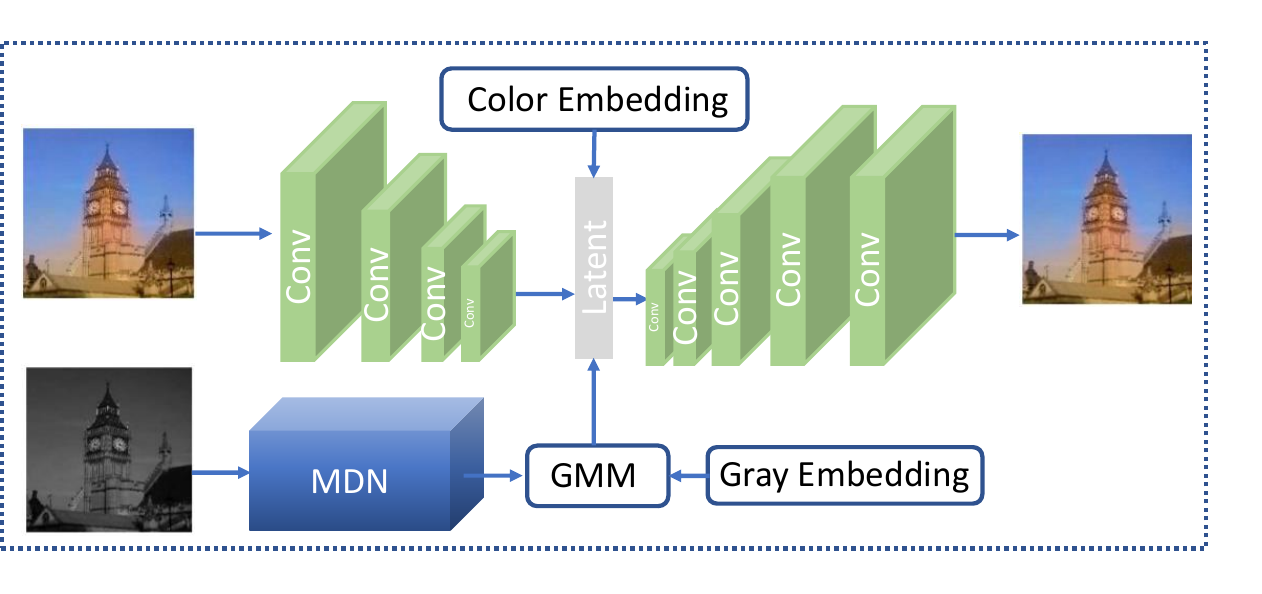}\\
UDCGAN~\cite{cao2017unsupervised}/ICGAN~\cite{nazeri2018image} & Diverse Image Colorization~\cite{deshpande2017learning}\\
\includegraphics[trim={0.5cm 0.5cm  1cm  0cm },clip,width=0.5\columnwidth,valign=t]{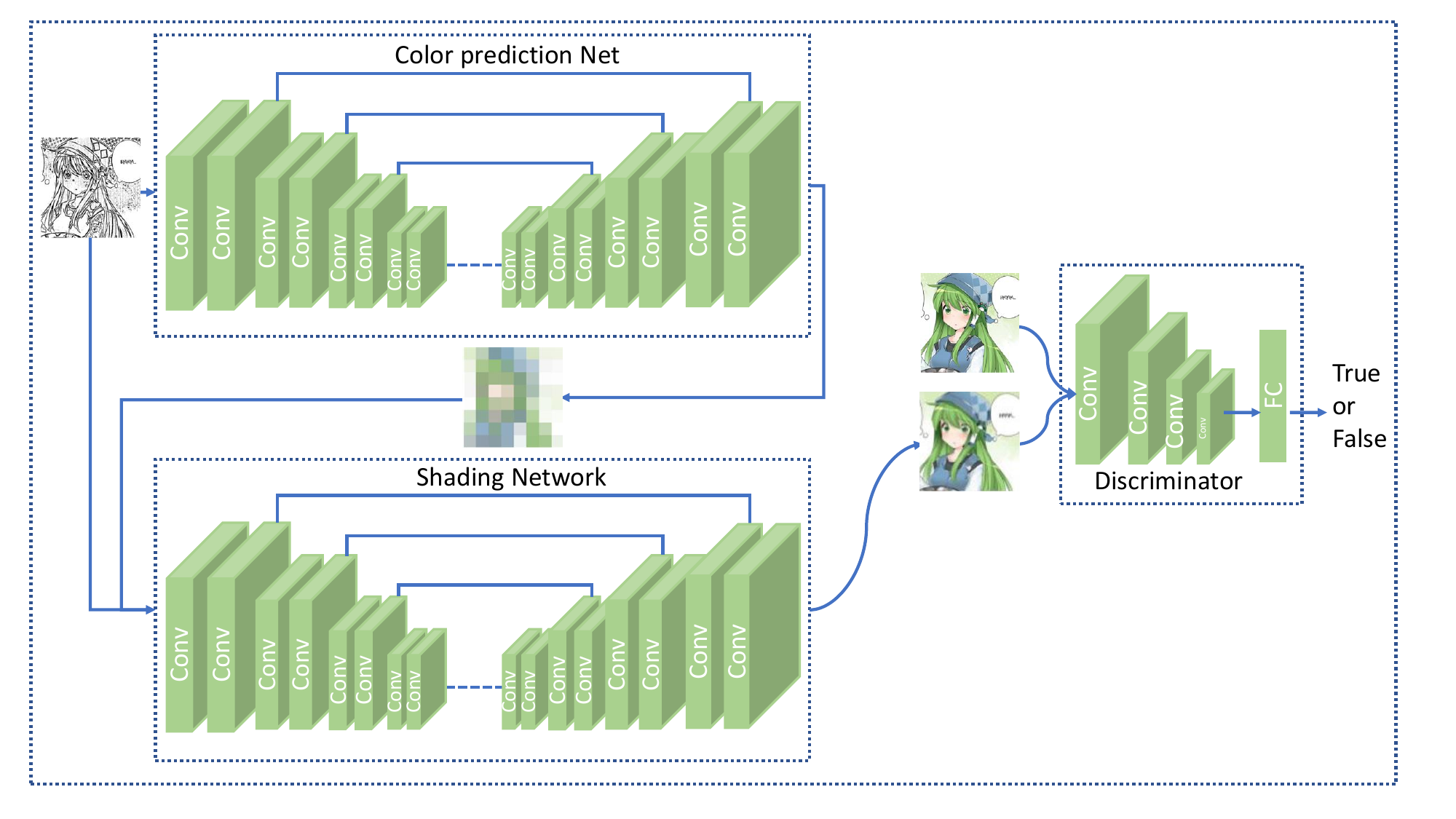}& 
\includegraphics[trim={7.5cm 6cm  8.5cm  -1cm },clip,width=0.5\columnwidth,valign=t]{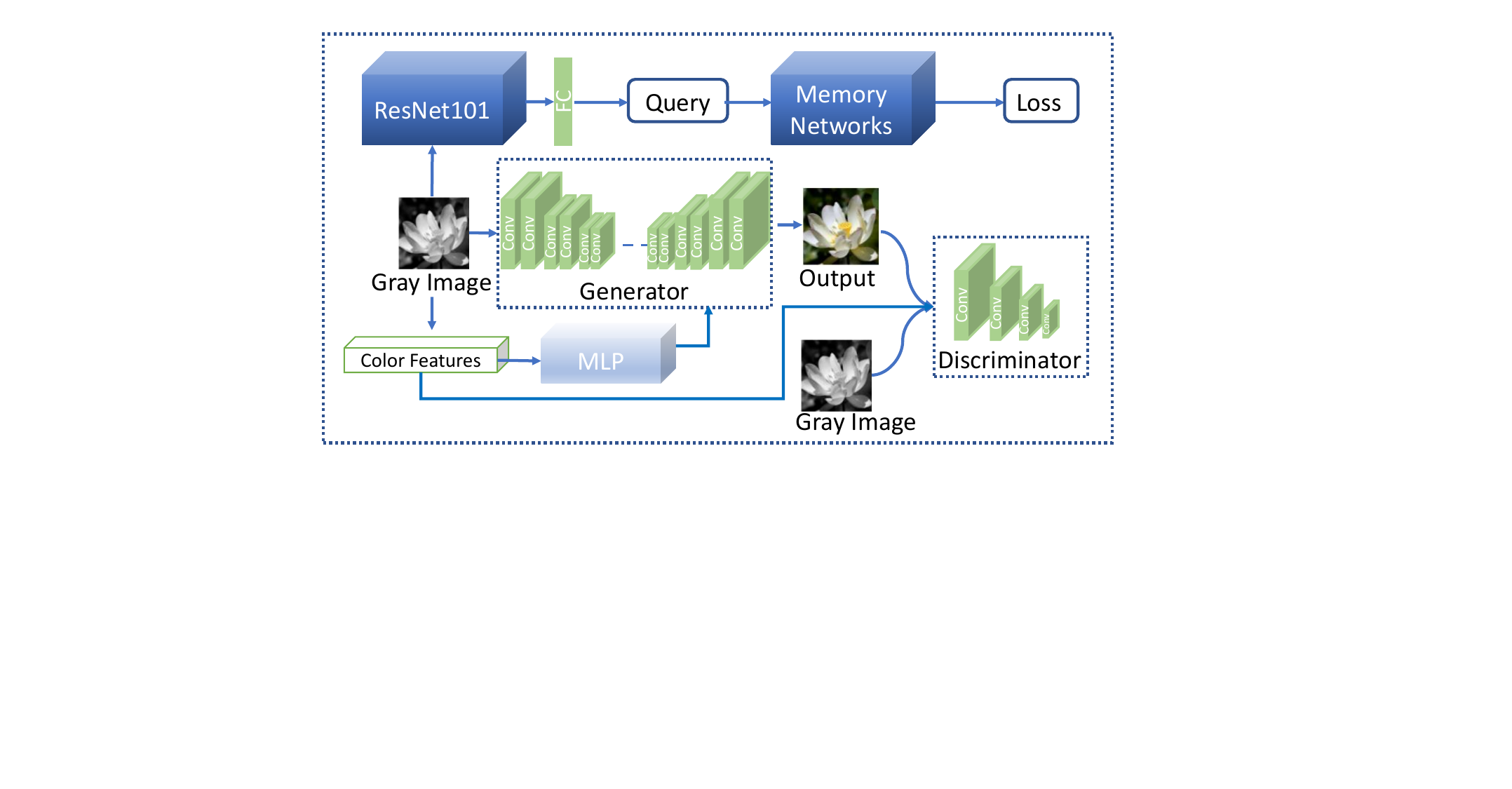}\\
TANDEM architecture~\cite{frans2017outline} & Memopainter~\cite{yoo2019coloring}\\
\multicolumn{2}{c}{\includegraphics[trim={0.5cm 8.5cm  2cm  0cm },clip,width=0.5\columnwidth,valign=t]{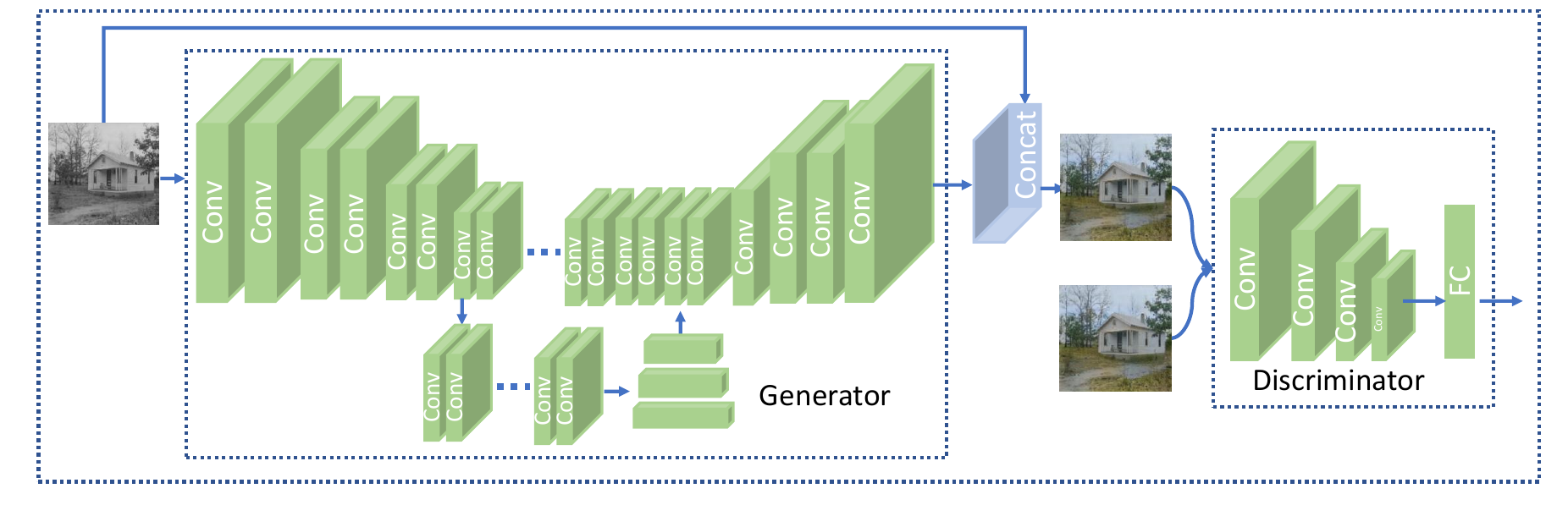}}\\
\multicolumn{2}{c}{ChromaGAN~\cite{vitoria2020chromagan}}\\

\end{tabular}
\end{center}
\caption{\textbf{Diverse colorization networks} generate different colorized images instead of aiming to restore the original color only. }
\label{fig:DiverseColorization}
\end{figure}

Diverse colorization aims to generate different colorized images rather than restore the original color, as shown in Figure~\ref{fig:DiverseColorization}. Diverse colorization is usually achieved via GANs or variational autoencoders. GANs attempt to generate the colors in a competitive manner, where the generator aims to fool the discriminator while the discriminator's goal is to differentiate between the ground truth and the generated colors. These models are useful for applications where diverse colors are possible and help explore more vibrant colors for a specific scene or item. However, they fail when the item's color is consistent. The output diversity may introduce ambiguity or inconsistency, and selecting the most appropriate colorization is challenging. Furthermore, the computational time and resource consumption also increase. We present the GANs employed for colorization in the following paragraphs.

\subsubsection{Unsupervised Diverse Colorization}
Cao~\etal~\cite{cao2017unsupervised} proposed the use of conditional GANs for the diverse colorization of real-world objects\footnote{Code is available at https://github.com/ccyyatnet/COLORGAN}. In the generator of the GAN network, the authors employed five fully convolutional layers with batch normalization and ReLU. Every generator layer concatenates the grayscale image to provide continuous conditional supervision for realistic results. Furthermore, noise channels are added to the first three convolutional layers of the generator network to diversify the colorization outputs. Four convolutional layers and a fully connected layer make up the discriminator, distinguishing between the image's fake and real values.

During the convolution operations, the stride is set to $1$ to keep the spatial information the same across all layers. The performance of the proposed method was assessed using the Turing test methodology. The authors provided questionnaire surveys to 80 subjects, asking them 20 questions regarding the results produced for the publicly available LSUN bedroom dataset~\cite{yu2015lsun}. The proposed model obtained a convincing rate of $62.6\%$ compared to $70\%$ for ground truth images. Additionally, a significance t-test generated a p-value of 0.1359, indicating that the generated colorized images are not significantly different from the real images.

For implementation, the authors opted for the TensorFlow framework. The authors selected a batch size of 64 and a learning rate of 0.0002 and 0.0001 for the discriminator and generator, respectively. The model was trained for 100 epochs using RMSProp as an optimizer. The network's output is 64$\times$64 in size.

\subsubsection{Tandem Adversarial Networks}
For the colorization of raw line art, Frans~\cite{frans2017outline} proposed two adversarial networks in tandem. The first network, namely the color-prediction network, predicts the color scheme from the outline, while the second network, called the shading network, produces the final image from the color scheme generated by the color-prediction network in conjunction with the outline. The color-prediction and shading networks have the same structure as U-Net~\cite{ronneberger2015u} while the discriminator has the same structure as ~\cite{cao2017unsupervised}. The adversarial training uses the discriminator formed by stacking the four convolutional layers and a fully connected layer at the end.

The convolutional layers are transposed once the density of the feature matrix reaches a certain level. The author also added skip connections between the corresponding layers to allow the gradient to flow through the network directly. Further, $\ell_2$ and adversarial losses are incorporated in the color-prediction and shading-networks, respectively. The convolutional filter size is set to $5\times5$ with a stride of two for every convolutional layer in the network. Starting from an initial 64 feature maps, the number of feature maps is increased twice after every layer.

\subsubsection{ICGAN} 
Image Colorization Using Generative Adversarial Networks abbreviated as ICGAN~\cite{nazeri2018image} proposed by Nazeri~\etal, has demonstrated better performance in image colorization tasks than traditional convolutional neural networks. Fully convolutional networks for semantic segmentation~\cite{long2015fully} inspire the baseline model and are constructed by replacing the fully connected layers of the network with fully convolutional layers. The basic idea is to downsample the input image gradually via multiple contractive encoding layers and then apply a reverse operation to reconstruct the output with many expansive decoding layers, similar to U-Net~\cite{ronneberger2015u}.

The ICGAN~\cite{nazeri2018image} generator takes grayscale input image instead of random noise, like traditional GANs. The authors also proposed the generator's modified cost function to maximize the discriminator's probability of being incorrect instead of minimizing the likelihood of being correct. Moreover, the baseline model's generator is used without any modifications, while the discriminator comprises a series of convolutional layers with batch normalization~\cite{ioffe2015batch}.

The filter size in each convolutional layer is 4$\times$4 as opposed to the traditional 3$\times$3, and the activation function is Leaky-ReLU~\cite{maas2013rectifier} with a slope value of 0.2. The final one-dimensional output is obtained by applying a convolution operation after the last layer, predicting the input's nature with a certain probability. 

To train the network, the authors employed ADAM optimization~\cite{kingma2014adam} with weight initialization using the guidelines from~\cite{he2015delving}. The system's performance is assessed using CIFAR10~\cite{krizhevsky2009learning} and Places356 datasets~\cite{zhou2016places}. Overall, the visual performance of the ICGAN~\cite{nazeri2018image} is favorable compared to traditional CNNs.

\begin{table*}[t]
\caption{The colorization algorithms summarization in terms of strength and weakness.}
\centering
\resizebox{\textwidth}{!}{
\begin{tabular}{|cllll|} 
\hline
\multicolumn{1}{|l}{Category}                     & \multicolumn{1}{c}{Methods}                                                                                                                                                                                                           & \multicolumn{1}{c}{References}                                                                             & \multicolumn{1}{c}{Strengths}                                                                                                                                                                                       & \multicolumn{1}{c|}{Weakness}                                                                                                                                                                                                                                           \\ 
\hline \hline
 
\rotatebox[origin=c]{90}{Plain}                & \begin{tabular}{@{\labelitemi\hspace{\dimexpr\labelsep+0.5\tabcolsep}}l@{}}Deep Colorization\\Colorful Colorization\\Deep Depth Colorization\\U-Net Grayscale Image Colorization\end{tabular}                                                                              & \begin{tabular}{@{\labelitemi\hspace{\dimexpr\labelsep+0.5\tabcolsep}}l@{}}\cite{cheng2015deep}\\\cite{zhang2016colorful}\\\cite{carlucci20182}\\\cite{hu2024grayscale-new2}\end{tabular}             & \begin{tabular}{@{\labelitemi\hspace{\dimexpr\labelsep+0.5\tabcolsep}}l@{}}Simple architecture\\Easy to implement\end{tabular}                                                                                       & \begin{tabular}{@{\labelitemi\hspace{\dimexpr\labelsep+0.5\tabcolsep}}l@{}}Limited performance\\Gradient vanish problem\\Create deeper networks challenging\end{tabular}                                                                           \\ \\ \\

\rotatebox[origin=c]{90}{User-Guided}         & \begin{tabular}{@{\labelitemi\hspace{\dimexpr\labelsep+0.5\tabcolsep}}l@{}}Scribbler\\Real-Time User-Guided\\Interactive Deep Colorization\\Anime Line Art Colorization\end{tabular}                                                   & \begin{tabular}{@{\labelitemi\hspace{\dimexpr\labelsep+0.5\tabcolsep}}l@{}}\cite{sangkloy2017scribbler}\\\cite{zhang2017real}\\\cite{xiao2019interactive}\\\cite{ci2018user}\end{tabular}          & \begin{tabular}{@{\labelitemi\hspace{\dimexpr\labelsep+0.5\tabcolsep}}l@{}}User choice incorporated\\Diverse colorization possible\end{tabular}                                                                      & \begin{tabular}{@{\labelitemi\hspace{\dimexpr\labelsep+0.5\tabcolsep}}l@{}}Colors are subject user preference \\Selected colors may not be natural~\\Requires human intervension\end{tabular}                                                                          \\ \\ \\

\rotatebox[origin=c]{90}{Domain-Specific}      & \begin{tabular}{@{\labelitemi\hspace{\dimexpr\labelsep+0.5\tabcolsep}}l@{}}Infrared Colorization\\SAR-GAN\\Radar Image Colorization\\Sketch Image Colorization\end{tabular}                                                            & \begin{tabular}{@{\labelitemi\hspace{\dimexpr\labelsep+0.5\tabcolsep}}l@{}}\cite{limmer2016infrared}\\\cite{wang2018generating}\\\cite{song2017radar}\\\cite{lee2020CVPR}\end{tabular}          & \begin{tabular}{@{\labelitemi\hspace{\dimexpr\labelsep+0.5\tabcolsep}}l@{}}Domain knowlege incorporated\\Faithful to the domain\end{tabular}                                                                         & \begin{tabular}{@{\labelitemi\hspace{\dimexpr\labelsep+0.5\tabcolsep}}l@{}}Limited applications\\Usually not Generalizable\end{tabular}                                                                                                          \\ \\ \\
\rotatebox[origin=c]{90}{Text-Based}          & \begin{tabular}{@{\labelitemi\hspace{\dimexpr\labelsep+0.5\tabcolsep}}l@{}}Learning to Color from Language\\Text2Colors\end{tabular}                                                                                                   & \begin{tabular}{@{\labelitemi\hspace{\dimexpr\labelsep+0.5\tabcolsep}}l@{}}\cite{manjunatha2018learning}\\\cite{bahng2018coloring}\end{tabular}                & \begin{tabular}{@{\labelitemi\hspace{\dimexpr\labelsep+0.5\tabcolsep}}l@{}}Combines text and visual information\end{tabular}                                               & \begin{tabular}{@{\labelitemi\hspace{\dimexpr\labelsep+0.5\tabcolsep}}l@{}}Requires text input\\Requires accurate text\\Requires multi-modal data for training\\Text may not be available always\end{tabular}    \\ \\ \\

\rotatebox[origin=c]{90}{Diverse Colorization}& \begin{tabular}{@{\labelitemi\hspace{\dimexpr\labelsep+0.5\tabcolsep}}l@{}}Unsupervised diverse colorization\\Tandem adversarial networks\\ICGAN\\Learning diverse colorization\\ChromaGAN\\Memopainter\\CycleGAN\\Transforming Color\end{tabular}                   & \begin{tabular}{@{\labelitemi\hspace{\dimexpr\labelsep+0.5\tabcolsep}}l@{}}\cite{cao2017unsupervised}\\\cite{frans2017outline}\\\cite{nazeri2018image}\\\cite{deshpande2017learning}\\\cite{vitoria2020chromagan}\\\cite{yoo2019coloring}\\\cite{li2023image-new3}\\\cite{shafiq2024transforming-new4}\end{tabular}    & \begin{tabular}{@{\labelitemi\hspace{\dimexpr\labelsep+0.5\tabcolsep}}l@{}}Produce different colorized images\\Can be trained unsupervised\end{tabular}                                                              & \begin{tabular}{@{\labelitemi\hspace{\dimexpr\labelsep+0.5\tabcolsep}}l@{}}Difficult to obtain original colors\\Dependent on the training data\\Fails to distinguish similar local texture regions\\Challenging to quantify the results\end{tabular}  \\ \\ \\

\rotatebox[origin=c]{90}{Multi-Path}           & \begin{tabular}{@{\labelitemi\hspace{\dimexpr\labelsep+0.5\tabcolsep}}l@{}}Let there be color\\Automatic colorization\\PixColor\\ColorCapsNet\\Pixelated\\Multiple Hypothesis Colorization\\Pixel-Level Semantics Guided\\Shadow-Aware Image Colorization~\end{tabular} & \begin{tabular}{@{\labelitemi\hspace{\dimexpr\labelsep+0.5\tabcolsep}}l@{}}\cite{iizuka2016let}\\\cite{larsson2016learning}\\\cite{guadarrama2017pixcolor}\\\cite{ozbulak2019image}\\\cite{zhao2019pixelated}\\\cite{baig2017multiple}\\\cite{Zhao0SH018}\\\cite{duan2024shadow-new1}\end{tabular} & \begin{tabular}[c]{@{}l@{}}\begin{tabular}{@{\labelitemi\hspace{\dimexpr\labelsep+0.5\tabcolsep}}l@{}}Solves vanishing gradient problem \\ Employed skip connections\\Easy flow of information\end{tabular}\\\\\end{tabular} & \begin{tabular}{@{\labelitemi\hspace{\dimexpr\labelsep+0.5\tabcolsep}}l@{}}May have complex architecture\\Computationally expensive\end{tabular} \\\\ \\

\rotatebox[origin=c]{90}{Exemplar-Based}      & \begin{tabular}{@{\labelitemi\hspace{\dimexpr\labelsep+0.5\tabcolsep}}l@{}}Deep Exemplar Colorization\\Instance-Aware Colorization\\Fast Deep Exemplar Colorization\end{tabular}                                                       & \begin{tabular}{@{\labelitemi\hspace{\dimexpr\labelsep+0.5\tabcolsep}}l@{}}\cite{he2018deep}\\\cite{su2020CVPR}\\\cite{xu2020CVPR}\end{tabular}             & \begin{tabular}{@{\labelitemi\hspace{\dimexpr\labelsep+0.5\tabcolsep}}l@{}}Guided via Exemplar images\\Abundant examples available on internet\end{tabular}                                                           & \begin{tabular}{@{\labelitemi\hspace{\dimexpr\labelsep+0.5\tabcolsep}}l@{}}Results dependent on exemplar images\\Examples may lack similar colors\\Difficult deal with unusual colors\\ Challenging artistic colors\end{tabular}              \\ \hline
\end{tabular}
}
\label{tab:Strenght_weakness}
\end{table*}

\subsubsection{Learning Diverse Image Colorization}
Deshpande~\etal~\cite{deshpande2017learning} employed Variational Auto-Encoder (VAE) and Mixture Density Network (MDN) to yield multiple diverse yet realistic colorizations for a single grayscale image\footnote{Code is available at \url{https://github.com/aditya12agd5/divcolor}}. The authors first employed VAE to learn a low-dimensional embedding for a color field of size $64\times 64 \times 2$, and then used MDN to generate multiple colorizations.

The VAE architecture encoder consists of four convolutional layers with a kernel size of $5\times 5$ and a stride of two. The feature channels start at 128 and double in the successive encoder layer. Batch normalization follows each convolutional layer, and ReLU serves as the activation function. The last layer of the encoder is fully connected. On the other hand, the decoder of the VAE architecture has five convolutional layers, each preceded by linear upsampling, batch normalization, and ReLU. The input to the decoder is a d-dimensional embedding, and the output is a $64\times 64 \times 2$ color field. The convolutional kernel size is $4\times 4$ in the first layer, and in the remaining, it is of size $5\times5$. The activation function in all layers is ReLU, except in the last layer, where the activation function is \emph{TanH}. The decoder halves the output channels at each layer.

MDN consists of twelve convolutional layers and two fully connected layers and is activated by the ReLU function, which is followed by batch normalization. During testing, embeddings sampled from the MDN output can be used by the decoder of the VAE to produce multiple colorizations. To enhance the overall performance of the proposed architecture, the authors designed three loss functions: specificity, colorfulness, and gradient.

\subsubsection{ChromaGAN}
ChromaGAN\footnote{Code available at \url{https://github.com/pvitoria/ChromaGAN}}~\cite{vitoria2020chromagan} exploits geometric, perceptual, and semantic features via an end-to-end self-supervised generative adversarial network. The proposed model can colorize the image realistically (actual colors) rather than merely focusing on aesthetic appeal by utilizing the semantic understanding of the real scenes.

The generator of ChromaGAN~\cite{vitoria2020chromagan} is composed of two branches, which receive a grayscale image of size 224$\times$224 as input. One of the branches yields chrominance information, whereas the other generates a class distribution vector. The network composition is as follows: the first stage is shared between branches that implement VGG16~\cite{simonyan2015vgg} while removing the last three fully connected layers. The pre-trained VGG16~\cite{simonyan2015vgg} weights are used to train this stage without freezing them. In the second stage, each branch follows its dedicated path. The first branch has two modules, each designed by combining a convolutional layer followed by batch normalization~\cite{ioffe2015batch} and ReLU. The second path has four modules with the same composite structure (Conv-BN-ReLU) as the first branch but is further followed by three fully connected layers and provides the class distribution. The third stage fuses the outputs of these two distinct paths. The features then pass through six modules, each with a convolutional layer and ReLU with two upsampling layers.

The discriminator is based on PatchGAN~\cite{isola2017PatchGAN} architecture, which holds high-frequency structural information about the generator's output by focusing on the local patches rather than the entire image. For five epochs, ChromaGAN~\cite{vitoria2020chromagan} is trained on 1.3M images of ImageNet~\cite{deng2009imagenet}. The optimizer used is ADAM, which has an initial learning rate of 2$e^{-5}$.

\subsubsection{MemoPainter: Coloring with Limited Data}
MemoPainter\footnote{\url{https://github.com/dongheehand/MemoPainter-PyTorch}}~\cite{yoo2019coloring} is capable of learning from limited data, which is made possible by the integration of external memory networks~\cite{kaiser2017MemoryNet} with the colorization networks. This technique effectively avoids the dominant color effect and preserves the color identity of different objects.

Memory networks keep the history of rare examples, enabling them to perform well even with insufficient data. To train memory networks in an unsupervised manner, a novel threshold triplet loss was introduced by the authors~\cite{yoo2019coloring}. In memory networks, the \enquote{key-memory} holds spatial information and computes cosine similarity against the input. Likewise, \enquote{value-memory} keeps a record of color information utilized by the colorization networks as a condition. Similarly, \enquote{age} keeps the time-stamp for each entry in the memory. The \enquote{key} and \enquote{value} memories are generated from the training data. The memory is updated by averaging the \enquote{key-memory} and a new query image owing to the distance between the color information of the new query image and the available top-1 \enquote{value-memory} element that lies below the threshold. Otherwise, a new record is stored for the input image color information.

The colorization networks are constructed using conditional GANs~\cite{mirza2014cgan} consisting of a generator and discriminator, and color information is learned from RGB images during training. The generator inspired by U-Net~\cite{ronneberger2015u} is composed of ten convolutional layers, while the discriminator is fully convolutional, consisting of four layers. The color feature is extracted and provided to the generator after being passed through the MLP.

During testing, the color information is retrieved from the memory networks and fed to the generator as a condition. The colorization networks employ adaptive instance normalization for enhanced colorization, considered a style transfer. The performance of MemoPainter~\cite{yoo2019coloring} is evaluated on different datasets, including Oxford102 Flower~\cite{nilsback2008flower}, Monster~\cite{Monsters2001}, Yumi~\footnote{https://comic.naver.com/webtoon/list.nhn?titleId=651673}, Superheroes~\cite{yoo2019coloring}, and Pokemon~\footnote{https://www.kaggle.com/kvpratama/pokemon-images-dataset}.

\subsubsection{CycleGAN based Image Colorization}
Li~\etal~\cite{li2023image-new3} developed SS-CycleGAN, an automatic image colorization method that considers high-level and detailed semantics and spatial information of the objects in the image. SS-CycleGAN consists of two generators and two discriminators. One of the generators is used for translating grayscale images to color ones, and the other is used for translating color images to grayscale ones. Similarly, the two discriminators are separately used to discriminate the generated color and grayscale images and maintain the rationality of the produced color and grayscale images.

The authors implemented a patch discriminator for SS-CycleGAN built on a self-attention mechanism, which can administer the patch discriminator to concentrate on spatial structure information and the semantic relevance of colored objects in an image. The authors integrate the detail loss term with the joint loss function of SS-CycleGAN to ensure the details of the generated image remain consistent with the original image. Moreover, they further implemented a multi-scale cascaded dilated convolution module in the generators of SS-CycleGAN for the extraction of multi-scale features of local image patches to obtain spatial information on different colored objects.

\subsubsection{Transforming Color}
Shafiq and Lee \cite{shafiq2024transforming-new4} proposed a novel technique for image colorization that integrates a VGG-based global feature extraction, a color encoder, a color transformer, and a GAN architecture. The transformer architecture extracts global information from the input image, and the GAN framework improves the image's overall subjective quality. After converting the RGB image into a Lab color space image, the input image is processed by a pre-trained VGG network, and the encoder captures high-level global semantic features. The output features of the VGG network are integrated with the encoder layers. Simultaneously, a color encoder employs convolutional layers to generate color information from a normal distribution. The fusion of color-encoded features at the bottleneck in the color transformer block and the global features in the encoder layers led to the integration of global and color-specific information. The Swin Transformer block captures the long-range dependencies and spatial relationships by processing the fused features. The proposed method can capture global features using two transformer blocks. Finally, the GAN's generator is composed of an encoder, a color transformer, a color decoder, and a discriminator. A combination of loss functions, including perceptual loss, adversarial loss, and color loss, achieves enhanced colorization during the training phase. The proposed architecture analyzes local image regions, enabling the proposed method to capture fine details of the entire image.

\subsection{Multi-path Colorization Networks}
\begin{figure}[htbp]
\begin{center}
\begin{tabular}{cc} 

\includegraphics[trim={0cm 3.5cm  0cm  1cm },clip,width=0.5\columnwidth,valign=t]{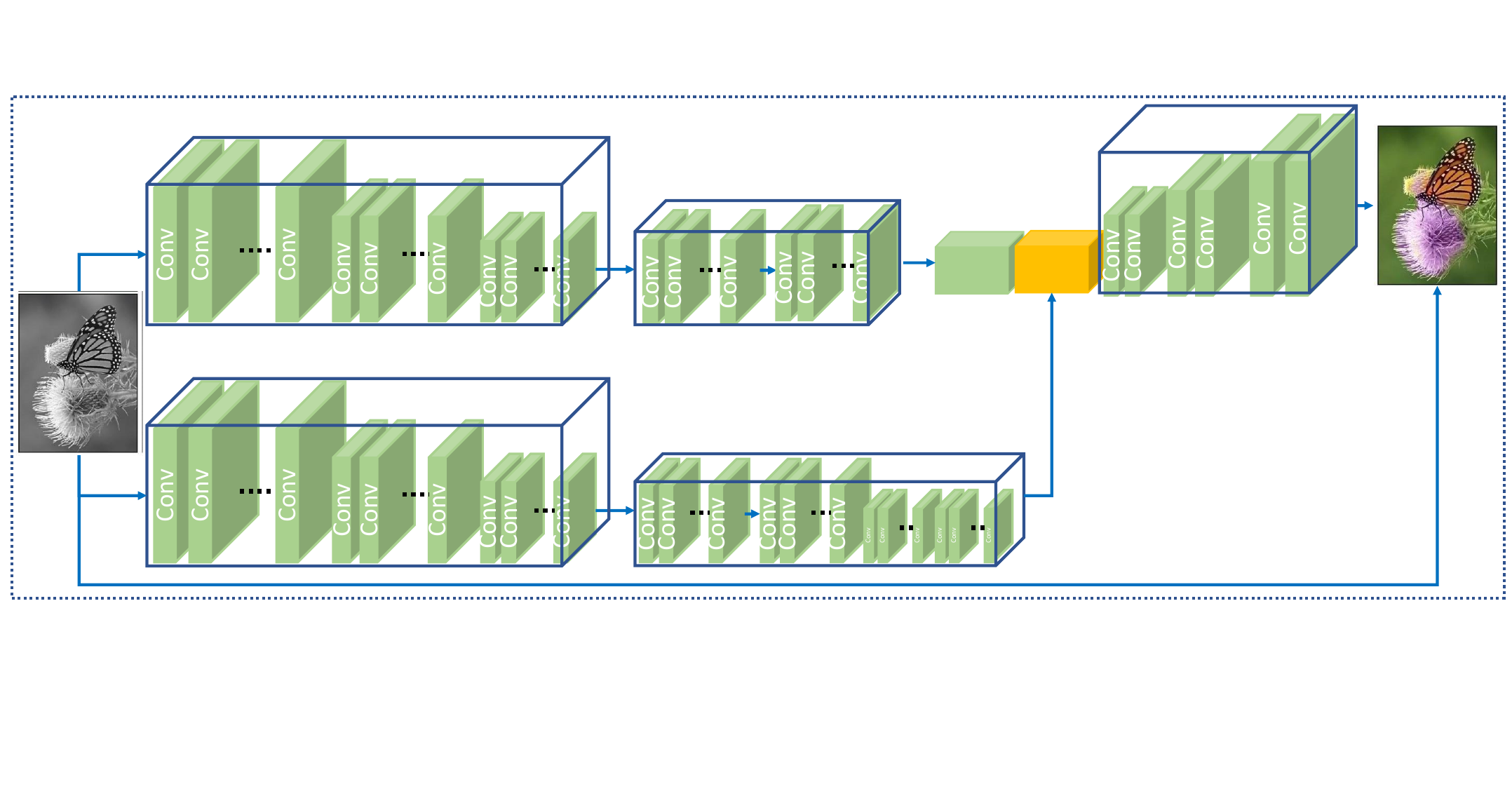}&
\includegraphics[trim={3cm 5.6cm  2cm  0cm },clip,width=0.5\columnwidth,valign=t]{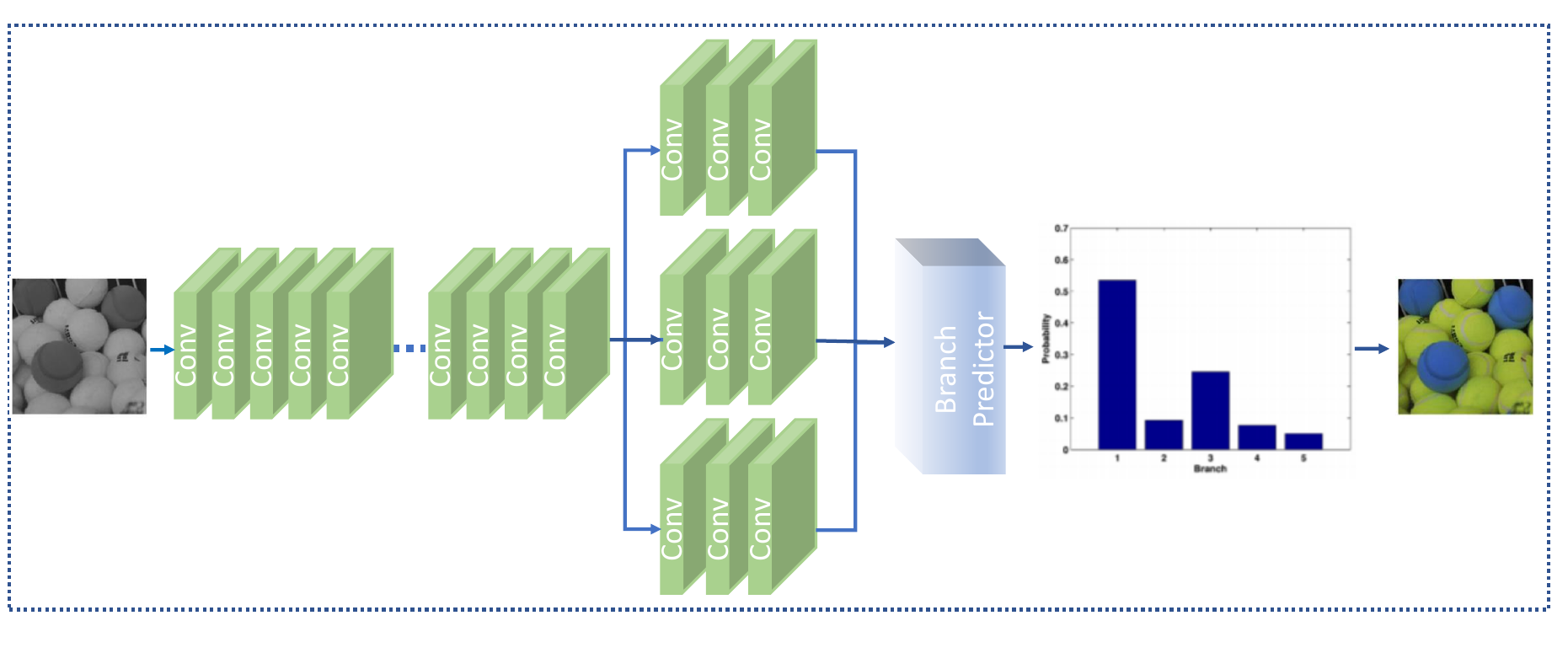}\\
Let there be color~\cite{iizuka2016let} & MHC~\cite{baig2017multiple}\\

\includegraphics[trim={3cm 6cm  6cm  2cm },clip,width=0.5\columnwidth,valign=t]{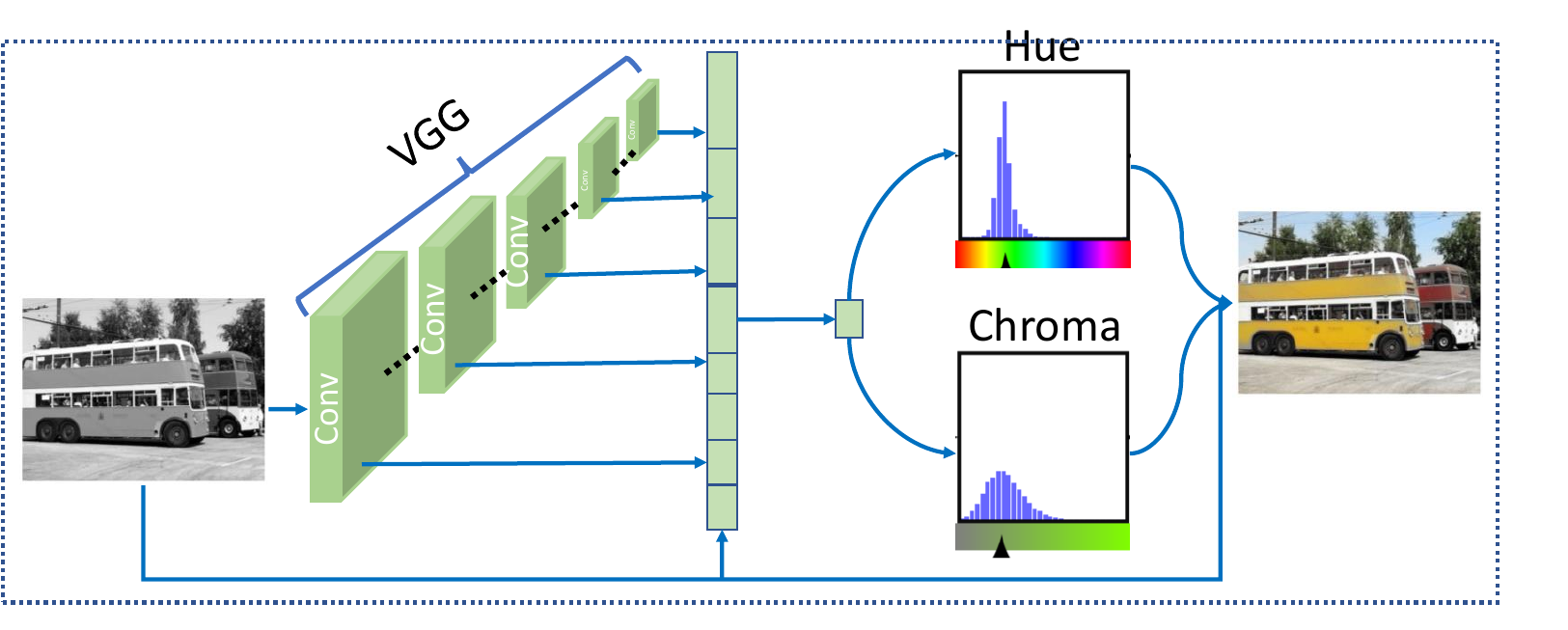}& 
\includegraphics[trim={1cm 9.5cm  2.2cm  -3.5cm },clip,width=0.5\columnwidth,valign=t]{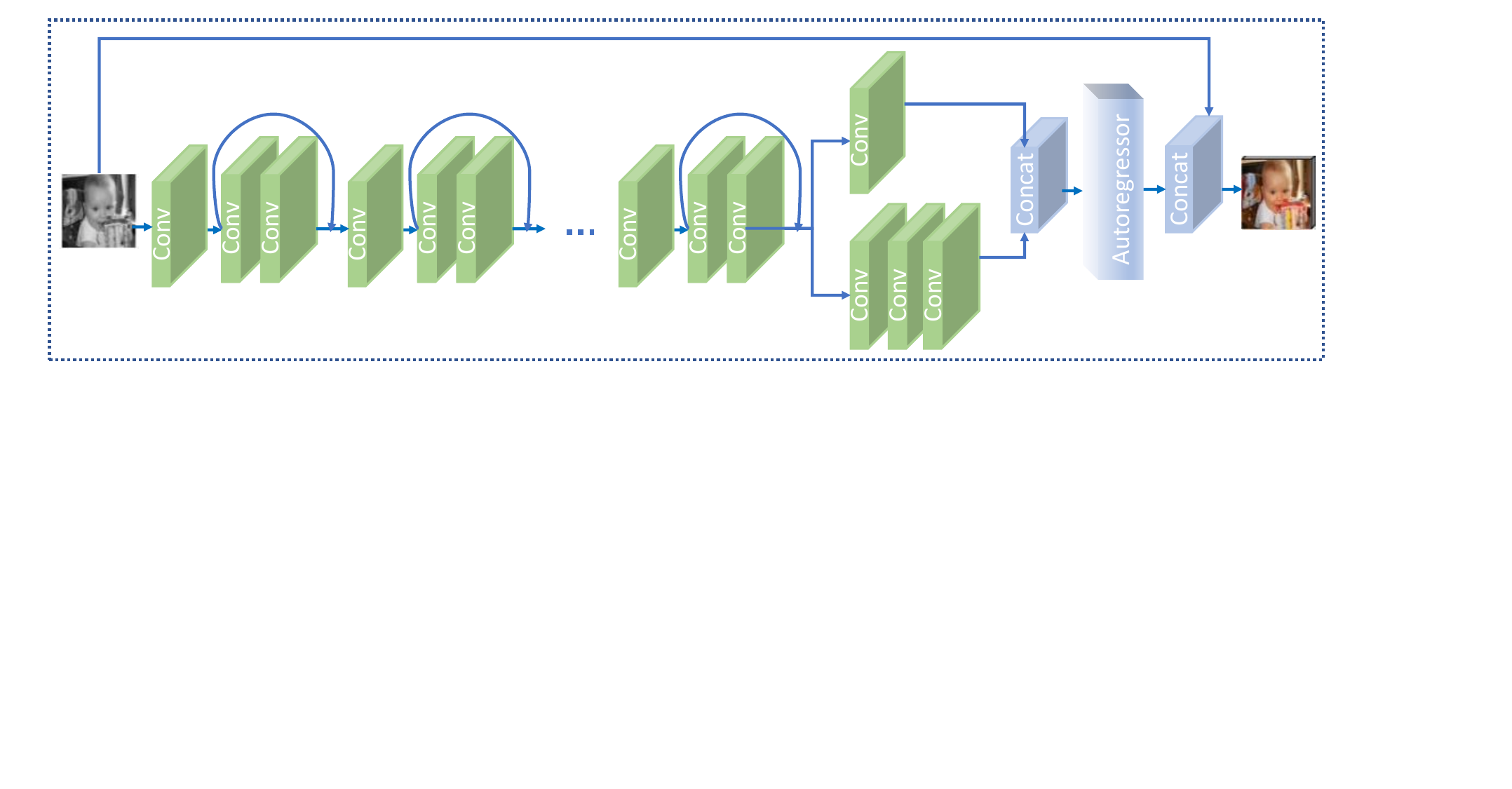}\\
Learning Automatic Colorization & Pixelated~\cite{zhao2019pixelated} \\

\includegraphics[trim={3cm 8.3cm  2cm  -1cm },clip,width=0.5\columnwidth,valign=t]{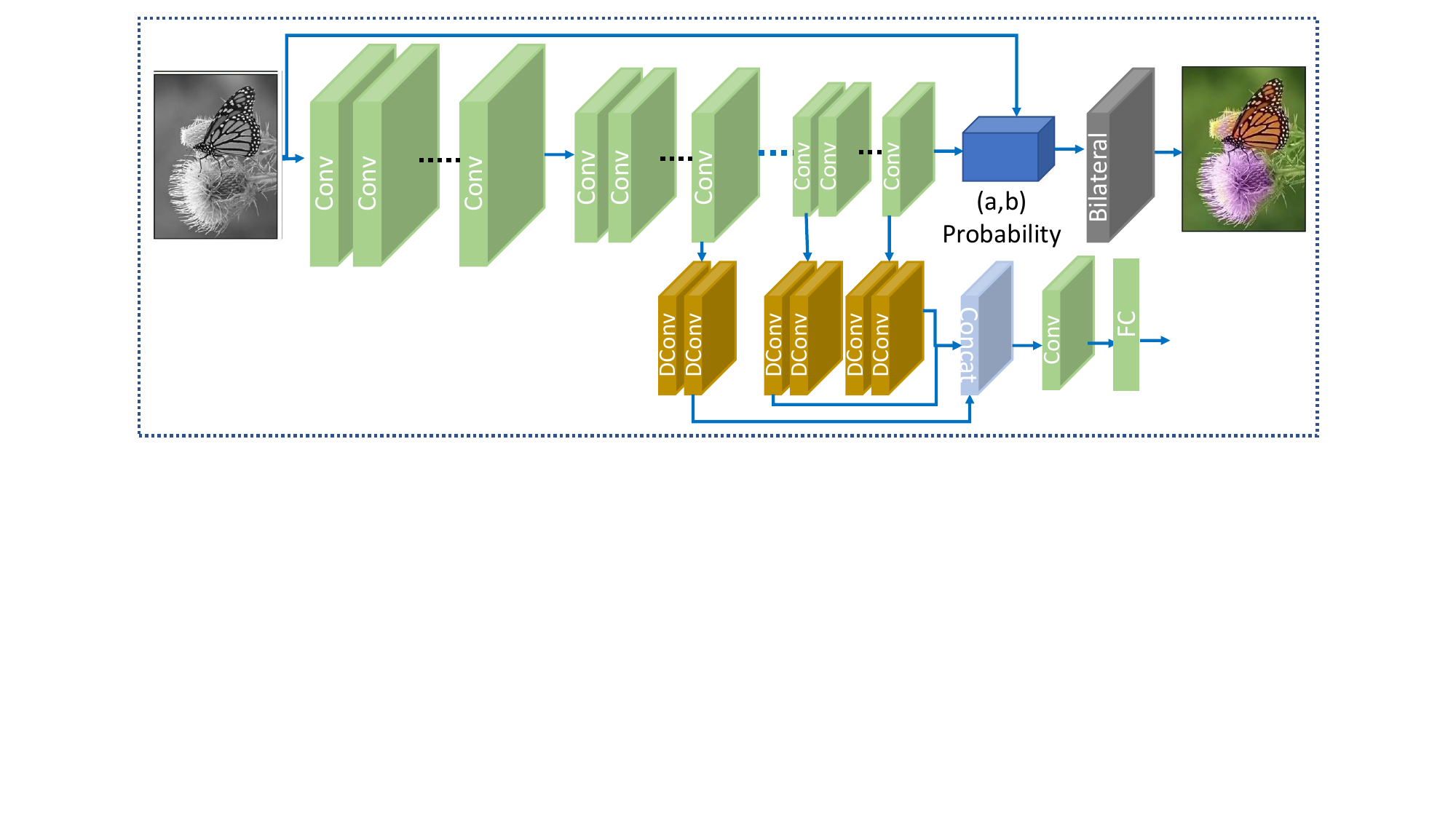}&
\includegraphics[trim={9cm 14cm  4cm  -1.5cm },clip,width=0.5\columnwidth,valign=t]{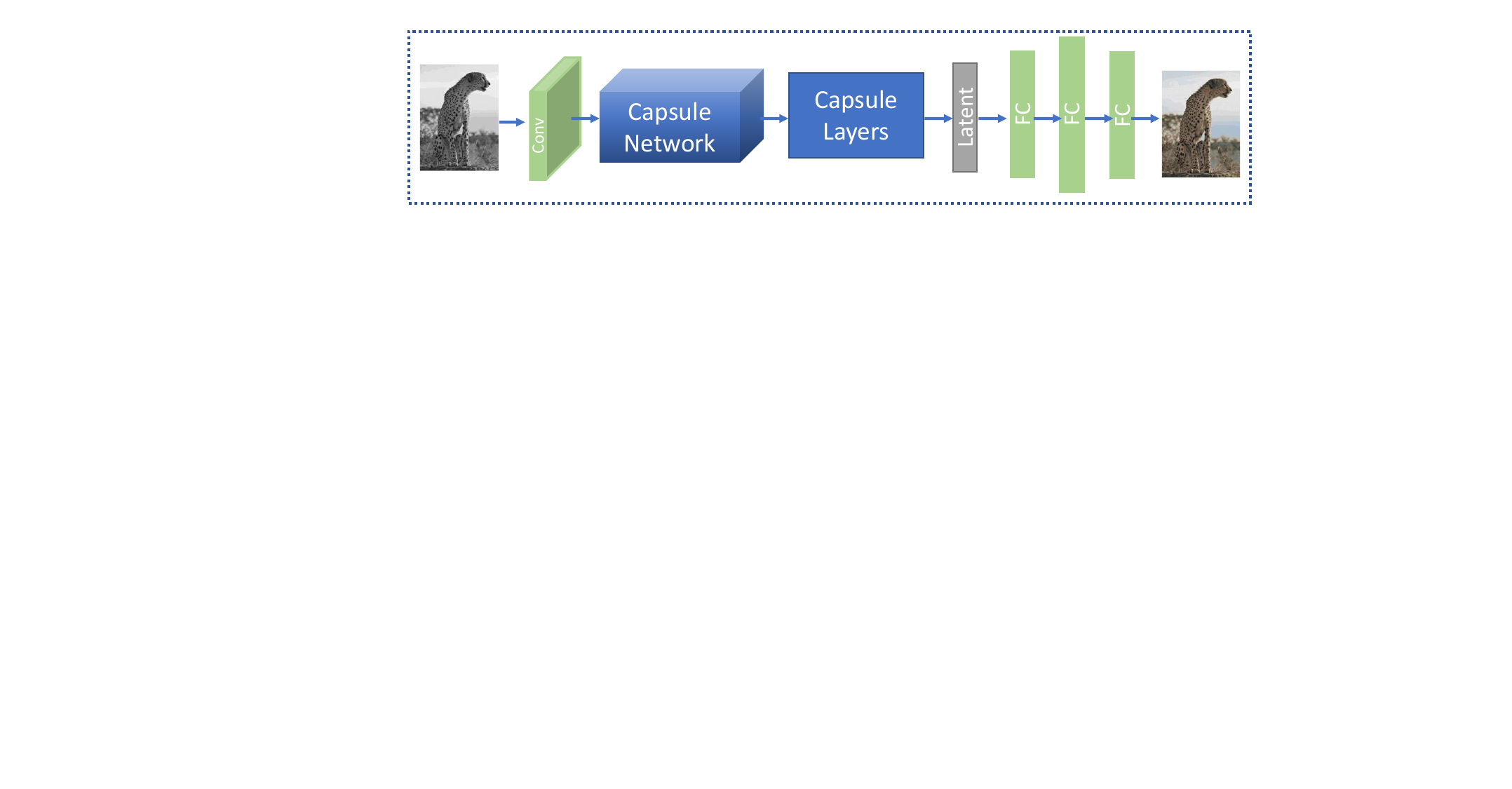}\\

 Pixel Level Colorization~\cite{Zhao0SH018} & ColorCapsNet~\cite{ozbulak2019image}\\

\multicolumn{2}{c}{\includegraphics[trim={7.5cm 7cm  10cm  0cm },clip,width=0.5\columnwidth,valign=t]{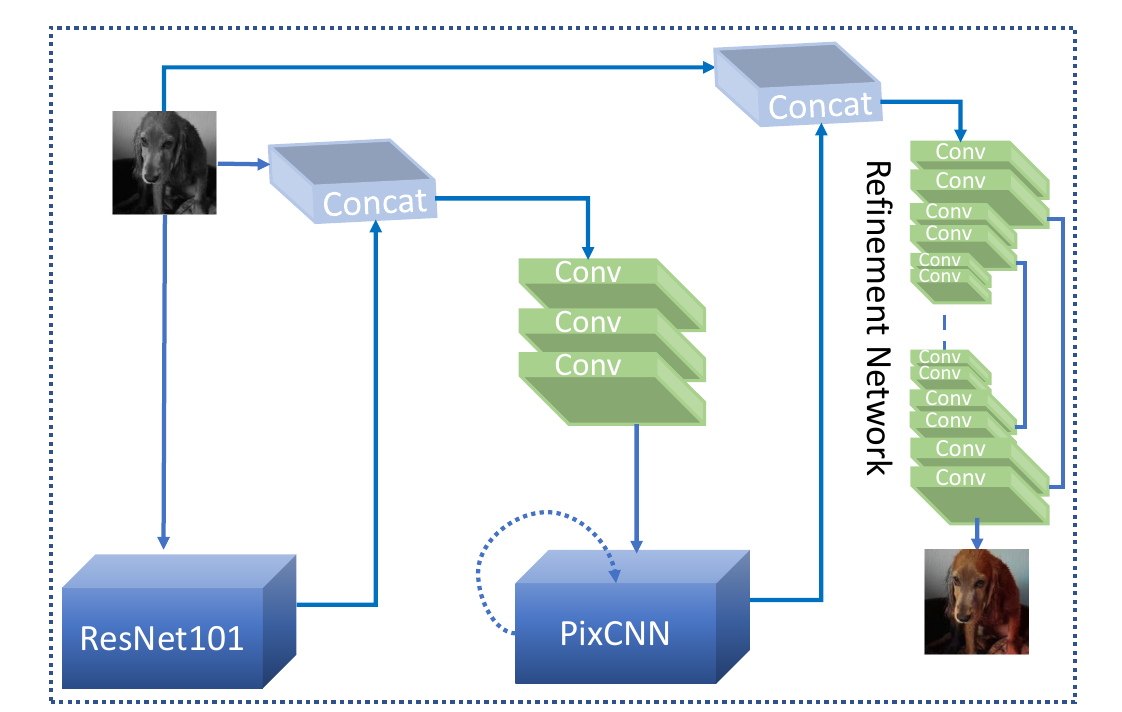}}\\
\multicolumn{2}{c}{PixColor~\cite{guadarrama2017pixcolor}}\\
\end{tabular}
\end{center}
\caption{\textbf{Examples of multi-path networks.} These colorization networks learn different features via several network paths.}
\label{fig:MultiPath}
\end{figure}

The multi-path networks follow multiple different parts to learn features at various levels or paths. The multi-path models colorize images by learning different visual features and contexts across several branches. This multi-path technique allows the accurate interpretation of a range of features. On the other hand, its disadvantage is that it often demands a significant amount of memory and computing resources to manage the multiple branches and feature concatenation; hence, it usually takes more time to train the models. The following are examples of multi-path networks, and Figure~\ref{fig:MultiPath} provides their architectures.

\subsubsection{Let there be Color}
Iizuka~\etal~\cite{iizuka2016let} designed a neural network\footnote{\url{https://github.com/satoshiiizuka/siggraph2016_colorization}} to primarily colorize the grayscale images and provide scene classification as a secondary task. The proposed system has two branches for learning features at multiple scales. Both branches are further divided into four subnets, i.e., the low-level features subnet, mid-level features subnet, global features subnet, and colorization network. The low-level subnet decreases the resolution and extracts edges and corners. Mid-level features come in handy when learning the textures.

Similarly, the global subnet computes a 256D vector to represent the image. The mid-level and global features are combined and fed into the colorization network to predict the chrominance channel. The final output is restored to the original input resolution. The global features assist the classification network in classifying the image scene. Both the colorization network and scene classification network are jointly trained.

The low-level features subnet consists of six convolution layers; the mid-level features subnet consists of two convolutional layers; and the global features network consists of four convolutional layers and three fully connected layers. The colorization network comprises a fusion layer, four convolutional, and two upsampling layers. To feed input to the network, the image is resized to 256$\times$256, and then the output is upsampled to its original resolution. The kernel is of size 3$\times$3; feature channels vary between 64 and 512, and striding is used to reduce the feature map size. Moreover, the learning rate is determined automatically using ADADELTA~\cite{zeiler2012adadelta}, the network is optimized using stochastic gradient descent (SGD)~\cite{bottou1991stochastic}, and the system is designed using Torch~\cite{collobert2011torch7}.

\subsubsection{Learning Representations for Automatic Colorization}
Learning Representations for Automatic Colorization\footnote{Code is available at \url{https://github.com/gustavla/autocolorize}}~\cite{larsson2016learning} aims to learn a mapping function taking per-pixel features as hypercolumns of localized slices from existing CNN networks to predict hue and chroma distributions for each pixel. The authors use VGG16~\cite{simonyan2015vgg} as a feature extractor, discarding the classification layer and using a grayscale image as input rather than RGB. For each pixel, a hypercolumn is extracted from all the VGG16 layers except the last classification layer, resulting in a 12k channel feature descriptor fed into a fully connected layer of 1k channels, which predicts the final hue and chroma outputs.

The KL divergence is employed as a loss to predict distributions over a set of color bins. The input size is 256$\times$256, and the framework used is Caffe~\cite{jia2014caffe}, where the network is fine-tuned for ten epochs, each taking about 17 hours. Other parameters and optimizations are similar to VGG16~\cite{simonyan2015vgg}.

\subsubsection{PixColor}
PixColor, proposed by \cite{guadarrama2017pixcolor}, employs a conditional PixelCNN to produce a low-resolution color image from a given grayscale image. A refinement CNN is then trained to process the original grayscale and the low-resolution color image to produce a high-resolution colorized image. The color of the previous pixels determines the colorization of an individual pixel.

PixelCNN is composed of two convolutional layers and three ResNet blocks~\cite{he2016ResNet}. The first convolutional layer uses kernels of size $7\times7$, while the last employs $3\times3$. Similarly, ResNet blocks contain 3, 4, and 23 convolutional layers, respectively. The PixelCNN colorization network is composed of three masked convolutional layers: one at the beginning and the second at the end of the network, whereas a gated convolutional block with ten layers is surrounded by the gated convolutional layers.

The PixelCNN model is trained by applying a maximum likelihood function with cross-entropy loss. Then, the refinement network, composed of 16 convolutional layers followed by two bilinear upsampling layers, each with two internal convolutional layers, is trained on the ground-truth chroma images downsampled to $28\times28$. The network ends with three convolutional layers, where the final layer outputs the colorized image.

\subsubsection{Colorize Capsule Network}
Colorize Capsule Network (ColorCapsNet)~\cite{ozbulak2019image} is built upon CapsNet~\cite{sabour2017dynamic} with three modifications. Firstly, the single convolutional layer is replaced by the first two convolutional layers of VGG19~\cite{simonyan2015vgg} and initialized via its weights. Secondly, batch normalization~\cite{ioffe2015batch} is inserted in between the first two convolutional layers. Thirdly, the capsule layer reduces the number of capsules from ten to six. The architecture of ColorCapsNet is similar to an autoencoder. The latent space processes the colorization-specific hidden variables between the encoder and decoder.

To train the model, the input RGB image is first converted into the CIE Lab colorspace, and extracted patches are fed to the network to learn the color distribution. The output is in the form of colorized patches, combined to obtain the complete image in Lab colorspace and later converted to RGB colorspace.

The difference between real and generated images is minimized using the mean squared error loss. ColorCapsNet is trained on ILSVRC 2012 to learn objects' general color distribution, which is then fine-tuned using the DIV2K dataset~\cite{agustsson2017DIV2K}. ColorCapsNet shows comparable performance to other models despite its shallow architecture. Adam is used as the optimizer with a learning rate of 0.001 and different kernel sizes.

\subsubsection{Pixelated}
Pixelated, introduced by Zhao~\etal~\cite{zhao2019pixelated}, is an image colorization model guided by pixelated semantics to keep the colorization consistent across multiple object categories. The network architecture is composed of a color embedding branch and a semantic segmentation branch. The network is built from gated residual blocks containing two convolutional layers, a skip connection, and a gating mechanism.

Three components make up the learning mechanism. \emph{Firstly}, an autoregressive model is adopted to utilize pixelated semantics for image colorization. The model uses a shared CNN to generate per-pixel distributions through a conditional Pixel-CNN, achieving colorization diversity. \emph{Secondly}, semantic segmentation is incorporated into color embedding by introducing atrous spatial pyramid pooling at the top layer capable of extracting multi-scale features using several parallel filters. The output is obtained by fusing multi-scale features. For semantic segmentation, the loss function is the cross-entropy loss with the softmax function. \emph{Thirdly}, to produce better and more accurate color distributions, pixelated color embedding is concatenated with semantic segmentation to create the semantic generator.

Color embedding, semantics, and color generation losses are combined during training. The network performance is evaluated on the Pascal VOC2012~\cite{everingham2010pascal} and COCO-stuff~\cite{lin2014coco} datasets. The images are rescaled to 128$\times$128 for use in the network.

\begin{figure}[t]
\begin{center}
\begin{tabular}{cc} 
\includegraphics[trim={3cm 6cm  2cm  5cm },clip,width=0.5\columnwidth,valign=t]{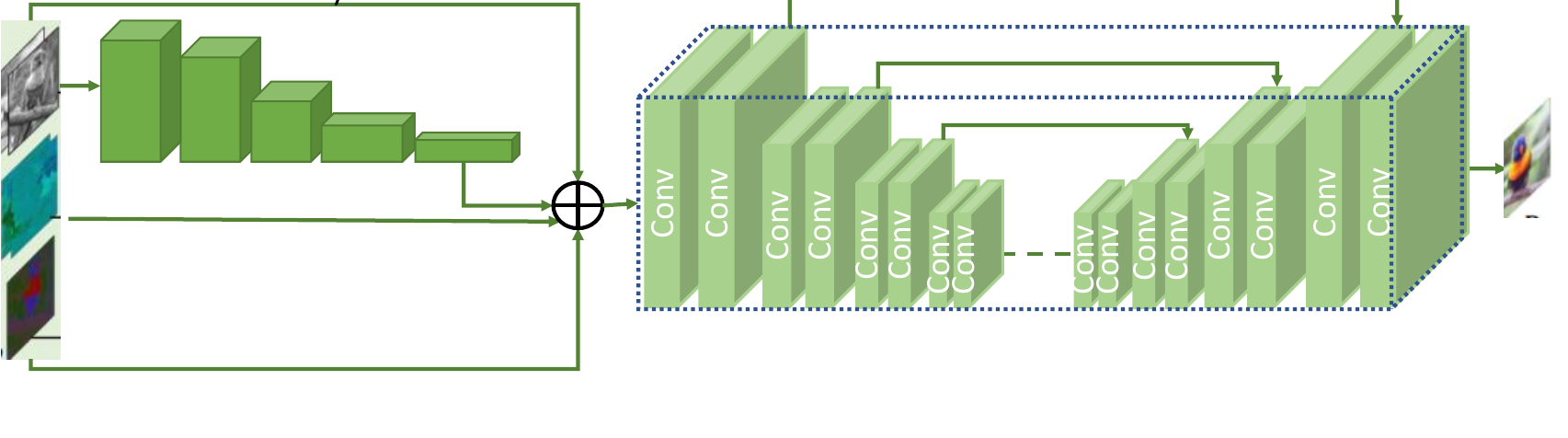}&
\includegraphics[trim={0cm 12cm  0cm  0cm },clip,width=0.5\columnwidth,valign=t]{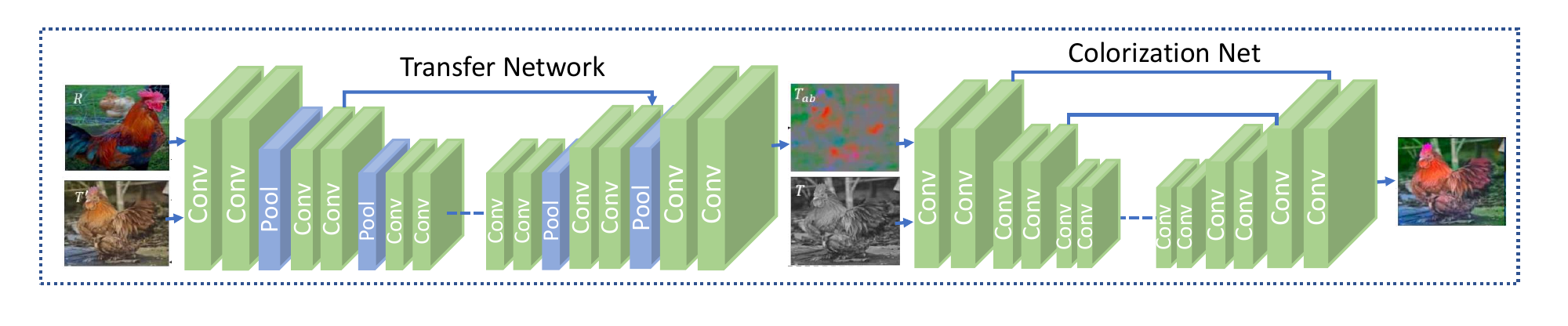}\\
Deep Exemplar Colorization~\cite{he2018deep} & Fast Deep Exemplar Colorization~\cite{xu2020CVPR}\\
\multicolumn{2}{c}{\includegraphics[trim={5.5cm 4cm  5cm  0cm },clip,width=0.5\columnwidth,valign=t]{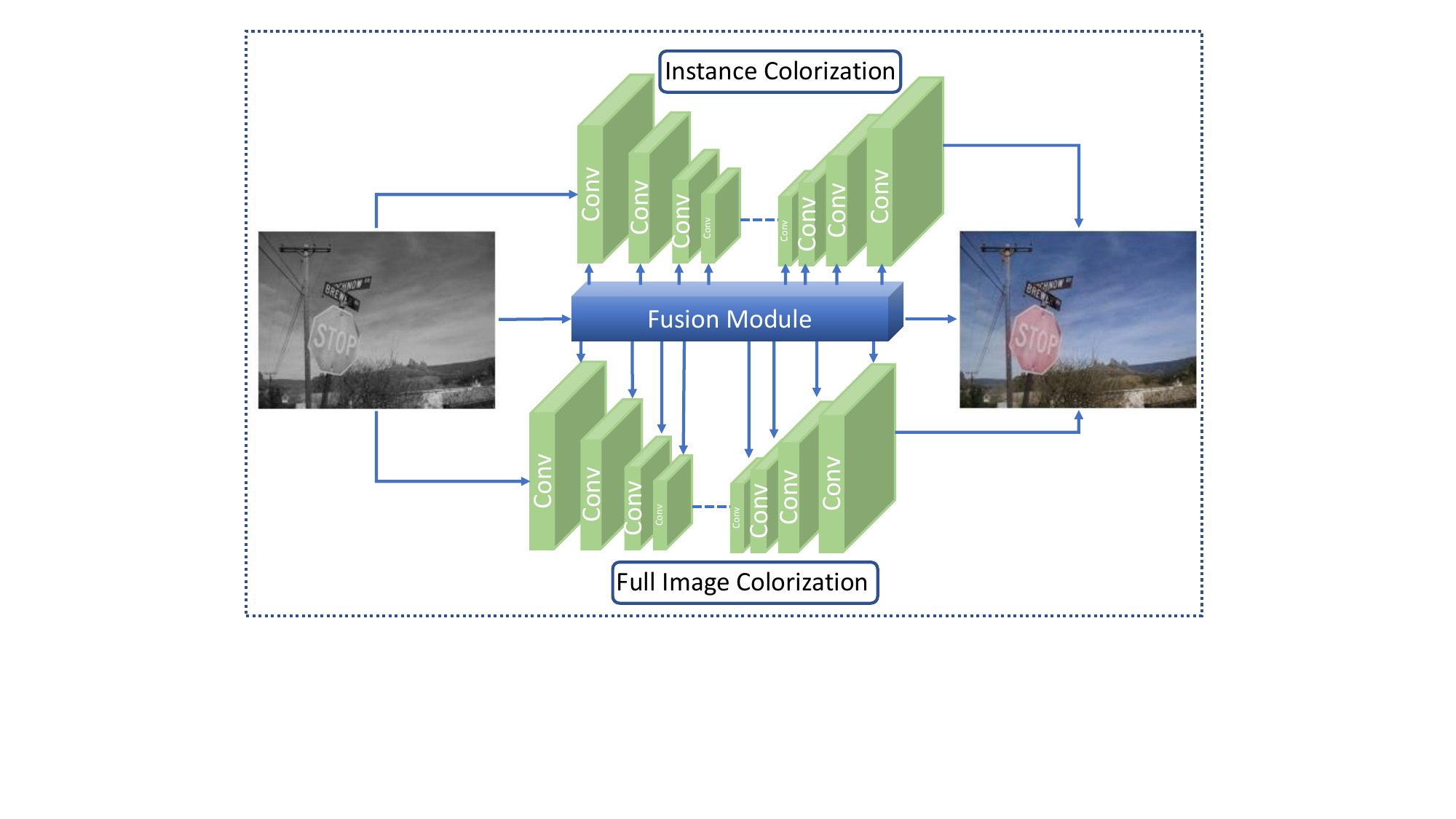}}\\
\multicolumn{2}{c}{Instance Colorization~\cite{su2020CVPR}}
\end{tabular}
\end{center}
\caption{\textbf{Exemplar-based Colorization networks.} These networks imitate the colors of the input example image provided along with the grayscale image.}
\label{fig:ExamplebasedColorization}
\end{figure}

\subsubsection{Multiple Hypothesis Colorization}
Mohammad \& Lorenzo~\cite{baig2017multiple} developed a multiple hypothesis colorization architecture, producing multiple color values for each pixel of the grayscale image. The low-cost colorization is achieved by storing the best pixel-level hypothesis. A common trunk consisting of convolutional layers computes the shared features. The trunk is then split into multiple branches, where each branch predicts color for each pixel. The main trunk's layers and subsequent branches are fully convolutional.

The authors developed two architectures, one for CIFAR100 and the other for ImageNet~\cite{deng2009imagenet}. The architecture for the CIFAR100 dataset is composed of 31 blocks preceded by a stand-alone convolutional layer. The number of channels in different layers across multiple blocks varies from 64 to 256. In contrast, the architecture for ImageNet~\cite{deng2009imagenet} is composed of 21 blocks. The number of channels across different layers inside blocks varies from 32 to 1024. Although the two models have a different number of blocks, the block structure remains the same. Both networks also utilize residual connections. Moreover, batch normalization is performed in all blocks, and ReLU is applied to all layers except the last of each block.

Softmax was employed as a loss function. The CIFAR100 model was trained for 40k iterations with a learning rate of 0.001, dropping it after 10k and 20k iterations by a factor of 10. Similarly, the ImageNet~\cite{deng2009imagenet} model was trained for 120k iterations with an initial learning rate of 0.0005 that is dropped after 40k and 80k iterations by a factor of 5.

\subsubsection{Pixel-Level Semantics Guided Image Colorization}
A hierarchical network comprised of a semantic segmentation branch and a colorization branch was developed by Zhao~\etal~\cite{Zhao0SH018}. The two branches share the first four convolution layers to learn low-level features. Four more convolutional layers extend the colorization branch, while three deconvolution layers come from the segmentation branch. The output of the deconvolutional layers is concatenated and passed through the final convolutional layer to produce class probabilities.

Semantic segmentation is achieved through weighted cross-entropy loss and softmax function, whereas colorization is performed via multinomial cross-entropy loss with class rebalance. During the training, the network jointly optimizes the two tasks. A joint bilateral upsampling layer is introduced during the testing to generate the final output.

The model was trained on 10,582 for semantic segmentation and tested on 1449 images from the PASCAL VOC2012 dataset for 40 epochs. Similarly, 9k images were used for training and 1k for testing on the COCO-stuff dataset~\cite{lin2014coco} for 20 epochs.

\subsubsection{Shadow-Aware Image Colorization}
Duan~\etal\cite{duan2024shadow-new1} introduced a novel end-to-end dual-branch shadow-aware image colorization network that accurately colorizes shadow areas by incorporating shadow information with semantic understanding and preserves saturated colors. The shadow-aware block helps identify the shadow areas due to its ability to integrate shadow-specific information. The proposed methodology is capable of colorizing shadow and non-shadow areas separately.

The color information colorizes the non-shadow areas of the input image, whereas colorless information of shadow areas is used to colorize the shadow areas of the image. Separate processing of colorizing the shadow and non-shadow areas preserves the saturation and diversity of the non-shadow regions. Similarly, the shadow areas are colorized using colorless shadow information so that the overall saturation of the image is not hampered. The two-branch network has a shadow detection branch and a shadow-aware colorization branch. The shadow detection branch is an encoder that uses transformers to pull out multi-level features, and the shadow-aware colorization branch is another encoder that pulls out shadow-aware color features. Both the encoders share features that lead to effective colorization of the input image.

\subsection{Exemplar-based Colorization}
Exemplar-based colorization utilizes the colors of example images provided along with input grayscale pictures and uses them to learn color distributions from related photos to improve the colorized output. When the exemplars closely match the content of the target pictures, they can capture subtle color changes and characteristics and generate compelling results. However, its effectiveness depends highly on the exemplars. Therefore, an inadequate or mismatched exemplar might provide an inferior colorized image. It is also expensive to find and process the reference images. Furthermore, the technique may only be applicable if suitable exemplars are available. Figure~\ref{fig:ExamplebasedColorization} presents the example networks for exemplar-based colorization. We provide more details about the networks in the subsequent subsections.

\subsubsection{Deep Exemplar-based Colorization}
Deep exemplar-based colorization\footnote{\url{https://github.com/msracver/Deep-Exemplar-based-Colorization}}~\cite{he2018deep} transfers the colors from a reference image to the grayscale one. The aim here is not to colorize images naturally but to provide diverse colors to the same image. The system consists of two subnetworks: the Similarity subnetwork and the Colorization subnetwork.

The similarity subnet takes the target and reference luminance channels aligned via Deep Image Analogy~\cite{liao2017visual}. The authors use the standard features of VGG19~\cite{simonyan2015vgg} after each block, resulting in five levels of coarse to fine feature maps. The features are upsampled to the same size. The similarity subnet computes a bidirectional similarity map using discrete cosine distance. Furthermore, the colorization subnet concatenates the grayscale image, chrominance channels, and the calculated similarity maps. U-Net~\cite{ronneberger2015u} inspires the structure of the colorization subnet.

The loss function was $\ell_2$ for a combination of chrominance channels and perceptual loss. For ten epochs, the network was trained using the ADAM~\cite{kingma2014adam} optimizer in the Caffe~\cite{jia2014caffe} framework, with a learning rate of $10^{-3}$. This rate was lowered by 0.1 after 33\% of training.






\begin{figure}[t]
\begin{center}
\begin{tabular}{c} 
\includegraphics[width=0.9\textwidth,valign=t]{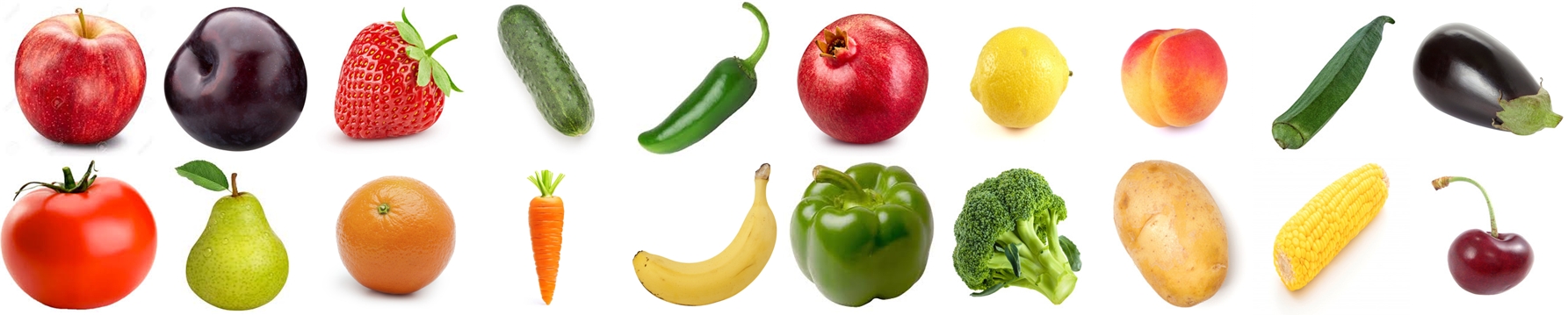}\\
\end{tabular}
\end{center}
\caption{Sample images for each category from our proposed Natural-Color dataset (NCD).}
\label{fig:Sample_images}
\end{figure}

\subsubsection{Fast Deep Exemplar Colorization}
CCurrent exemplar-based colorization methods suffer from two challenges: 1) they are sensitive to selecting reference images, and 2) they have high time and resource consumption. Inspired by stylization characteristics in feature extraction and blending, Xu~\etal~\cite{xu2020CVPR} proposed a stylization-based architecture for fast deep exemplar colorization. The proposed architecture consists of a transfer sub-net that learns a coarse chrominance map (ab map in CIELab color space) and a colorization sub-net that refines the map to generate the final colored result. The proposed method seeks to produce realistic colorization results in real time, regardless of the semantic relationship between the input and exemplar image.

More specifically, an encoder-decoder structure is used for the transfer sub-net, which takes the target-reference image pairs as the input and output initial ab map. The pre-trained VGG19 module (from the $conv1_1$ layer to the $conv4_1$ layer) is treated as the encoder and a symmetrical decoder for image reconstruction. In addition, the fast Adaptive Instance Normalization (AdaIN) \cite{AdaIN} is utilized after convolutional layers to accelerate feature matching and blending. The ab map generated by the transfer sub-net is inaccurate and has some artifacts. To refine it, a colorization sub-net that adopts an analogous U-Net structure is designed, taking a known luminance map and an initial chrominance ab map as input.

In the original implementations, the transfer sub-net is trained on the Microsoft COCO dataset~\cite{lin2014coco} by minimizing the sum of the $\ell_2$ loss, while the colorization sub-net is trained on the ImageNet dataset~\cite{deng2009imagenet} by minimizing the Huber loss~\cite{Huberloss}.

\begin{figure}[t]
\begin{center}
\begin{tabular}{c@{ }  c@{ } c@{ } c@{ } c@{ }	c@{ }	c@{ }  c}
    
    \includegraphics[width=.115\textwidth]{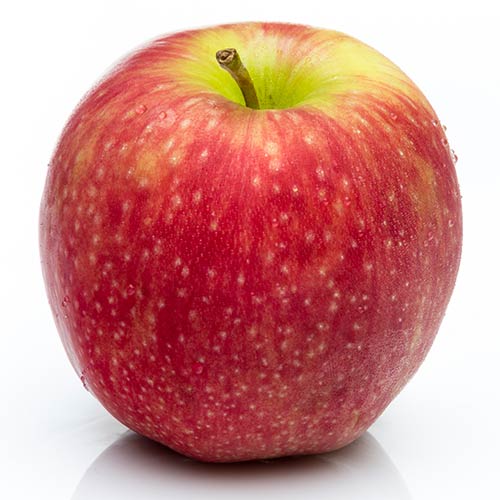} &  
    \includegraphics[width=.115\textwidth]{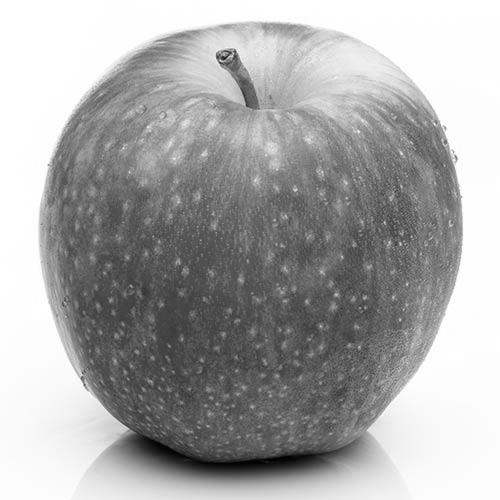}&
    \includegraphics[width=.115\textwidth]{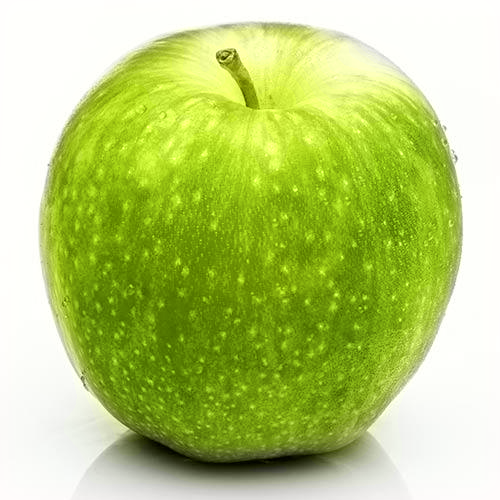}&   
	\includegraphics[width=.115\textwidth]{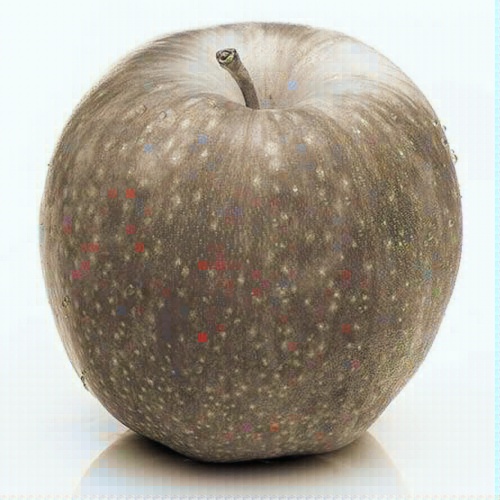}&    
	\includegraphics[width=.115\textwidth]{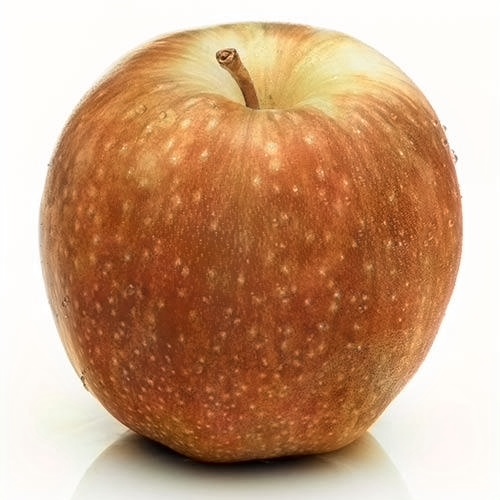}&    
	\includegraphics[width=.115\textwidth]{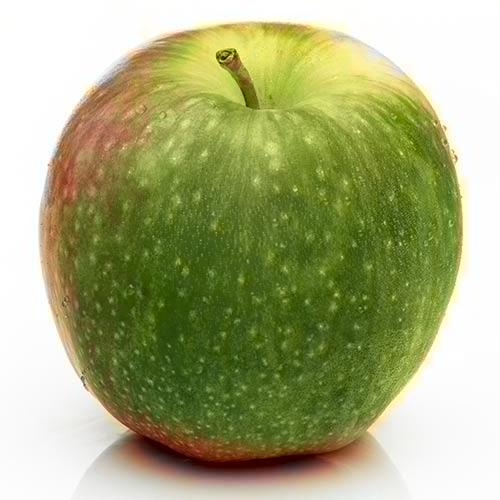}& 
	\includegraphics[width=.115\textwidth]{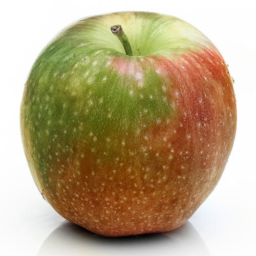}& 
  	\includegraphics[width=.115\textwidth]{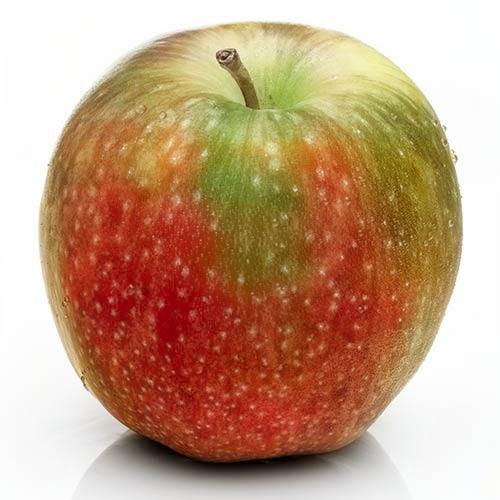}\\
  	
  	\includegraphics[width=.115\textwidth]{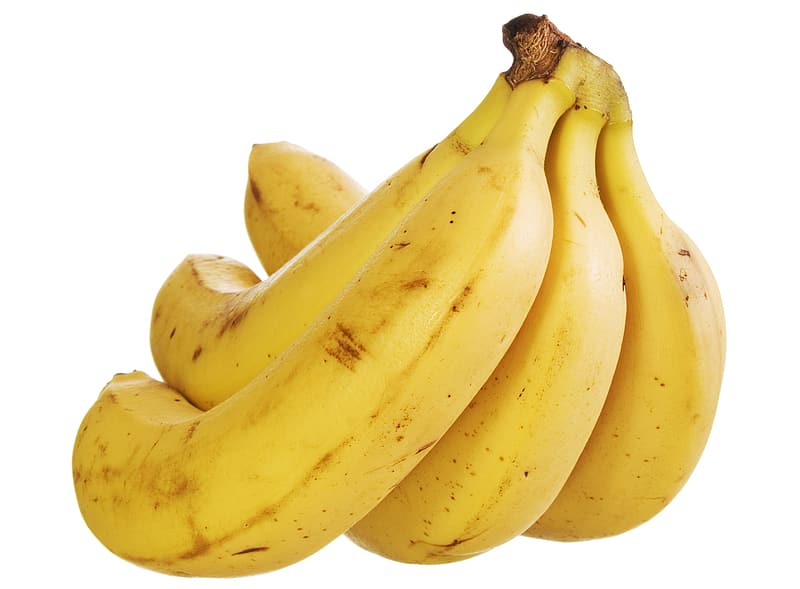}&  
    \includegraphics[width=.115\textwidth]{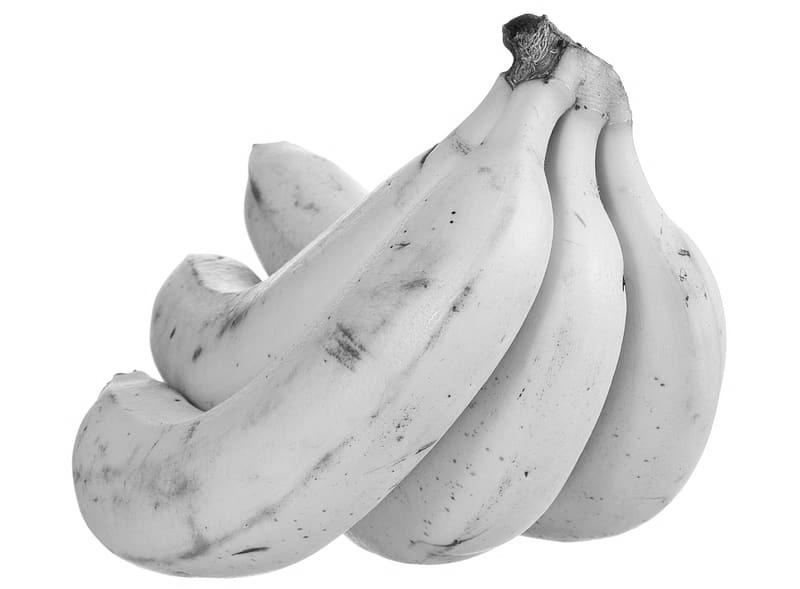}&
    \includegraphics[width=.115\textwidth]{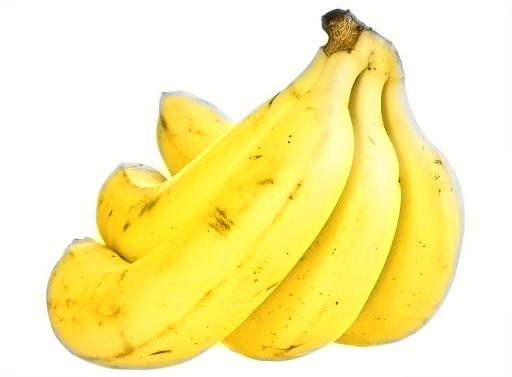}&   
	\includegraphics[width=.115\textwidth]{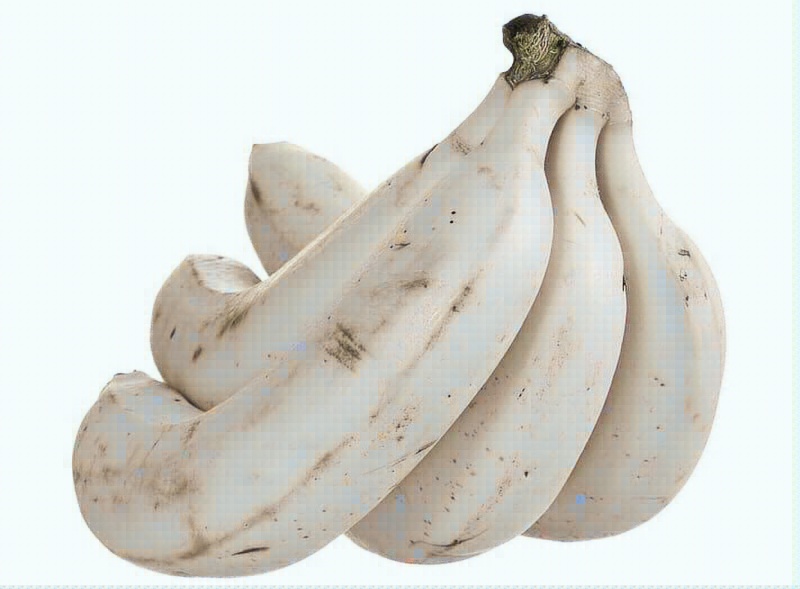}&    
	\includegraphics[width=.115\textwidth]{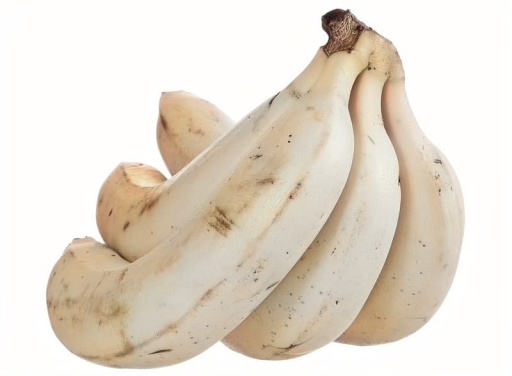}&    
	\includegraphics[width=.115\textwidth]{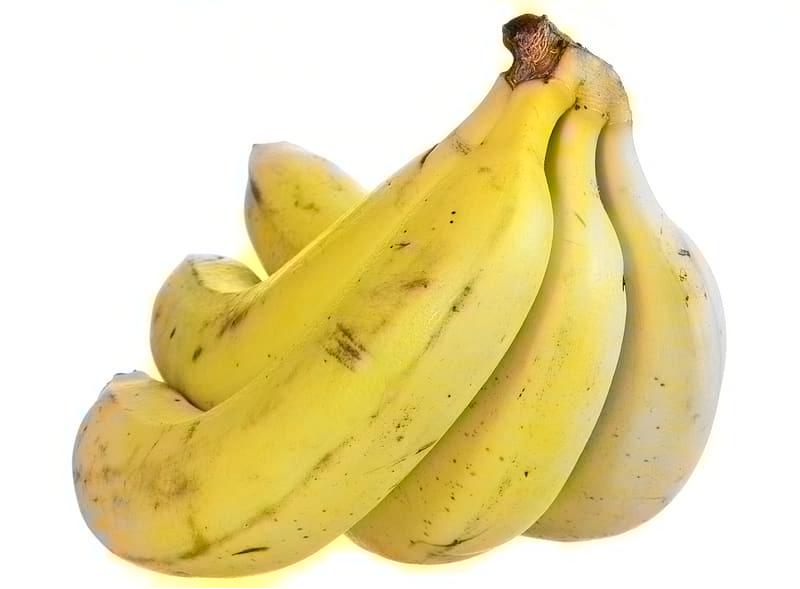}& 
	\includegraphics[width=.115\textwidth]{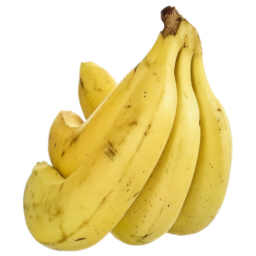}& 
  	\includegraphics[width=.115\textwidth]{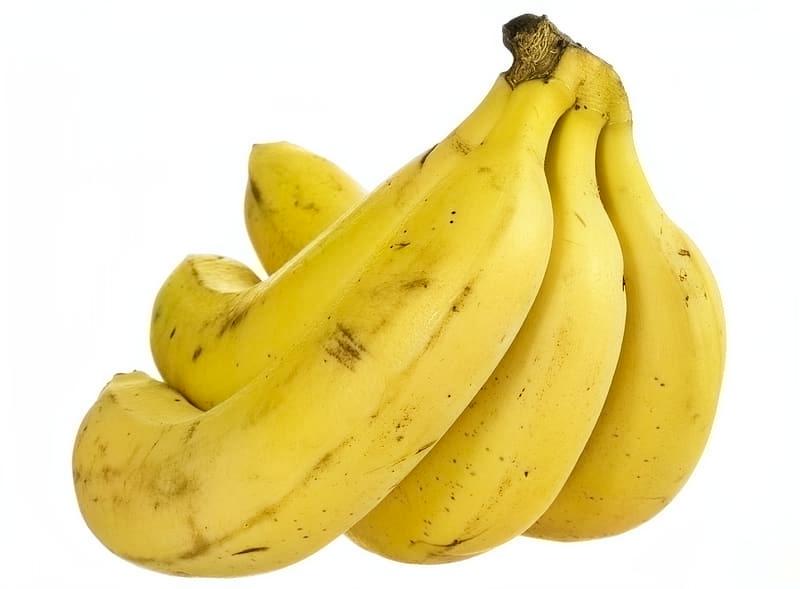}\\

  	\includegraphics[width=.115\textwidth]{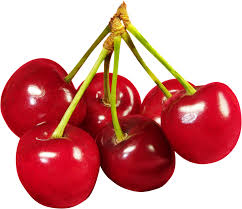}&  
    \includegraphics[width=.115\textwidth]{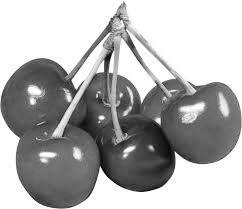}&
    \includegraphics[width=.115\textwidth]{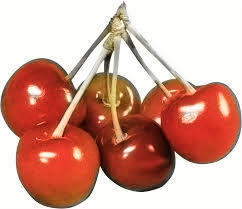}&   
	\includegraphics[width=.115\textwidth]{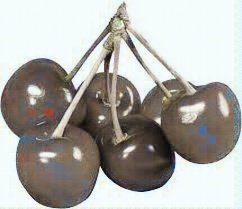}&    
	\includegraphics[width=.115\textwidth]{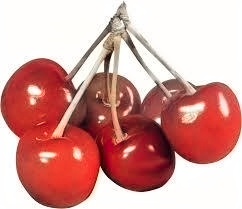}&    
	\includegraphics[width=.115\textwidth]{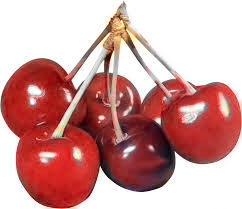}& 
	\includegraphics[width=.115\textwidth]{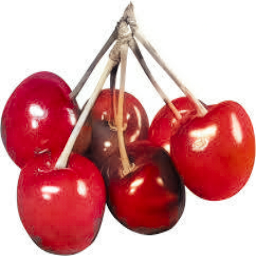}& 
  	\includegraphics[width=.115\textwidth]{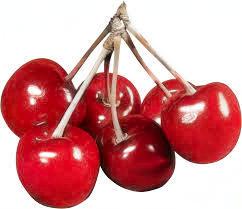}\\

  	\includegraphics[width=.115\textwidth]{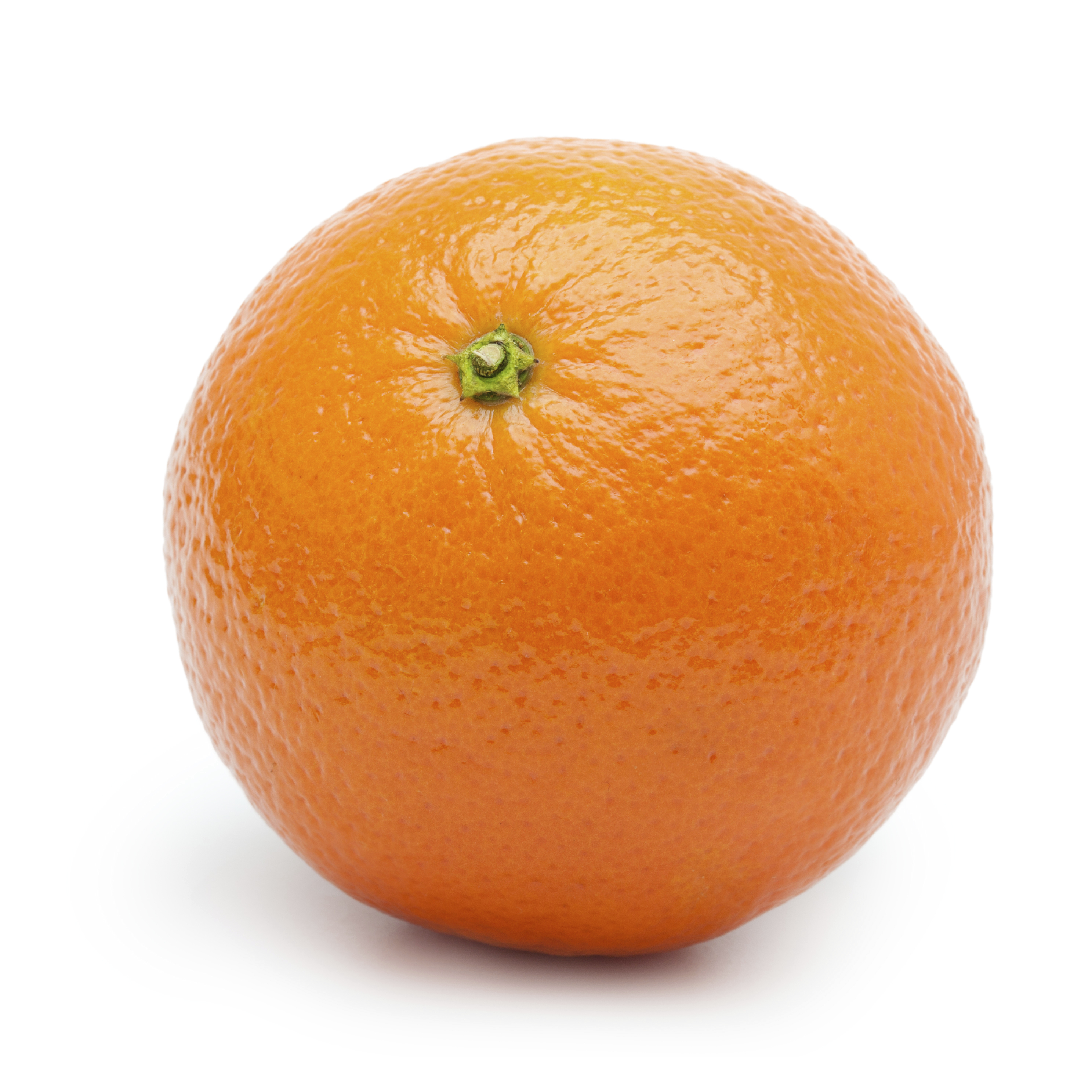}&  
    \includegraphics[width=.115\textwidth]{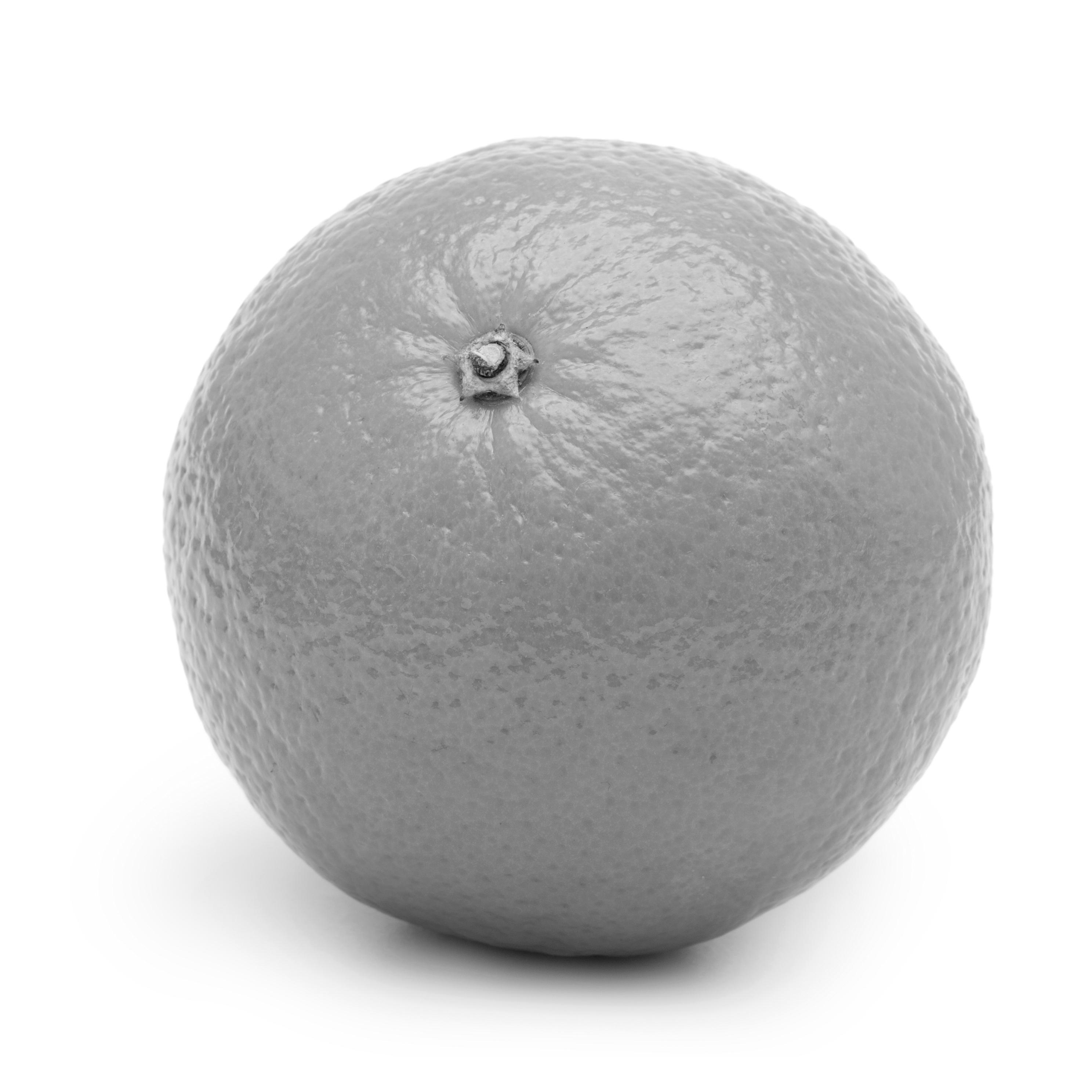}&
    \includegraphics[width=.115\textwidth]{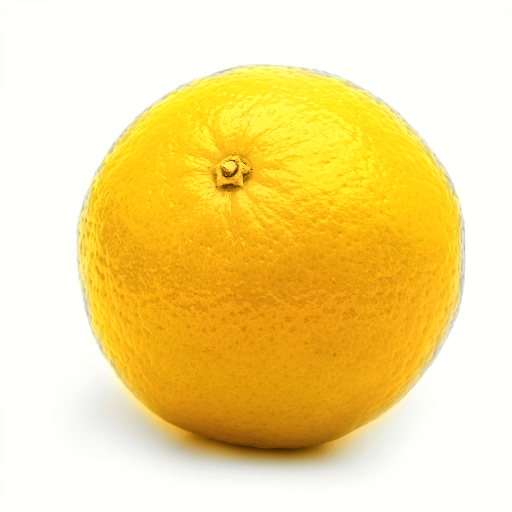}&   
	\includegraphics[width=.115\textwidth]{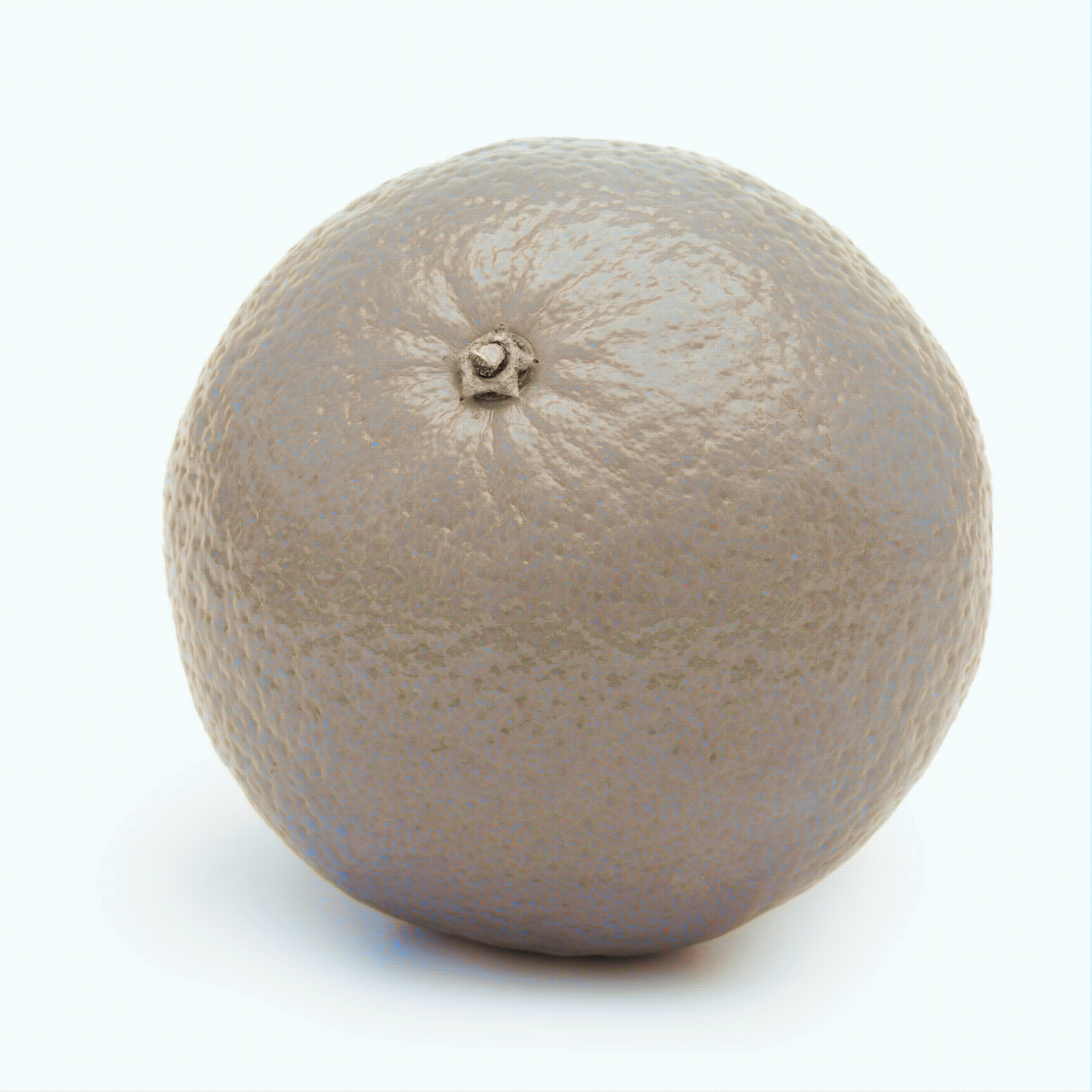}&    
	\includegraphics[width=.115\textwidth]{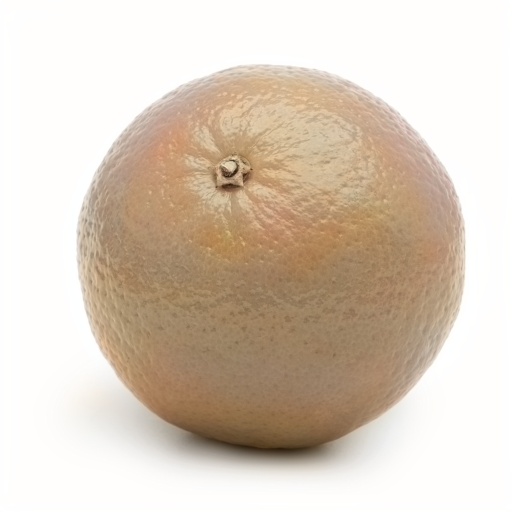}&    
	\includegraphics[width=.115\textwidth]{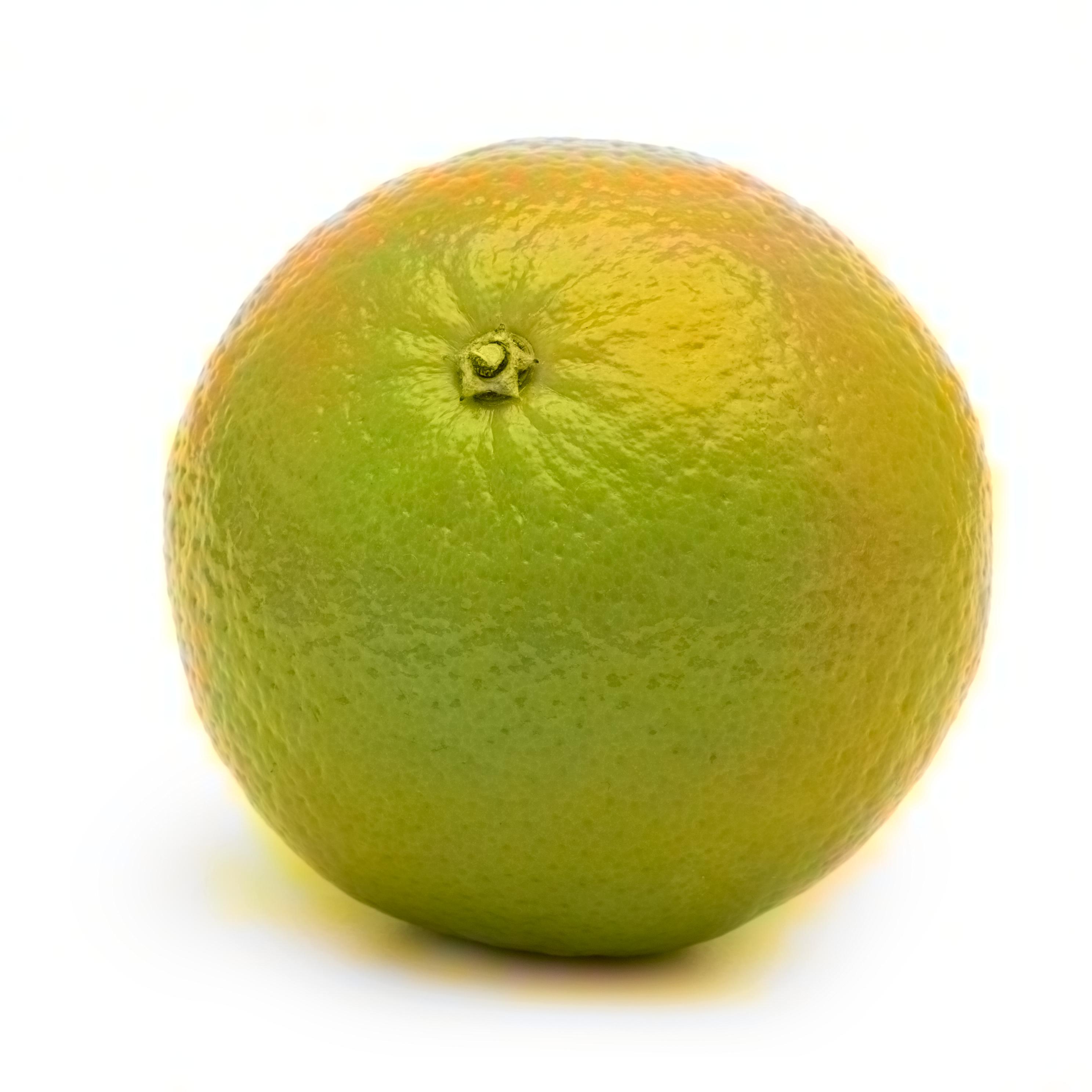}& 
	\includegraphics[width=.115\textwidth]{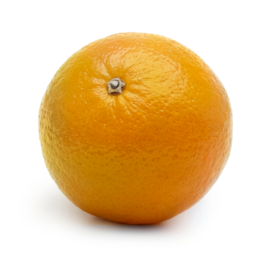}& 
  	\includegraphics[width=.115\textwidth]{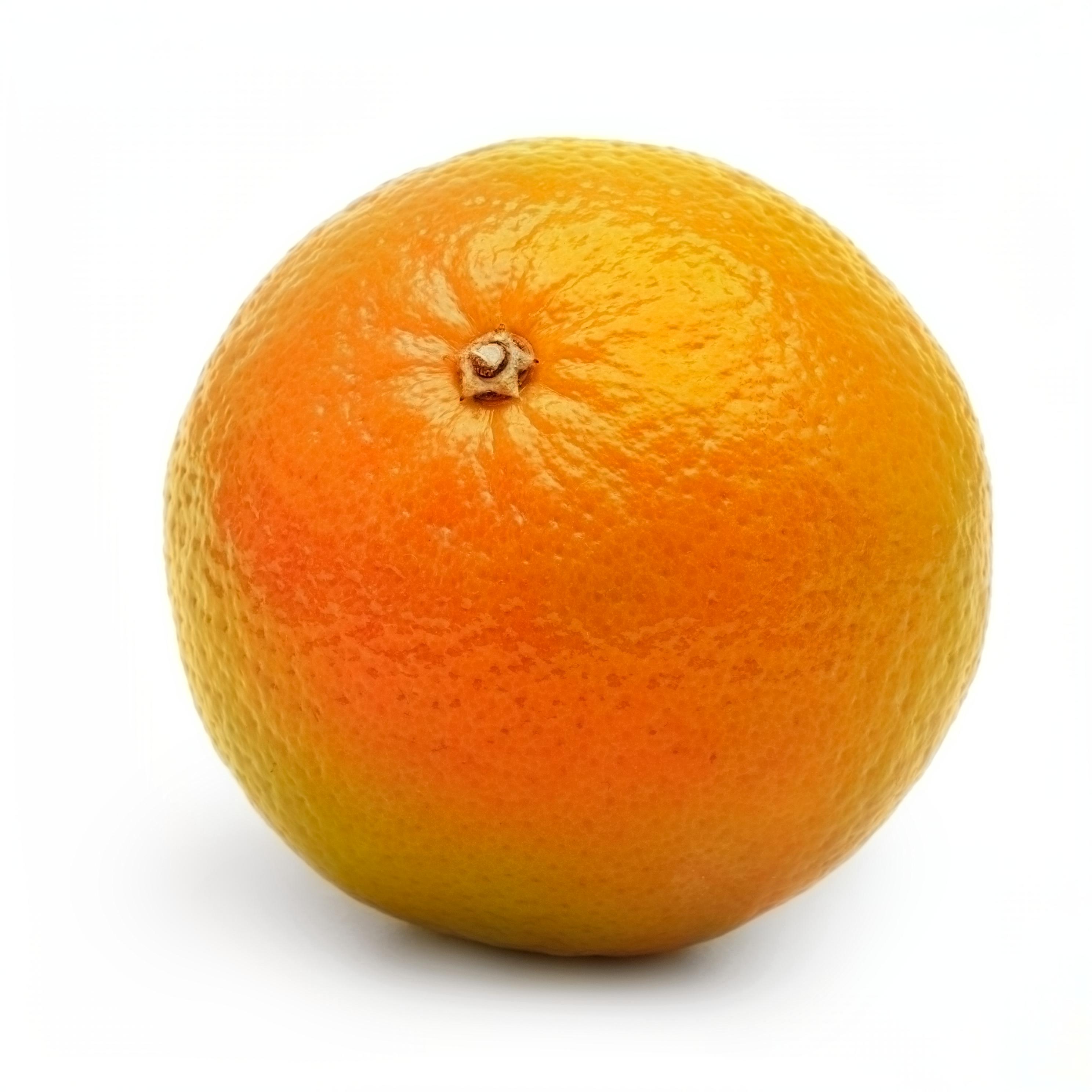}\\

  	\includegraphics[width=.115\textwidth]{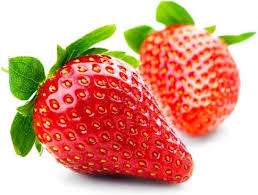}&  
    \includegraphics[width=.115\textwidth]{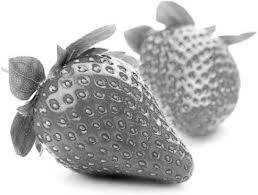}&
    \includegraphics[width=.115\textwidth]{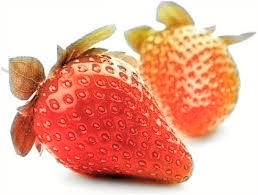}&   
	\includegraphics[width=.115\textwidth]{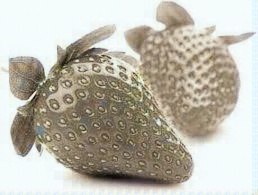}&    
	\includegraphics[width=.115\textwidth]{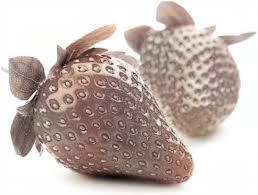}&    
	\includegraphics[width=.115\textwidth]{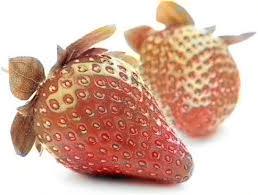}& 
	\includegraphics[width=.115\textwidth]{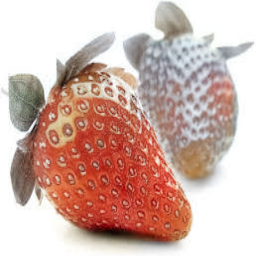}& 
  	\includegraphics[width=.115\textwidth]{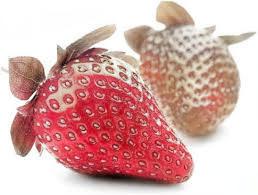}\\

  	\includegraphics[width=.115\textwidth]{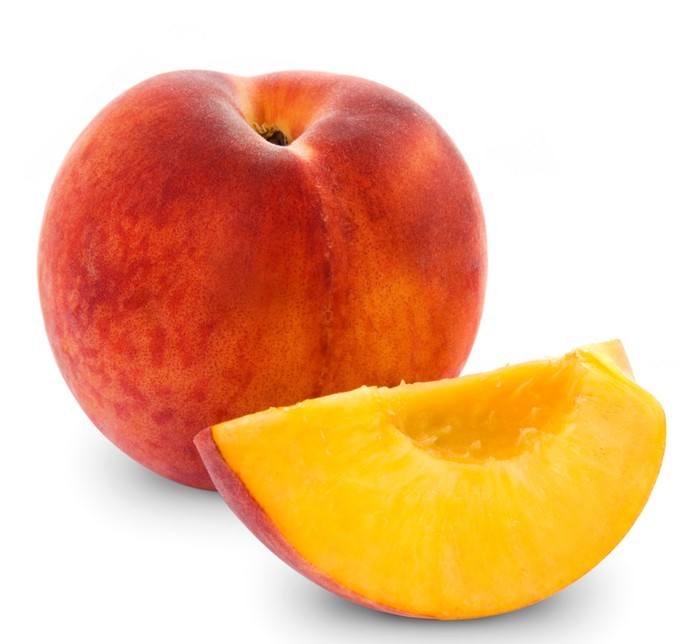}&  
    \includegraphics[width=.115\textwidth]{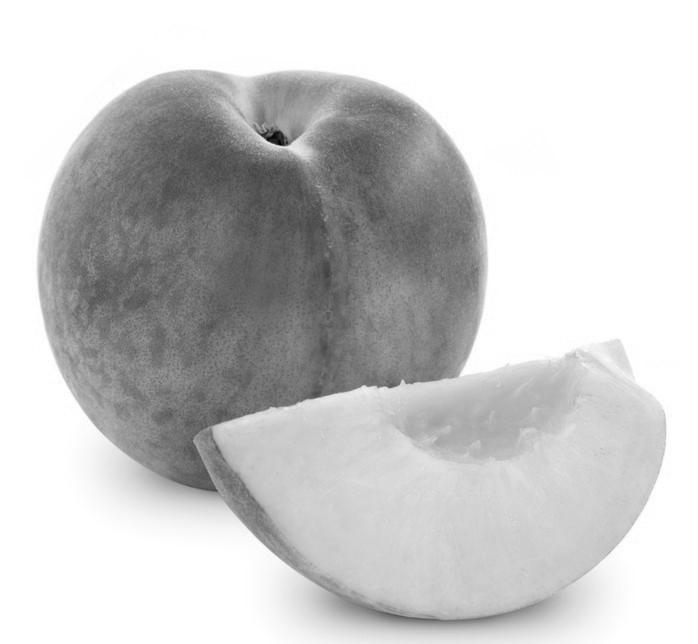}&
    \includegraphics[width=.115\textwidth]{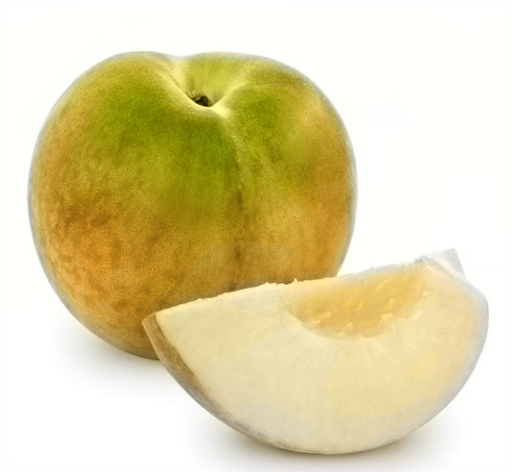}&   
	\includegraphics[width=.115\textwidth]{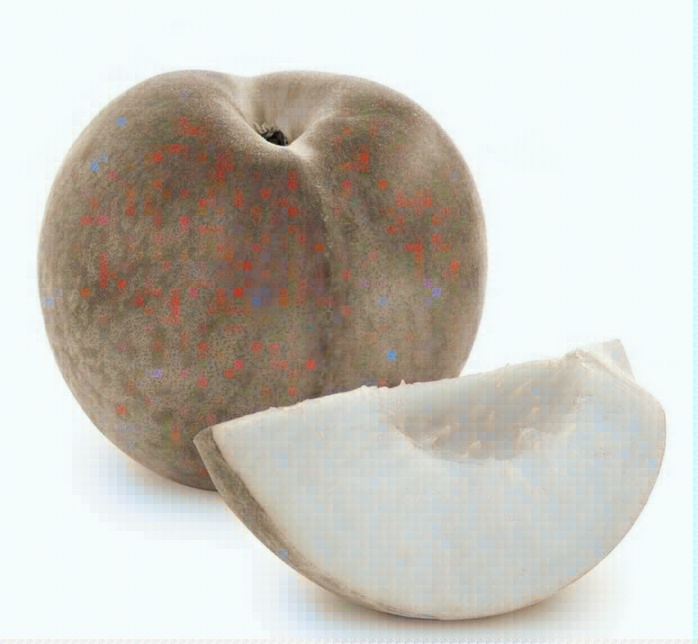}&    
	\includegraphics[width=.115\textwidth]{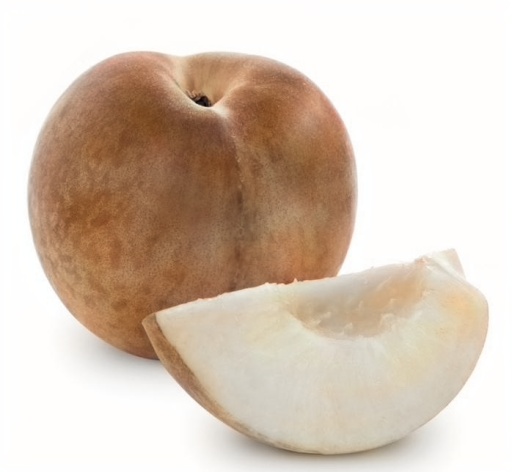}&    
	\includegraphics[width=.115\textwidth]{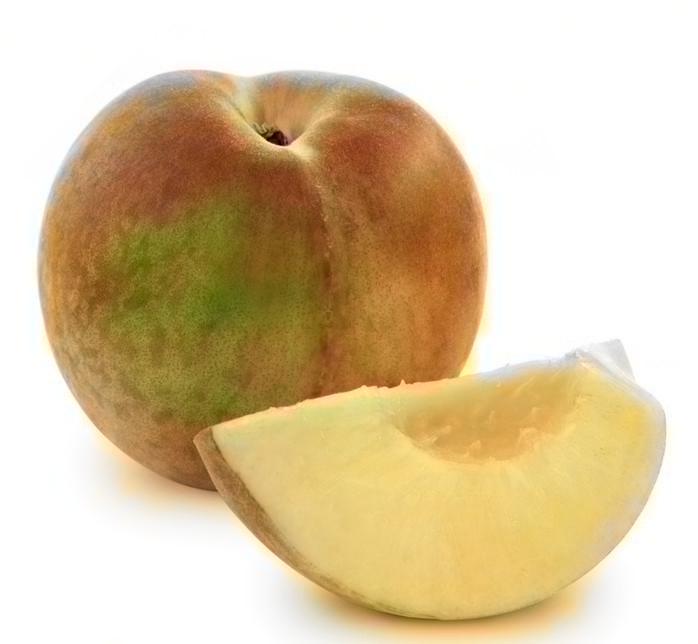}& 
	\includegraphics[width=.115\textwidth]{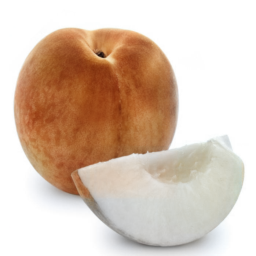}& 
  	\includegraphics[width=.115\textwidth]{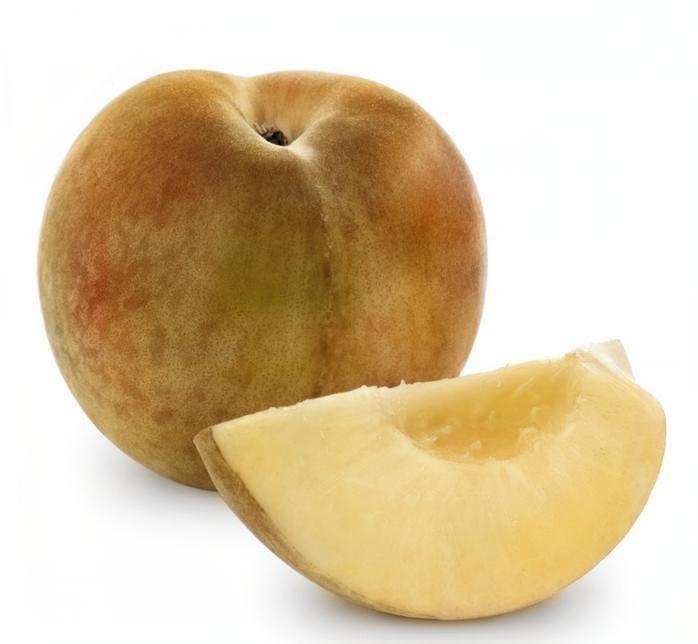}\\

  Original & Grayscale& \cite{larsson2016learning} & \cite{ozbulak2019image} &  \cite{iizuka2016let} &  \cite{zhang2016colorful} &  \cite{su2020CVPR} &  \cite{zhang2017real}\\
 \end{tabular}
\end{center}
\caption{Visual comparison of colorization algorithms on different fruit images from the Natural-Color Dataset. State-of-the-art colorization algorithms are unable to colorize the images effectively.}
\label{fig:Comparisons_fruit}
\end{figure}
\subsubsection{Instance-Aware Image Colorization}
The existing colorization models usually fail to colorize the images with multiple objects. To solve this issue, Su~\etal~\cite{su2020CVPR} proposed an instance-aware image colorization method\footnote{Code is available at \url{https://cgv.cs.nthu.edu.tw/projects/instaColorization}}. The network includes three parts: 1) an off-the-shelf pre-trained model to detect object instances and produce cropped object images, 2) two backbone networks trained end-to-end, for instance, and full-image colorization, and 3) a fusion module to selectively blend features extracted from different layers of the two colorization networks. Specifically, a grayscale image is fed to the network as input. The network first detects the object bounding boxes using an off-the-shelf object detector. Then, each detected instance is cropped out and forwarded to the instance colorization network, for instance, colorization. At the same time, the input grayscale image is also sent to another instance colorization network (with the same structure but different weights) for full-image colorization. Finally, a fusion module fuses all the instance features with the full image feature at each layer until the output colored image is obtained. In the training phase, a sequential approach strategy is adopted that trains the full-image network followed by the instance network and finally trains the feature fusion module by freezing the above two networks. The smooth-$\ell_{1}$ loss is used to train the network.

The authors use three datasets for training and evaluation, including ImageNet~\cite{deng2009imagenet}, COCO-Stuff~\cite{caesar2018coco}, and Places205 \cite{Place205}. Moreover, the images are resized to a size of 256 $\times$ 256.

\section{Experiments}

\subsection{Datasets}
The data sets available for evaluation are the most commonly used in the literature for other tasks such as detection, classification, segmentation \etc The images are first converted to grayscale; then, colorization models are applied to analyze their performance. Hence, we provide a new dataset explicitly designed for the colorization task in the next section while we list the currently used datasets below. 

\begin{itemize}

\item \textbf{COCO-stuff dataset~\cite{caesar2018coco}}: The Common Objects in COntext-stuff (COCO-stuff) dataset is constructed by annotating the original COCO dataset~\cite{lin2014coco}, which originally came with annotations of only things while neglecting the annotations of stuff. There are 164k images in the COCO-stuff dataset, which spans over 172 categories, including 80 things, 91 stuff, and one unlabeled class. The methods are evaluated using 5k images from the original validation set.

\item \textbf{Places205~\cite{zhou2014places}} has 20,500 test images from 205 categories. This dataset is commonly used to evaluate various colorization algorithms to investigate their relative performance. It only evaluates the transferability of algorithms and does not have a training partition.

\begin{figure}[t]
\begin{center}
\begin{tabular}{c@{ }  c@{ } c@{ } c@{ } c@{ }	c@{ }	c@{ }  c}
    \includegraphics[width=.115\textwidth]{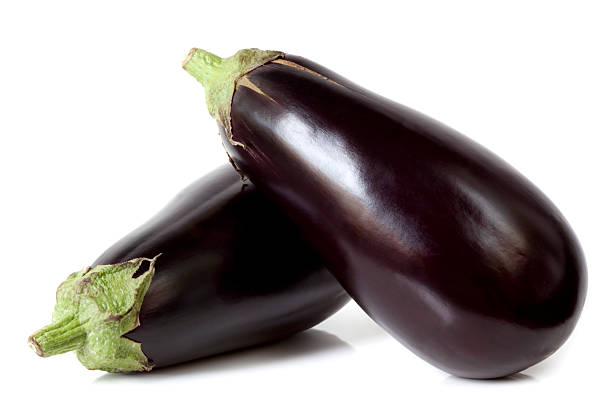}&  
    \includegraphics[width=.115\textwidth]{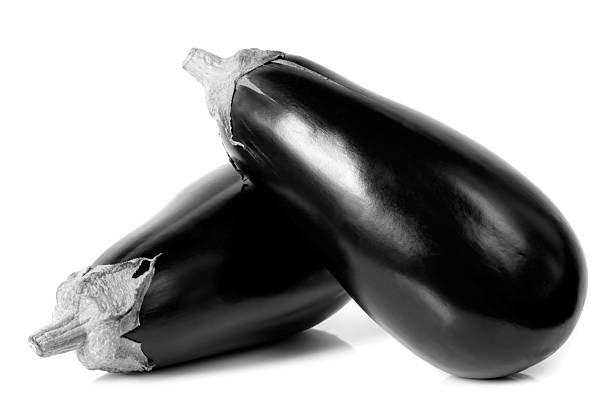}&
    \includegraphics[width=.115\textwidth]{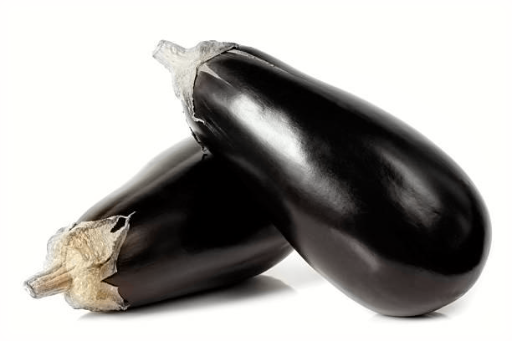}&   
	\includegraphics[width=.115\textwidth]{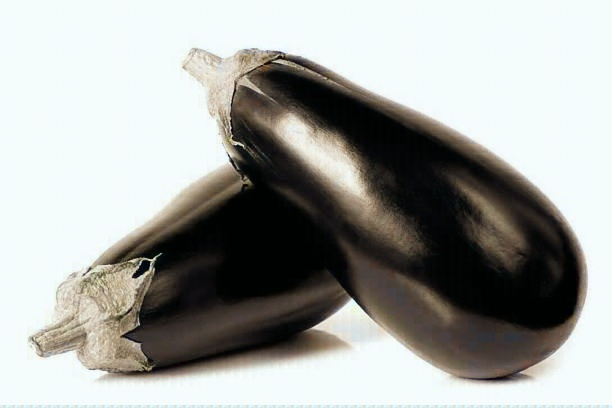}&    
	\includegraphics[width=.115\textwidth]{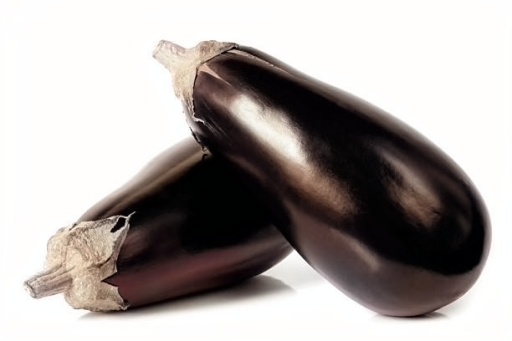}&    
	\includegraphics[width=.115\textwidth]{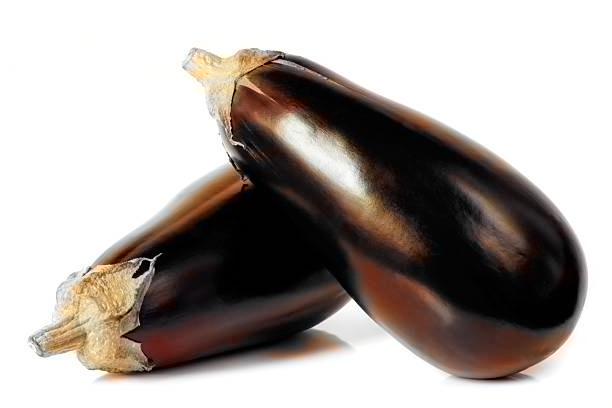}& 
	\includegraphics[width=.115\textwidth]{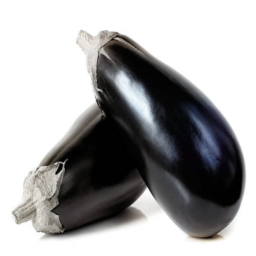}& 
  	\includegraphics[width=.115\textwidth]{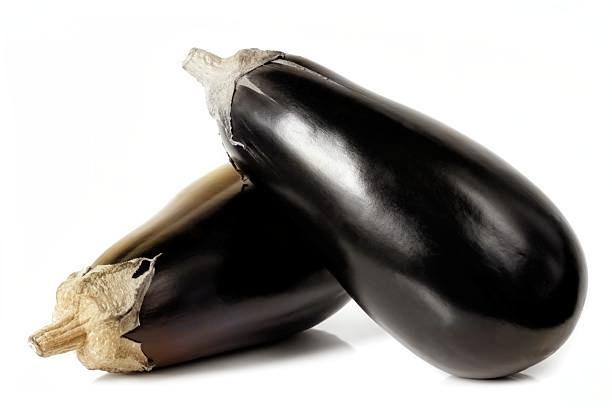}\\
  	
  	\includegraphics[width=.115\textwidth]{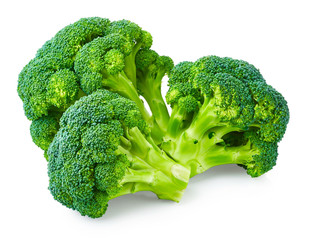}&  
    \includegraphics[width=.115\textwidth]{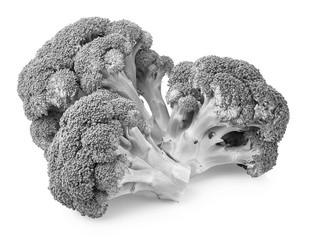}&
    \includegraphics[width=.115\textwidth]{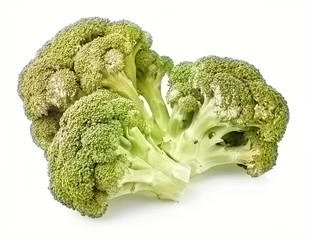}&   
	\includegraphics[width=.115\textwidth]{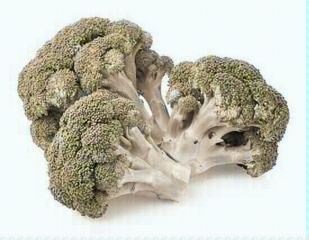}&    
	\includegraphics[width=.115\textwidth]{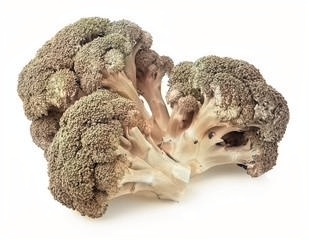}&    
	\includegraphics[width=.115\textwidth]{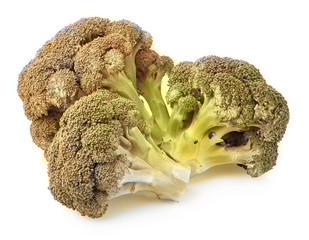}& 
	\includegraphics[width=.115\textwidth]{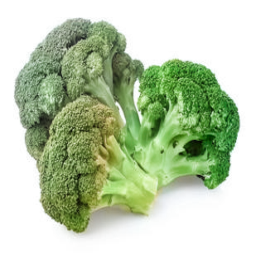}& 
  	\includegraphics[width=.115\textwidth]{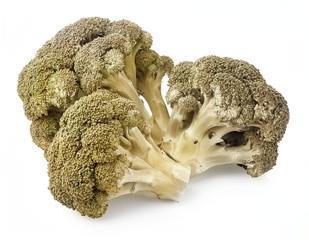}\\

  \includegraphics[width=.115\textwidth]{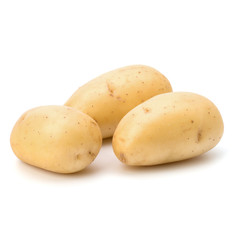}&  
    \includegraphics[width=.115\textwidth]{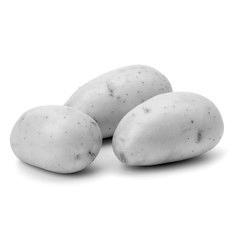}&
    \includegraphics[width=.115\textwidth]{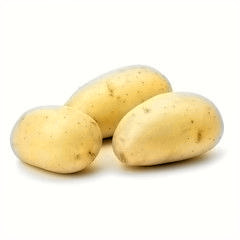}&   
	\includegraphics[width=.115\textwidth]{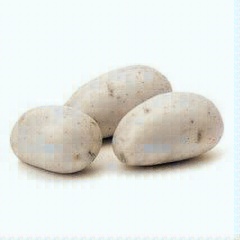}&    
	\includegraphics[width=.115\textwidth]{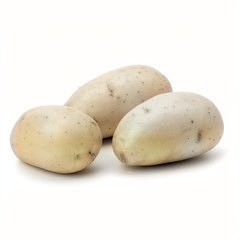}&    
	\includegraphics[width=.115\textwidth]{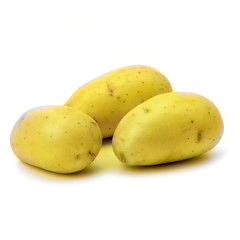}& 
	\includegraphics[width=.115\textwidth]{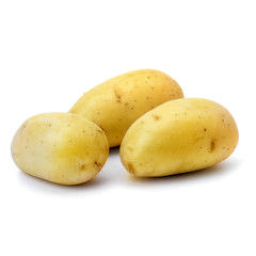}& 
  	\includegraphics[width=.115\textwidth]{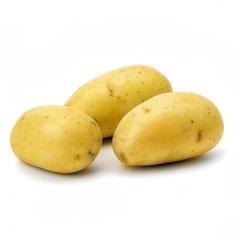}\\

  	\includegraphics[width=.115\textwidth]{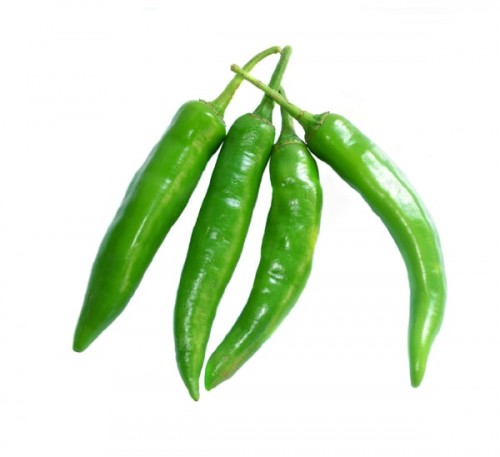}&  
    \includegraphics[width=.115\textwidth]{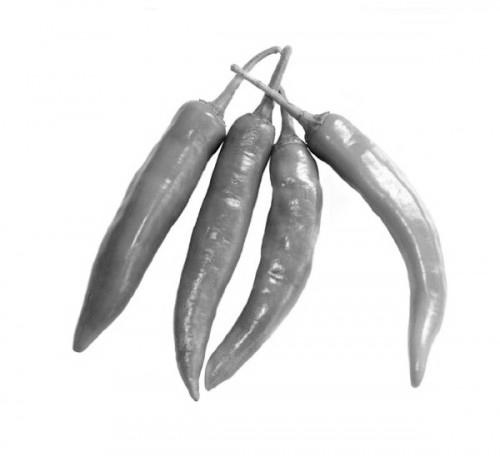}&
    \includegraphics[width=.115\textwidth]{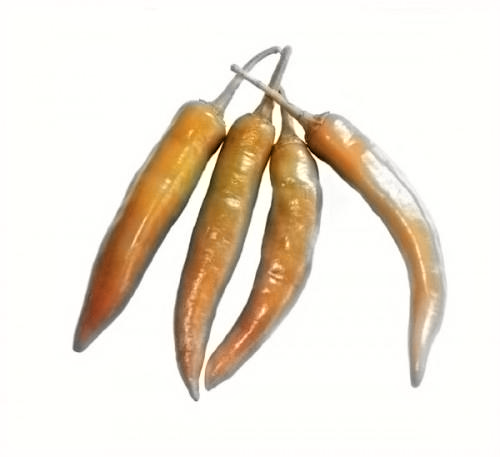}&   
	\includegraphics[width=.115\textwidth]{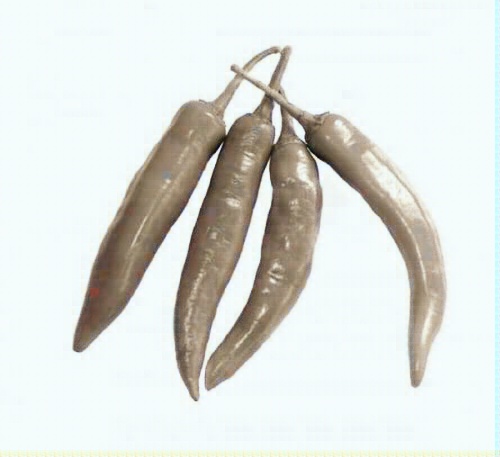}&    
	\includegraphics[width=.115\textwidth]{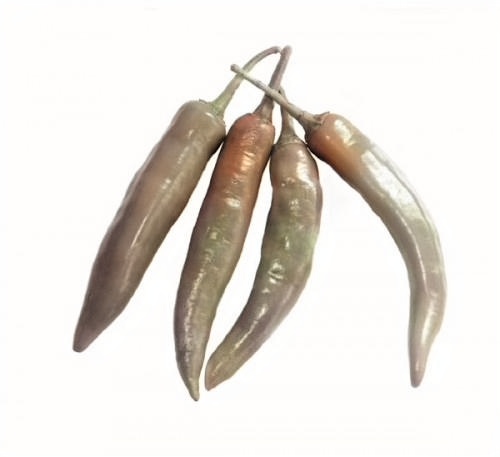}&    
	\includegraphics[width=.115\textwidth]{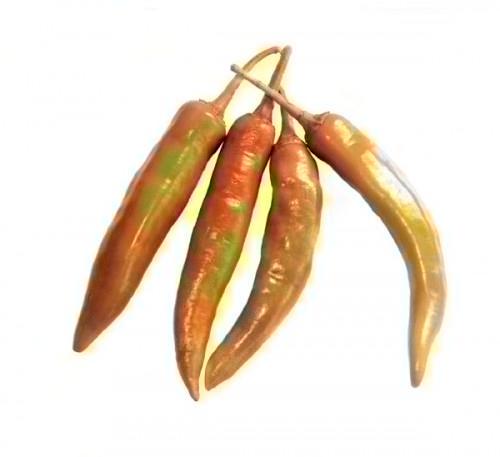}& 
	\includegraphics[width=.115\textwidth]{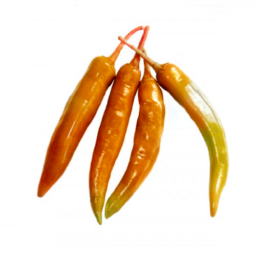}& 
  	\includegraphics[width=.115\textwidth]{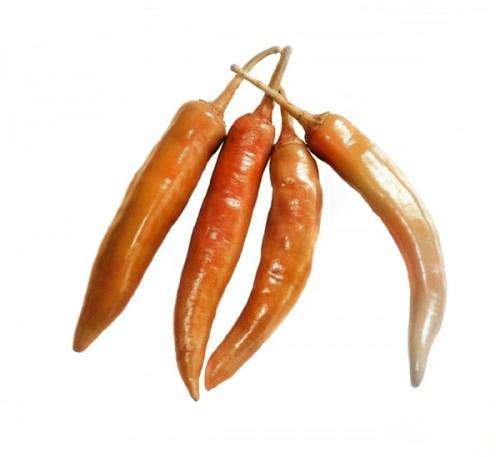}\\

  	\includegraphics[width=.115\textwidth]{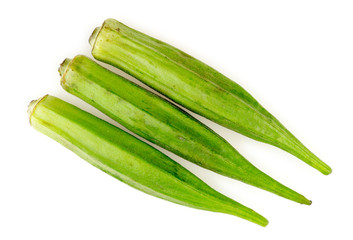}&  
    \includegraphics[width=.115\textwidth]{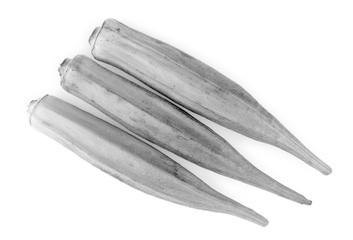}&
    \includegraphics[width=.115\textwidth]{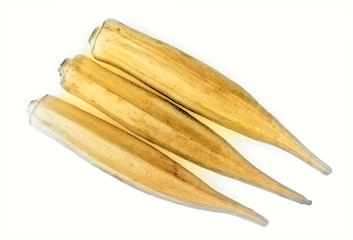}&   
	\includegraphics[width=.115\textwidth]{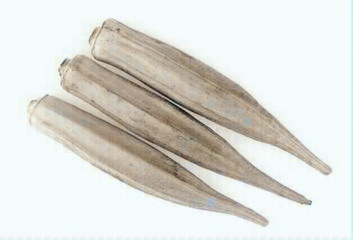}&    
	\includegraphics[width=.115\textwidth]{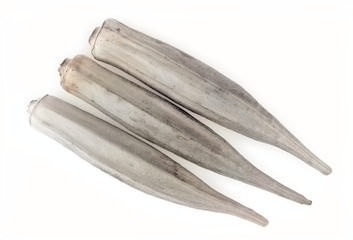}&    
	\includegraphics[width=.115\textwidth]{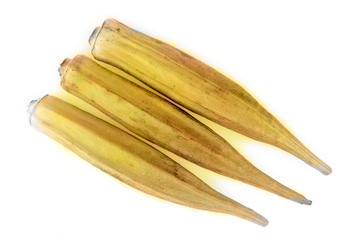}& 
	\includegraphics[width=.115\textwidth]{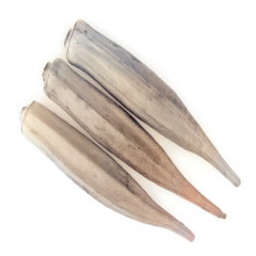}& 
  	\includegraphics[width=.115\textwidth]{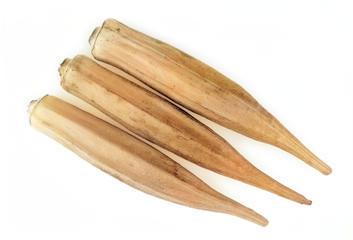}\\

  Original & Grayscale& \cite{larsson2016learning} & \cite{ozbulak2019image} &  \cite{iizuka2016let} &  \cite{zhang2016colorful} &  \cite{su2020CVPR} &  \cite{zhang2017real}\\
 \end{tabular}
\end{center}
\caption{Qualitative comparison on a few sample images of vegetables from Natural-Color Dataset. Most of the algorithms fail to reproduce the original colors.}
\label{fig:Comparisons_vegetables}
\end{figure}

\item \textbf{PASCAL VOC dataset~\cite{everingham2015pascal}}: PASCAL Visual Object Classes (PASCAL VOC) dataset has more than 11000 images that are divided into 20 object categories.

\item \textbf{CIFAR datasets~ \cite{krizhevsky2009learning}}: CIFAR-10 and CIFAR-100 are two subsets created and reliably labeled from 80 million tiny image dataset \cite{torralba200880}. CIFAR-10 is comprised of 60k images equally distributed over ten mutually exclusive categories, with 6k in each image category. On the other hand, CIFAR-100 has the same images distributed over 100 categories, with 600 images assigned to each category. Each image in both the subsets is of size $32\times32$ pixels. In CIFAR-100, two-level labeling is used. There are 20 superclasses at the higher level, each further divided into five subclasses.Overall, 50k and 1k images comprise training and testing sets, respectively.

\item \textbf{ImageNet ILSVRC2012~\cite{deng2009imagenet}}: This dataset contains 1.2 million high-resolution training images spanning over 1k categories, where 50k images comprise the hold-out validation set. The split employed for testing, i.e., ctest10k, is from~\cite{larsson2016learning} containing 10k images for evaluation.

\item \textbf{Palette-and-Text dataset~\cite{bahng2018coloring}}: is constructed by modifying the data collected from color-hex.com, where users upload user-defined color palettes with label names of their choice. The authors first collected 47,665 palette-text pairs, removing non-alphanumeric and non-English words from the collection. After removing text-palette pairs that lack semantic relationships, the final curated dataset contains 10,183 textual phrases with their corresponding five-color palettes.
\end{itemize}

\subsection{Evaluation Metrics}
The metrics typically used to assess colorization quality are either subjective or commonly used mathematical metrics, such as PSNR and SSIM~\cite{Wang2004}. Subjective evaluation is the gold standard for many applications in which humans determine the accuracy of the algorithm's output.

Although colorization uses subjective evaluation to some extent, it has two limitations: 1) scaling is extremely challenging, and 2) accurately determining the color is also very difficult. On the other hand, the mathematical metrics are general and may not provide the algorithms' accurate performance. We offer more insights into the metrics for colorization in the \enquote{Future Directions} section.

\section{Comparisons}
\noindent
\textbf{Qualitative Comparisons:}
We first compare the colorization methods on the existing datasets and then on our newly collected natural color dataset (NCD).

\textit{Existing Datasets Comparisons:} The commonly used datasets for image colorizations are Places205~\cite{zhou2014places}, ImageNet ILSVRC2012~\cite{deng2009imagenet} and COCO-stuff~\cite{caesar2018coco} datasets. The performances of various methods are given in Table~\ref{table:other_Comparisons}, where the numbers generated by each method are more or less similar for SSIM and PSNR, but the results are better for instance-aware colorization in terms of LPIPS.

\begin{table*}[t]
\caption{Comparison of state-of-the-art methods for colorization in terms of PSNR, SSIM, and LPIPS on our ImageNet, COCOStuff, and Places205 datasets. Higher values of metrics indicate better performance.}
\rowcolors{2}{gray!12}{white}
\centering
\begin{adjustbox}{max width=\textwidth}
\begin{tabular}{l||ccc|ccc|ccc}\hline
\rowcolor{gray!25}
\multicolumn{1}{c}{Method}               				&\multicolumn{3}{c}{Imagenet ctest10k}     &\multicolumn{3}{c}{COCOStuff} 	&\multicolumn{3}{c}{Places205}\\ \hline 
\rowcolor{gray!25}									&LPIPS$\downarrow$ &PSNR$\uparrow$    &SSIM$\uparrow$   &LPIPS$\downarrow$ &PSNR$\uparrow$    &SSIM$\uparrow$         	&LPIPS$\downarrow$ &PSNR$\uparrow$    &SSIM$\uparrow$   \\ \hline \hline
Let there be Color~\cite{iizuka2016let}   		&0.200 &23.64 &0.917 &0.185 &23.86 &0.922       	&0.146 &25.58 &0.950 \\
Automatic Colorizer~\cite{larsson2016learning}  &0.188 &25.11 &0.927 &0.183 &25.06 &0.930       	&0.161 &25.72 &0.951 \\
Colorful Colorization~\cite{zhang2016colorful}  &0.238 &21.79 &0.892 &0.234 &21.84 &0.895       	&0.205 &22.58 &0.921 \\
Real-Time Colorization~\cite{zhang2017real}     &0.145 &26.17 &0.932 &0.138 &26.82 &0.937       	&0.149 &25.82 &0.948 \\
Deoldify~\cite{antic2018deep} 					&0.187 &23.54 &0.914 &0.180 &23.69 &0.920       	&0.161 &23.98 &0.939 \\
Fully-Automatic Colorization~\cite{lei2019fully}&0.202 &24.52 &0.917 &0.191 &24.59 &0.922       	&0.175 &25.07 &0.942 \\
Instance-aware Colorization~\cite{su2020CVPR}   &0.134 &26.98 &0.933 &0.125 &27.78 &0.940       	&0.130 &27.17 &0.954 \\ \hline 
\end{tabular}
\end{adjustbox}
\label{table:other_Comparisons}
\end{table*}

\textit{NCD Comparisons:} The networks mentioned in section~\ref{sec:SIDC} are evaluated on the peak signal-to-noise ratio (PSNR), the structural similarity index (SSIM)~\cite{Wang2004}, patch-based contrast quality index (PCQI), and underwater image quality measure (UIQM) measures. Table~\ref{table:NCD_Comparisons} presents the results for each category for all measures on NCD. Real-time colorization~\cite{zhang2017real} achieves the high performance of \textbf{21.93 dB} and \textbf{0.881} for PSNR and SSIM against other competitive measures. Furthermore, the PCQI performance of instance-aware colorization~\cite{su2020CVPR} and IQM achievement of colorful colorization~\cite{zhang2016colorful} are higher than those of state-of-the-art methods. However, declaring one method against the other may not be a simple task due to the involvement of many various elements such as the number of parameters, depth of the network, the number of images for training, the datasets employed, the size of the training patch, the number of feature maps, the network complexity, etc. For a fair comparison, the only possible approach is to ensure that all methods have similar elements, as mentioned earlier.

\vspace{1mm}
\noindent
\textbf{Quantitative Comparisons:} We present the visual colorization comparisons on fruits and vegetables in Figure~\ref{fig:Comparisons_fruit} and~\ref{fig:Comparisons_vegetables}, respectively, for a few state-of-the-art algorithms. We can observe that most algorithms fail to recover the original natural colors for most images. Though Real-Time~\cite{zhang2017real} and instance-aware~\cite{su2020CVPR} colorization algorithms provide consistent performance and colors closer to the original objects; however, the algorithms are still far from delivering accurate colorization performance. The experiments on the proposed Natural-Color Dataset (NCD) show the limitations of the state-of-the-art algorithms and encourage the authors to explore novel colorization techniques.

\begin{table*}
\caption{Comparisons of the state-of-the-art methods for the colorization in terms of PSNR, SSIM, PCQI, and IQM on our Natural-Color Dataset. The higher value of the metrics indicates better performance.}
\rowcolors{2}{gray!12}{white}
\centering
\begin{adjustbox}{max width=\textwidth}
\begin{tabular}{l||c|cccc|cccc|cccc|cccc|cccc|cccc}\hline 
\rowcolor{gray!25}

Category & No. of  & \multicolumn{4}{|c|}{Automatic Colorizer \cite{larsson2016learning}} & \multicolumn{4}{|c|}{ColorCapsNet \cite{ozbulak2019image}} & \multicolumn{4}{|c|}{Let there be Color \cite{iizuka2016let}} & \multicolumn{4}{|c|}{Colorful Colorization \cite{zhang2016colorful}} & \multicolumn{4}{|c|}{Instance-Aware Colorization \cite{su2020CVPR}} & \multicolumn{4}{|c}{Real-Time Colorization \cite{zhang2017real}}\\
\rowcolor{gray!25}
& Images& PSNR & SSIM & PCQI & IQM & PSNR & SSIM & PCQI & IQM & PSNR & SSIM & PCQI & IQM & PSNR & SSIM & PCQI & IQM & PSNR & SSIM & PCQI & IQM & PSNR & SSIM & PCQI & IQM\\ \hline \hline
Apple  	       &39  & 16.66 & 0.772 & 0.881 & 1.009 & 17.64 & 0.625 & 0.825 & 0.869 & 20.17 & 0.820 & 0.918 & 0.944 & 18.71 & 0.792 & 0.922 & 1.095 & 21.94 & 0.903 & 0.918 & 0.918 & 21.27 & 0.888 & 0.915 & 1.104 \\
Banana         &44  & 17.83 & 0.792 & 0.937 & 0.688 & 16.43 & 0.598 & 0.820 & 0.652 & 16.74 & 0.733 & 0.936 & 0.634 & 22.22 & 0.894 & 0.948 & 0.813 & 22.14 & 0.889 & 0.947 & 0.643 & 23.34 & 0.923 & 0.948 & 0.780 \\ 
Brinjal        &35  & 27.21 & 0.849 & 0.946 & 1.179 & 26.32 & 0.720 & 0.876 & 1.015 & 27.74 & 0.866 & 0.939 & 1.094  & 24.52 & 0.829 & 0.946 & 1.304 & 25.59 & 0.843 & 0.949 & 1.134  & 27.56  &  0.855  &  0.944  &  1.242 \\
Broccoli       &35  & 18.57 & 0.870 & 0.895 & 1.419 & 18.48 & 0.717 & 0.865 & 1.232 & 18.59 & 0.816 & 0.918 & 1.301 & 19.34 & 0.842 & 0.920 & 1.483 & 20.11 & 0.851 & 0.964 & 1.431  & 19.54 & 0.854 & 0.918 & 1.435 \\ 
Capsicum green &35  & 18.72  & 0.812 & 0.894 & 1.097 & 19.67 & 0.646 & 0.840 & 0.918 & 19.35 & 0.731 & 0.914 & 0.970 & 19.91 & 0.807 & 0.917 & 1.168 & 18.02 & 0.752 & 0.927 & 1.140 & 19.32 &  0.789 & 0.910 & 1.118 \\ 
Carrot         &39  & 23.10 & 0.929 & 0.931 & 0.921 & 17.65 & 0.655 & 0.852 & 0.748 & 19.40 & 0.826 & 0.927 & 0.825  & 21.16 & 0.917 & 0.939 & 1.026 & 22.00 & 0.908 & 0.932 & 0.832  & 22.62 & 0.937 & 0.936 & 1.005 \\ 
Cherry         &34  & 23.09 & 0.892 & 0.920 & 1.055 & 19.71 & 0.656 & 0.839 & 0.927 & 22.74 & 0.868 & 0.919 & 0.993  & 23.93 & 0.896 & 0.918 & 1.197 & 23.47 & 0.891 & 0.919 & 1.042  & 24.28 & 0.902 & 0.924 & 1.169 \\ 
Chilli green   &36  & 20.82 & 0.868 & 0.944 & 1.115 & 20.77 & 0.716 & 0.889 & 0.882 & 20.86 & 0.844 & 0.944 & 0.962  & 20.94 & 0.872 & 0.948 & 1.224 & 20.38 & 0.858 & 0.963 & 1.193  &  21.39 & 0.878 & 0.946 & 1.154 \\ 
Corn           &36  & 19.41 & 0.873 & 0.911 & 1.079 & 15.34 & 0.584 & 0.820 & 0.925  & 17.03 & 0.780 & 0.903 & 0.997  & 20.25 & 0.888 & 0.905 & 1.163 & 19.15 & 0.849 & 0.912 & 1.035  & 20.44 & 0.910 & 0.902 & 1.146 \\ 
Cucumber       &35  & 22.58 & 0.879 & 0.919 & 1.304 & 22.73 & 0.725 & 0.862 & 1.059 & 22.70 & 0.834 & 0.927 & 1.157  & 21.63 & 0.837 & 0.931 & 1.409 & 21.34 & 0.817 & 0.955 & 1.310  &  23.24 & 0.865 & 0.929 & 1.321 \\ 
Lady Finger    &36  & 20.66 & 0.874 & 0.925 & 1.189 & 21.42 & 0.721 & 0.876 & 0.878 & 21.69 & 0.848 & 0.926 & 0.990  & 21.64 & 0.882 & 0.928 & 1.246 & 21.50 & 0.866 & 0.954 & 1.110  & 23.04  &  0.903 & 0.926 & 1.148 \\ 
Lemon          &39  & 19.15 & 0.877 & 0.873 & 0.899 & 14.47 & 0.551 & 0.766 & 0.786  & 15.37 & 0.697 & 0.873 & 0.816  & 20.43 & 0.878 & 0.875 & 0.978 & 21.13 & 0.894 & 0.895 & 0.841  & 21.20 & 0.901 & 0.876 & 0.962
 \\ 
Orange        & 29  & 19.27 & 0.905 & 0.873 & 0.896 & 13.98 & 0.536 & 0.784 & 0.793  & 15.16 & 0.696 & 0.894 & 0.788  & 19.49 & 0.895 & 0.899 & 0.989 & 21.44 & 0.915 & 0.905 & 0.816  & 21.06  &  0.909 &   0.898 &   0.991 \\ 
Peach          &27  & 17.99 & 0.841 & 0.849 & 0.882 & 16.09 & 0.550 & 0.789 & 0.820 & 18.63 & 0.797 & 0.876 & 0.811  & 19.07 & 0.857 & 0.902 & 1.010 & 19.39 & 0.833 & 0.878 & 0.739  & 19.53  &  0.861  &  0.901 &   0.997 \\ 
Pear           &28  & 19.60 & 0.883 & 0.926 & 0.985 & 16.72 & 0.591 & 0.835 & 0.831 & 18.59 & 0.796 & 0.932 & 0.888  & 23.13 & 0.917 & 0.943 & 1.057 & 21.14 & 0.893 & 0.952 & 0.915 & 23.87  &  0.935  &  0.940 &    1.026 \\ 
Plum           &35  & 21.48 & 0.726 & 0.888 & 1.212 & 22.14 & 0.626 & 0.836 & 1.063 & 23.43 & 0.768 & 0.898 & 1.114  & 21.87 & 0.747 & 0.901 & 1.310 & 20.59 & 0.734 & 0.912 & 1.195  & 23.37  &  0.764 &   0.901 &   1.263 \\ 
Pomegranate    &55  & 19.47 & 0.875 & 0.901 & 1.285 & 16.44 & 0.602 & 0.843 & 1.112 & 17.56 & 0.740 & 0.915 & 1.165  & 19.25 & 0.853 & 0.925 & 1.338 & 19.92 & 0.863 & 0.949 & 1.240  & 18.72 &   0.817 &  0.922 & 1.295 \\ 
Potato         &35  & 24.48 & 0.939 & 0.862 & 1.028 & 18.83 & 0.622 & 0.766 & 0.869  & 21.81 & 0.854 & 0.862 & 0.945 & 22.43 & 0.935 & 0.860 & 1.102 & 23.53 & 0.935 & 0.892 & 1.022 & 23.19 & 0.946  &  0.857 &   1.046 \\ 
Strawberry  & 35  & 20.84 & 0.920 & 0.955 & 1.444 & 15.48 & 0.621 & 0.880 & 1.157 & 15.96 & 0.743 & 0.932 & 1.222  & 19.13 & 0.881 & 0.932 & 1.485 & 19.59 & 0.891 & 0.972 & 1.440  & 20.24  &  0.898  &  0.920 &   1.466 \\ 
Tomato         &48  & 18.66 & 0.883 & 0.893 & 0.943 & 15.43 & 0.584 & 0.819 & 0.815 & 17.44 & 0.777 & 0.918 & 0.858 & 19.49 & 0.876 & 0.918 & 1.043 & 20.04 & 0.878 & 0.924 & 0.901  & 21.30 &   0.895 &   0.913 &    1.036 \\ \hline \hline
\textbf{Dataset-level} &\textbf{735}   & 20.48 & 0.863 & 0.906 & 1.082 & 18.29 & 0.632 & 0.834 & 0.917 & 19.55 & 0.792 & 0.914 & 0.974 & 20.93 & 0.865 & 0.919 & \textbf{1.172} & 21.12 & 0.863 & \textbf{0.931} & 1.045  & \textbf{21.93} & \textbf{0.881} & 0.916 & 1.135 \\ \hline
\end{tabular}
\end{adjustbox}
\label{table:NCD_Comparisons}
\end{table*}

\section{Limitations and Future Directions}
\noindent
\textbf{Lack of Appropriate Evaluation Metrics}: As mentioned earlier, colorization papers typically employ metrics from image restoration tasks, which may not be appropriate for the task at hand. We also suggest comparing only the predicted color channels, i.e., U-channel and V-channel, instead of YUV. Similarly, it will be highly sought after to use metrics specifically designed to take color into account, such as PCQI~\cite{wang2015PCQI} and UIQM~\cite{panetta2015UIQM}.

PCQI stands for patch-based contrast quality index. It is a patch-based evaluation metric. It considers three statistics of a patch, i.e., mean intensity ($p_m$), signal strength or contrast change ($p_c$), and structure distortion ($p_s$) for comparison with ground truth. It can be expressed as

\begin{equation}
PCQI = p_m(x, y) \cdot p_c(x, y) \cdot p_s(x, y).
\label{eq:pcqi}
\end{equation}

Similarly, UIQM is an abbreviation for the underwater image quality measure. It differs from earlier defined evaluation metrics because it does not require a reference image. Like PCQI, it also depends on three measures, i.e., image colorfulness, image sharpness, and image contrast. It can be formulated as:

\begin{equation}
IQM = w_1 (ICM) +w_2 (ISM)+w_3 (IConM),
\label{eq:uiqm}
\end{equation}
where ICM, ISM, and IConM stand for image colorfulness, image sharpness and image contrast, respectively, while \enquote{$w$} controls the weight of each quantity.

\vspace{1mm}
\noindent
\textbf{Lack of Benchmark Dataset}:
Due to the lack of purposefully built colorization datasets, researchers typically evaluate image colorization techniques on grayscale images from various datasets available in the literature. Originally, researchers collected the public datasets for tasks like detection, classification, etc., not for image colorization. The quality of the images may not be sufficient for image colorization. Moreover, the datasets contain objects, such as buses, shirts, doors, etc., that can take any color. Hence, such an evaluation environment is not appropriate for fair comparison in terms of PSNR or SSIM.

We aim to remove this unrealistic setting for image colorization by collecting images that are true to their colors. For example, a carrot will have an orange color in most images. Bananas will be either greenish or yellowish. Hence, our choice of images deliberately evaluates the method's strength and any bias toward specific colors. Furthermore, we purposefully put a white background to test for any algorithm-induced colorization spill. We have collected \emph{723} images from the internet distributed in \emph{20} categories. Each image has an object and a white background. Our benchmark outlines a realistic evaluation scenario that differs sharply from those generally employed by image colorization techniques. Figure~\ref{fig:Sample_images} shows the images from each category from our Natural-Color Dataset (NCD).

\vspace{1mm}
\noindent
\textbf{Lacking of Competitions}: Currently, competitions for most vision tasks are held across various top-tier conferences, such as CVPR, ECCV workshops like NTIRE, PBVS, etc., and online submission platforms like Kaggle. These competitions help push the state-of-the-art. Unfortunately, there is no such arrangement for image colorization. Making competitions a regular feature in top-tier conferences and online venues would thus be a drastic step forward for image colorization.

\vspace{1mm}
\noindent
\textbf{Limited Availability of Open-Source Codes}: Open-source code plays a vital role in the advancement of the research fields, as can be seen in classification~\cite{he2016ResNet}, image super-resolution~\cite{anwar2019densely}, image denoising~\cite{anwar2019real}, etc. In image colorization, open-source codes are rare, or codes are obsolete now as they were built in earlier CNN frameworks. In other research fields, volunteers recycle or re-implement the codes for the new frameworks and environments. However, this is not currently done for image colorization, hindering progress.

\section{Conclusion}
Single image colorization is a research problem with critical real-life applications. The exceptional success of deep learning approaches has led to a rapid growth in deep convolutional techniques for image colorization. Based on exciting innovations, various methods are proposed for exploiting network structures, training methods, learning paradigms, etc. This article presents a thorough review of deep-learning methods for image colorization.

We observe that image colorization performance has improved in recent years at the cost of increasing network complexity. However, inadequate metrics, network complexity, and failure to handle real-life degradations restrict the application of state-of-the-art methods to critical real-world scenarios.

We identified the following trends in image colorization: 1) GAN-based methods deliver diverse colorization visually compared to CNN-based methods; 2) the existing models typically produce a suboptimal result for complex scenes having a large number of objects with small sizes; 3) deep models with higher complexity have slight improvement in terms of numbers; 4) the diversity of networks in image colorization as compared to other image restoration is significant; 5) a future direction for image colorization is unsupervised learning; 6) many recent advancements and techniques such as attention mechanisms and loss functions can be incorporated for performance. We believe this article and our novel dataset will inspire additional efforts to address the aforementioned critical issues.

\bibliographystyle{spmpsci}   
\bibliography{ref}
\end{document}